%% file: main.tex
\documentclass[10pt,twocolumn,letterpaper]{article}

\usepackage{cvpr}              % To produce the CAMERA-READY version
\usepackage{minitoc}

\input{preamble}

\definecolor{cvprblue}{rgb}{0.21,0.49,0.74}
\usepackage[pagebackref,breaklinks,colorlinks,citecolor=blue,linkcolor=blue,urlcolor=blue]{hyperref}

\def\paperID{} % *** Enter the Paper ID here
\def\confName{CVPR}
\def\confYear{2026}

\title{Video-R2: Reinforcing Consistent and Grounded Reasoning \\ in Multimodal Language Models}

\author{
  Muhammad Maaz$^{1}$,~~
  Hanoona Rasheed$^{1}$,~~Fahad Shahbaz Khan$^{1,2}$,~~
  Salman Khan$^{1,3}$\\[0.25cm]
  \fontsize{10.5pt}{12pt}\selectfont
  $^{1}$Mohamed bin Zayed University of AI, $^{2}$Linköping University, $^{3}$Australian National University\\
  {\hypersetup{}
    \fontsize{11.5pt}{12pt}\selectfont
    \href{https://github.com/mbzuai-oryx/Video-R2}{https://github.com/mbzuai-oryx/Video-R2}
  }
}

\begin{document}

% % % % % % % % % TOC % % % % % % % % % 
\doparttoc 
\faketableofcontents 
% % % % % % % % % TOC % % % % % % % % % 

\twocolumn[{
\maketitle
\maketitle
\begin{center}
    \captionsetup{type=figure}
    \includegraphics[width=1\linewidth]{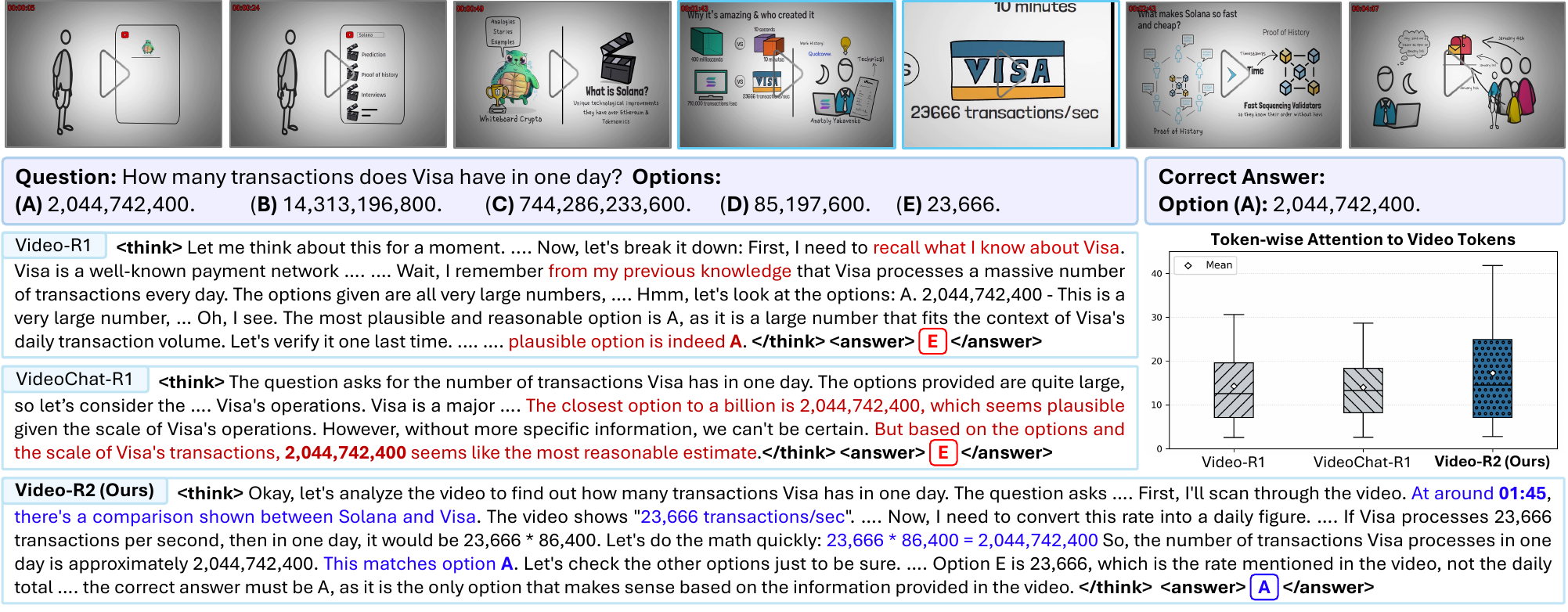}
    \vspace{-2em}
    \captionof{figure}{
    \textbf{Inconsistent reasoning in prior video-LLMs and improved visual reliance with~\MyModel.} 
Given the video and the question, \textit{``How many transactions does Visa have in one day?''}, 
both \textbf{Video-R1}~\cite{videor12025} and \textbf{VideoChat-R1}~\cite{videochatR12025} conclude option~A during their reasoning but finally predict option~E as answer, revealing inconsistent reasoning behavior where the model’s conclusion and final answer mismatch. 
%As shown by the attention scores on the right, 
This behaviour emerges because these models primarily rely on textual context and world knowledge, attending little to the video itself. 
In contrast, our \textbf{\MyModel} correctly identifies the on-screen visual cue at \texttt{01:45} 
(\textit{``23,666 transactions/sec''}), performs temporal conversion, and arrives at the correct daily value.
The box plot on the right shows the average attention from generated tokens to video tokens across all attention heads in the final transformer layer.
Compared with baselines, \textbf{\MyModel} allocates higher and more distributed attention to video tokens, 
indicating stronger and more adaptive \textit{visual reliance}. 
While prior models can produce plausible yet inconsistent reasoning,  \textbf{\MyModel}\ reasons coherently and depends on actual video evidence.
(Additional examples in \texttt{Appendix}~\ref{supp:tac},~\ref{supp:vas}).
}
    \label{fig:motivation}
\end{center}
}]

\input{sec/0_abstract}    
\input{sec/1_intro}
\input{sec/2_reasoning_quality_metrics}
\input{sec/3_temporal_alignment_reward}
\input{sec/4_experiments}
% \input{sec/5_discussion}
% \input{sec/related_work}
\input{sec/6_conclusion}
% \clearpage
% {
%     \small
%     \bibliographystyle{ieeenat_fullname}
%     \bibliography{main}
% }
\clearpage
\appendix

\input{supplemental/suppl}

\clearpage
{
    \small
    \bibliographystyle{ieeenat_fullname}
    \bibliography{main}
}

\end{document}

%% file: preamble.tex
\usepackage{lipsum}

\usepackage{booktabs}
\usepackage[table]{xcolor}
\usepackage{graphicx}
\usepackage{array}

\newcommand{\rothead}[1]{%
  \rotatebox[origin=br]{90}{\parbox[b]{2.6cm}{\raggedright #1}}%
}

\newcommand{\rotheadablation}[1]{%
  \rotatebox[origin=br]{90}{\parbox[b]{1.5cm}{\raggedright #1}}%
}

\usepackage{adjustbox}
\usepackage{amssymb}
\usepackage{multirow}
\usepackage{stfloats}
\usepackage{booktabs}
\usepackage{adjustbox}
\usepackage{pifont}
\usepackage{tcolorbox}
\tcbuselibrary{breakable}
\usepackage[normalem]{ulem}

\makeatletter
\renewcommand\numberline[1]{#1\kern0.06em\hspace*{1em}}
\makeatother

\definecolor{checkgreen}{HTML}{3e993b}
\definecolor{checkred}{HTML}{ea796c}
\definecolor{darkblue}{HTML}{252c94}

\newcommand{\cmark}{\ding{51}}
\newcommand{\xmark}{\ding{55}}

\newcommand{\MyModel}{Video-R2}

\usepackage{marvosym} % for \Letter

\makeatletter
\newcommand\emailfootnotetext[1]{%
  \begingroup
    \renewcommand\thefootnote{}% empty marker
    \footnotetext{\Letter~\href{mailto:#1}{#1}}%
    \addtocounter{footnote}{-1}% keep counter unchanged
  \endgroup
}

\newcommand\emailthanks[1]{%
  \begingroup
    \renewcommand\thefootnote{}% suppress star in author line
    \footnotemark
  \endgroup
  \protected@xdef\@thanks{\@thanks\protect\emailfootnotetext{#1}}%
}
\makeatother

%% file: sec/0_abstract.tex
\begin{abstract}
Reasoning over dynamic visual content remains a central challenge for multimodal large language models. 
Recent thinking models generate explicit reasoning traces for interpretability; however, their reasoning often `appears' convincing while being logically inconsistent or weakly grounded in visual evidence. 
We identify and formalize these issues through two diagnostic metrics: Think–Answer Consistency (TAC), which measures the alignment between reasoning and answers, and Video Attention Score (VAS), which captures the extent to which reasoning depends on visual versus textual cues. 
Analysis across 11 video reasoning benchmarks shows that current models rely heavily on linguistic priors rather than visual content.
To address this, we propose a reinforcement learning approach that enhances both temporal precision and reasoning consistency. 
Our approach combines timestamp-aware supervised fine-tuning with Group Relative Policy Optimization (GRPO) guided by a novel Temporal Alignment Reward (TAR). 
This dual-step post-training stage encourages temporally aligned and causally coherent video reasoning. 
The resulting model, \MyModel, achieves consistently higher TAC, VAS, and accuracy across multiple benchmarks, demonstrating that improvements in temporal alignment and reasoning coherence lead to more accurate and trustworthy video understanding.
\end{abstract}

%% file: sec/1_intro.tex
\section{Introduction}
\label{sec:intro}

Recent progress in multimodal large language models (MLLMs) has advanced visual understanding~\cite{liu2023llava,zhu2023minigpt4,maaz2023videochatgpt,videollama,llavav1.5,li2024llavanextinterleaved,maaz2024videogptplus,deitke2025molmo,chen2024internvl,qwen2025qwen25technicalreport,cho2025PerceptionLM}.
However, reasoning over dynamic visual content remains a persistent challenge.
Unlike static image tasks such as captioning or visual question answering, video reasoning requires the ability to infer temporal and causal relationships across sequences of frames~\cite{huang2023vtimellm,buch2025flexible,chen2024mecd,fei2024video,zhang2025vtimecot,han2025videoespresso,shi2025agentofthoughts,yuan2025videoexplorer,sridhar2025video,he2025enhancing,zheng2025villa,luo2025thinking}. 
This demands not only frame-level recognition but also reasoning about how fine-grained visual elements (objects, actions, events) evolve and interact over time.

Recent \textit{thinking} models such as Video-R1~\cite{videor12025}, VideoChat-R1/1.5~\cite{videochatR12025,yan2025videochat_r1.5}, and VideoRFT~\cite{videorft2025} 
address this challenge by generating explicit reasoning traces using \texttt{<think>} and \texttt{<answer>} formats. 
These traces 
aim
to make the 
reasoning process transparent and interpretable. 
However, our analysis reveals a concerning trend: \emph{the reasoning often sounds coherent but fails to correspond either to the final answer or to the actual visual evidence in the video.}
This observation motivates a deeper look at how current SoTA models reason about videos and where they fail. 

We begin by diagnosing two key issues that limit existing thinking models. 
\emph{a) Inconsistency between reasoning and answers.}~As shown in Fig.~\ref{fig:motivation}, the model may conclude one option in its reasoning but outputs a different answer, indicating poor internal alignment. 
We formalize this as \textbf{Think–Answer Consistency (TAC)} metric that measures whether the model’s reasoning logically supports its final answer.
\emph{b) Weak reliance on visual evidence.} 
Although the reasoning text often references the video, closer inspection shows minimal use of concrete visual details e.g., specific objects, actions, or timestamps.
Attention analysis (Fig.~\ref{fig:motivation}, right) confirms that the current models overwhelmingly attend to text tokens, assigning minimal attention weights to visual tokens.
To quantify this imbalance, we introduce the \textbf{Video Attention Score (VAS)}, which captures how strongly a model's reasoning is grounded in visual evidence rather than linguistic priors.
Fig.~\ref{fig:motivation} illustrates both issues. Although current models often seem to reason from visual content, their reasoning remains largely text-driven rather than visually grounded. This shows that answer accuracy alone is insufficient to assess reasoning quality, motivating TAC and VAS as complementary diagnostic metrics.

% % % % % % % % % % % % %  Intro Figure % % % % % % % % % % % % %
\begin{figure}[!t]
  \centering
  \includegraphics[width=1.0\columnwidth]{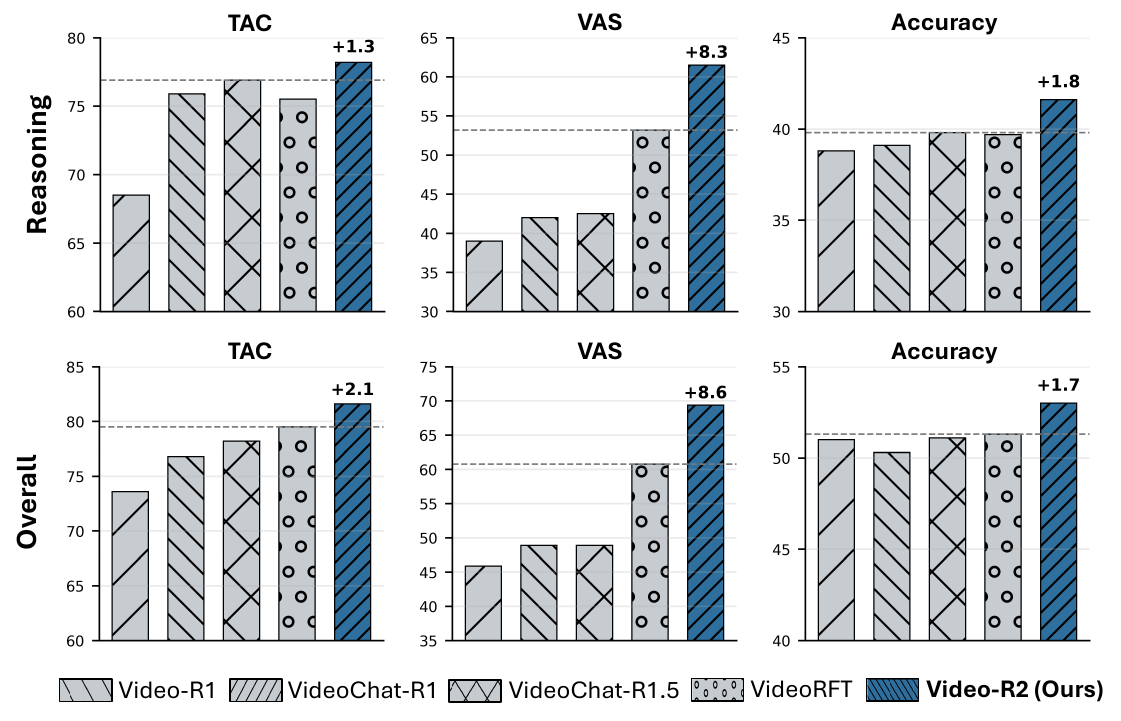}
  \vspace{-2em}
    \caption{\textbf{Comparison of \MyModel}~with recent video reasoning models, Video-R1~\cite{videor12025}, VideoChat-R1/1.5~\cite{videochatR12025,yan2025videochat_r1.5}, and VideoRFT~\cite{videorft2025}, across three metrics: 
    \textbf{TAC} (Think--Answer Consistency), \textbf{VAS} (Video Attention Score), and \textbf{Accuracy}. 
    The upper row reports average scores over six reasoning benchmarks, VideoMathQA~\cite{rasheed2025videomathqa}, Video-MMMU~\cite{hu2025video_mmmu}, MMVU~\cite{mmvu2024}, VSIBench~\cite{vsibench}, MINERVA~\cite{minerva2024}, and SciVideoBench~\cite{deng2025scivideobench}, 
    while the lower row shows averages over all 11 benchmarks including the five generic ones, MVBench~\cite{li2023mvbench}, VideoMME~\cite{li2023videomme}, TempCompass~\cite{tempcompass}, MLVU~\cite{zhou2025mlvu}, and LongVideoBench~\cite{wu2024longvideobench}.~\MyModel~performs better across both reasoning and overall evaluations, achieving higher consistency (TAC) and video-focused reasoning (VAS) while maintaining competitive accuracy (Details in Sec.~\ref{sec:experiments}).
}
  \label{fig:teaser}
\vspace{-1.5em}
\end{figure}
% % % % % % % % % % % % %  Intro Figure % % % % % % % % % % % % %

To address these limitations, we propose a reinforcement learning framework that strengthens both temporal precision and reasoning consistency. 
Our method begins with supervised fine-tuning to teach the model timestamp-aware thinking, enabling it to generate intermediate reasoning steps linked to the video timeline. 
We then apply Group Relative Policy Optimization (GRPO)~\cite{shao2024deepseekmath} using a new \textbf{Temporal Alignment Reward (TAR)} that encourages accurate temporal reasoning. 
TAR evaluates the alignment between predicted and reference timestamps and is applied only when the reasoning and the final answer are consistent. This combination allows the model to learn coherent and temporally accurate reasoning traces. We refer to the resulting model as \textbf{\MyModel}, 
the next step in the video reasoning family~\cite{videor12025,videochatR12025,yan2025videochat_r1.5,videorft2025,luo2025thinking,dang2025twgrpo}.

Our approach enhances the capabilities of Video-R1 by improving temporal alignment and reasoning consistency, resulting in higher overall accuracy.
As shown in Fig.~\ref{fig:teaser},~\MyModel~achieves consistent improvements across multiple video reasoning and non-reasoning benchmarks under a unified experimental setup.
All methods are tested using the same number of frames, resolution, and input formatting across 11 established video understanding and reasoning benchmarks~\cite{li2023mvbench,li2023videomme,tempcompass,zhou2025mlvu,wu2024longvideobench,rasheed2025videomathqa,hu2025video_mmmu,mmvu2024,vsibench,minerva2024,deng2025scivideobench}, along with our proposed TAC and VAS metrics.
These results demonstrate that reinforcing temporal alignment and logical coherence leads to more reliable and grounded video reasoning without sacrificing performance.

Our main contributions are summarized as follows.
\begin{itemize}
    \item We introduce two diagnostic metrics for video reasoning: \textbf{Think--Answer Consistency (TAC)} and \textbf{Video Attention Score (VAS)}, which measure logical coherence and perceptual focus in model reasoning. (Sec.~\ref{sec:reasoning_quality_metrics})
    \item We propose a reinforcement learning framework with a new \textbf{Temporal Alignment Reward (TAR)} that promotes temporally precise and self-consistent reasoning. (Sec.~\ref{sec:temporal_alignment_reward})
    \item We curate a \textbf{timestamp-aligned reasoning dataset} by filtering high-difficulty samples using an automated pipeline, followed by manual verification. This dataset provides the foundation for temporal alignment and reasoning supervision. (Sec.~\ref{sec:dataset})
    \item We conduct a comprehensive comparison with existing video reasoning and non-reasoning models under a unified evaluation protocol, showing consistent gains in TAC, VAS and accuracy across multiple video understanding and reasoning benchmarks. (Sec.~\ref{sec:results})
\end{itemize}

%% file: sec/2_reasoning_quality_metrics.tex
\section{Reasoning Quality Metrics}
\label{sec:reasoning_quality_metrics}

Existing video understanding benchmarks primarily measure answer accuracy, which alone does not reflect how a model reaches its predictions~\cite{li2023mvbench,li2023videomme,tempcompass,wu2024longvideobench,zhou2025mlvu,rasheed2025videomathqa,hu2025video_mmmu,mmvu2024,minerva2024,vsibench,deng2025scivideobench,xiao2021nextqa,patraucean2023perception_test,cheng2025v,khattak2025good,wang2025lvbench,mangalam2023egoschema}. As shown in Fig.~\ref{fig:motivation}, a model can produce the correct answer while its reasoning remains inconsistent or unrelated to the video. Such cases reduce interpretability and raise questions about the model's actual reasoning ability. To assess reasoning quality more comprehensively, we introduce two complementary metrics: \textbf{Think--Answer Consistency (TAC)} and the \textbf{Video Attention Score (VAS)}. These metrics quantify distinct aspects of reasoning quality, namely logical coherence and visual dependence, providing a more comprehansive understanding of model behavior.

\noindent
\textbf{Think--Answer Consistency (TAC).}~TAC measures the logical alignment between the model's reasoning and its final answer. A reasoning model usually responds in the format \texttt{<think>}~\textit{reasoning}~\texttt{</think>}~\texttt{<answer>}~\textit{final answer}~\texttt{</answer>}. 
In many cases, the conclusion drawn in the reasoning section contradicts the final answer, suggesting that the model's reasoning is disconnected from its actual decision. 
A high TAC value indicates coherent reasoning, where the model concludes the same output in both reasoning and answer, while a low TAC reflects reasoning inconsistencies that limit interpretability. We compute TAC only for the subset containing correctly answered samples, since consistency of incorrect answers can convey a flase sense of better performance.
For a model $M$ and dataset $D$:
\begin{equation}
\text{TAC}(M, D) = \frac{1}{|D_{\text{correct}}|} 
\sum_{i \in D_{\text{correct}}} 
\mathbb{I}[\hat{a}_i^{\text{think}} = \hat{a}_i^{\text{answer}}],
\label{main:equation_1}
\end{equation}
where $D_{\text{correct}}$ represents samples for which the model's predicted final answer matches the ground truth. 
$\hat{a}_i^{\text{think}}$ is the answer concluded in the reasoning trace (\texttt{<think>...</think>}), 
and $\hat{a}_i^{\text{answer}}$ is the explicit final answer of the model (\texttt{<answer>...</answer>}). 
The operator $\mathbb{I}[\cdot]$ is an indicator function that equals~1 when the condition inside the brackets holds true and~0 otherwise. 
TAC $\in [0,1]$ thus represents the fraction of correct samples where the reasoning and final answer are consistent. 
The value of $(1-\text{TAC})$ indicates the proportion of ``uninterpretable correctness,'' where the model outputs the right answer for unclear or contradictory reasons.

\noindent
\textbf{Video Attention Score (VAS).}~While TAC measures logical coherence, it does not capture whether the reasoning relies on the visual cues in the video. The \textbf{Video Attention Score (VAS)} quantifies how much the reasoning \textit{claims} to rely on visual information rather than on textual patterns or prior knowledge. 
We employ an \textit{LLM-as-a-Judge} framework, where a language model evaluates the reasoning trace based on predefined guidelines and assigns a visual-reliance score between 0 and 10, later normalized to the range $[0,1]$. 
The LLM-judge considers cues such as explicit mentions of visual entities, spatial or temporal relations, and causal statements linked to observed actions. The complete judging prompt is provided in the \texttt{Appendix}~\ref{supp:vas}. 
Importantly, the \textit{token-wise attention distributions} shown in Fig.~\ref{fig:motivation} serve only as a visualization of the internal model behavior and are not used to compute VAS. 
VAS is independent of attention weights and provides a complementary, text-based proxy for visual reasoning. 
We can represent it as:
\vspace{-0.3em}
\begin{equation}
\text{VAS}(M, D) = \frac{1}{|D|} \sum_{i=1}^{|D|} s_i,
\end{equation}
\vspace{-0.1em}
where $s_i \in [0,1]$ is the normalized LLM-judge score assigned to the reasoning trace of sample $i$. Although VAS evaluates \textit{claimed} rather than verified reliance on video evidence, it complements accuracy and TAC by quantifying how visually grounded a model's reasoning appears.

\noindent
\textbf{Complementarity.} TAC and VAS together form a two-axis evaluation of reasoning quality. TAC reflects the model's internal logical coherence, while VAS captures its degree of visual focus. Both metrics correlate with intuitive model behavior and qualitative observations, revealing how well a model reasons rather than merely predicts. In the following section, we use these insights to design a reinforcement learning framework that directly promotes consistent and temporally aligned reasoning through the proposed Temporal Alignment Reward (TAR) with consistency gating.

%% file: sec/3_temporal_alignment_reward.tex
\section{RL with Temporal Alignment Reward}
\label{sec:temporal_alignment_reward}
% % % % % % % % % % % % % % % % % % % % % % % % % % % % % % % 
\begin{figure*}[!t]
\centering
\includegraphics[width=1\linewidth]{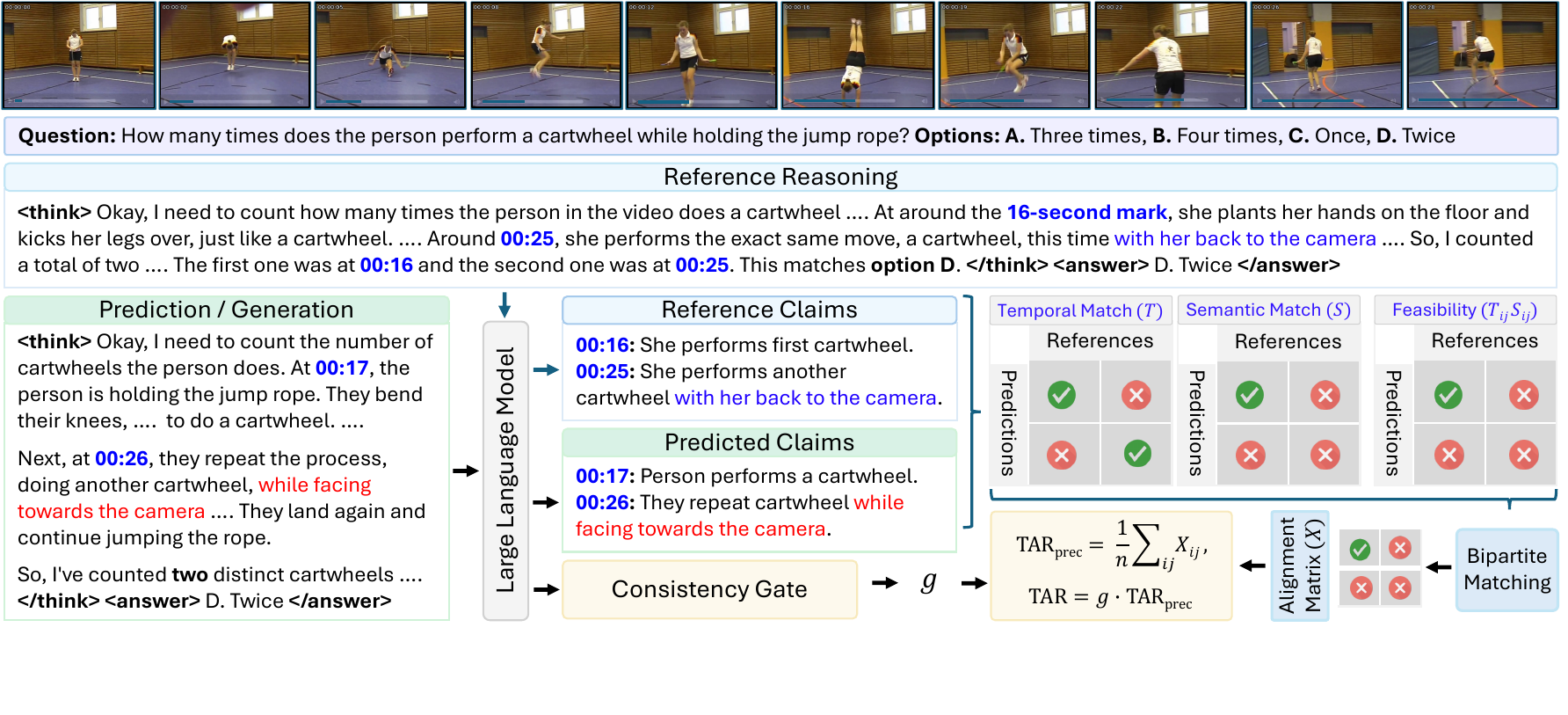}
\vspace{-4.5em}
\caption{
    \textbf{Temporal Alignment Reward (TAR).} 
    The figure illustrates how the proposed TAR is computed for a video reasoning example. 
    We have a \textit{reference reasoning} containing grounded timestamps and a \textit{predicted reasoning} generated by our model. 
    Both are processed by an LLM that extracts timestamp-sentence pairs, referred to as \textit{claims}. 
    The \textbf{Temporal Match} matrix~($T$) checks if a predicted timestamp lies within a temporal tolerance~$\Delta$ of a reference timestamp ({\color{checkgreen}True}~if $|t_i^{\mathrm{pred}} - t_j^{\mathrm{ref}}|\le \Delta$, otherwise~{\color{checkred}False}). 
    The \textbf{Semantic Match} matrix~($S$) compares sentence similarities using embeddings; a {\color{checkgreen}True}/tick indicates that the cosine similarity exceeds a threshold~$\tau$. 
    A one-to-one bipartite matching is then applied over pairs satisfying both conditions to obtain the binary alignment matrix~$X$, ensuring that each predicted claim aligns with at most one reference claim. 
    The precision-based temporal alignment reward is computed as 
    $\mathrm{TAR}_{\text{prec}} = \tfrac{1}{n}\sum_{i,j} X_{ij}$. 
    This score represents the fraction of predicted timestamps that are temporally and semantically valid. 
    Finally, TAR is gated by the consistency indicator~$g$ that verifies if the reasoning and answer are coherent, yielding $\mathrm{TAR} = g \times \mathrm{TAR}_{\text{prec}}$. 
    Rows correspond to predictions and columns to references. 
    The example demonstrates how correct temporal matches contribute to higher rewards under logical consistency.
}
\label{fig:temporal_grounding_reward}
\vspace{-1em}
\end{figure*}
% % % % % % % % % % % % % % % % % % % % % % % % % % % % % % % 
The reasoning quality metrics, TAC and VAS, reveal that models often produce reasoning traces that are only weakly grounded in video content or logically inconsistent with their final answers.~To explicitly encourage temporally precise and consistent reasoning, we follow a two-stage post-training setup consisting of \emph{supervised fine-tuning (SFT)} followed by \emph{reinforcement learning (RL)} using GRPO~\cite{shao2024deepseekmath}. 
Within the RL stage, we introduce a novel reinforcement signal called the \textbf{Temporal Alignment Reward (TAR)}, designed to promote reasoning traces where predicted timestamps are both temporally and semantically aligned with the reference reasoning while remaining logically consistent. 
The formulation of TAR is shown in Fig.~\ref{fig:temporal_grounding_reward}.

\noindent
\textbf{Claim extraction.}~For both the reference and predicted reasoning traces, we first extract timestamped \textit{claims}. Each claim consists of a timestamp $t$ and a short sentence $u$ describing the observed visual event at that time. We denote the sets of reference and predicted claims as:
\begin{equation}
\mathcal{R} = \{(t_j^{\mathrm{ref}}, u_j^{\mathrm{ref}})\}_{j=1}^{m}, \quad 
\mathcal{P} = \{(t_i^{\mathrm{pred}}, u_i^{\mathrm{pred}})\}_{i=1}^{n}.
\end{equation}
Where $m$ and $n$ are the number of reference and predicted claims respectively. These claims are automatically obtained using an LLM that identifies temporal mentions and associated descriptions within the reasoning text.~We refer to the annotated reasoning as \textit{reference reasoning} rather than ground truth, since it may not be fully exhaustive.

\noindent
\textbf{Temporal and semantic matching.}~To measure how well each predicted claim corresponds to the reference reasoning, we define two binary matrices that capture temporal and semantic agreement. The \textit{temporal match matrix} $T \in \{0,1\}^{n \times m}$ is defined as:
\begin{equation}
T_{ij} = 
\begin{cases}
    1, & \text{if } |t_i^{\mathrm{pred}} - t_j^{\mathrm{ref}}| \leq \Delta, \\
    0, & \text{otherwise,}
\end{cases}
\end{equation}
where $\Delta$ is the temporal tolerance. The \textit{semantic match matrix} $S \in \{0,1\}^{n \times m}$ is given by:
\begin{equation}
S_{ij} = 
\begin{cases}
    1, & \text{if } \mathrm{sim}\big(e(u_i^{\mathrm{pred}}), e(u_j^{\mathrm{ref}})\big) \geq \tau, \\
    0, & \text{otherwise,}
\end{cases}
\end{equation}
where $e(\cdot)$ denotes the sentence embedding function and $\mathrm{sim}(\cdot,\cdot)$ denotes the cosine similarity. The parameters $\Delta$ and $\tau$ represent the temporal and semantic thresholds, respectively, selected empirically (details in \texttt{Appendix}~\ref{supp:tar},~\ref{supp:additional_ablations}). A valid match must satisfy both conditions.

\noindent
\textbf{One-to-one assignment.}~After computing the temporal and semantic matches between predicted and reference claims, we enforce a one-to-one alignment to establish correspondence between reasoning events. 
We formulate a bipartite matching problem that maximizes 
valid alignments while satisfying both temporal and semantic match constraints:
\begin{equation}
\max_{X \in \{0,1\}^{n \times m}} \sum_{i,j} X_{ij},
\end{equation}
\begin{equation}
\begin{aligned}
\text{s.t. } & X_{ij} \le T_{ij} S_{ij}, \quad 
\sum_j X_{ij} \le 1, \quad \sum_i X_{ij} \le 1,
\end{aligned}
\end{equation}
where $T_{ij}$ indicates whether the $i$-th predicted timestamp lies within the temporal tolerance~$\Delta$ of the $j$-th reference timestamp, and 
$S_{ij} = \mathbb{I}\{\mathrm{sim}(e(u_i^{\mathrm{pred}}), e(u_j^{\mathrm{ref}})) \ge \tau\}$ denotes semantic similarity above a threshold~$\tau$. 
Among all feasible pairs, we perform a greedy one-to-one matching that selects, for each predicted claim, the most similar unused reference claim above the threshold. 
The resulting binary matrix~$X$ represents the final alignment and ensures that duplicate predictions referring to the same reference event are counted only once when computing the reward.

\noindent
\textbf{Precision-based temporal alignment.}~Given the alignment matrix~$X$, the precision-based 
alignment score is defined as:
\begin{equation}
\mathrm{TAR}_{\text{prec}} = \frac{1}{n} \sum_{ij} X_{ij},
\end{equation}
where $X_{ij} = 1$ if the $i$-th predicted claim is both temporally and semantically matched to the $j$-th reference claim, and $0$ otherwise.
This score measures the fraction of predicted timestamps that are 
temporally and semantically supported
by the reference reasoning. 
We adopt a precision-based formulation because with a sparse reference, recall is an unreliable signal that fails to penalize overgeneration. 
Precision instead provides a stable learning signal by rewarding only verifiable predictions and prioritizing strong support over exhaustive coverage. This precision score serves as the base component of TAR before applying the consistency gate.

\noindent
\textbf{Consistency gating.}~To ensure that temporal precision is only rewarded when reasoning is logically coherent, we apply a consistency gate. 
Let $g = \mathbb{I}\!\left[\text{TAC} = 1\right]$ denote the sample-level consistency indicator, 
where $\mathbb{I}[\cdot]$ is the indicator function that equals~1 if the conclusion from the model’s reasoning matches its predicted final answer 
(i.e., the sample is consistent) and~0 otherwise. 
The gated temporal alignment reward is then computed as:
\begin{equation}
\mathrm{TAR} = g \cdot \mathrm{TAR}_{\text{prec}}.
\end{equation}

\noindent
\textbf{Total reward formulation.}~The total GRPO reward combines the accuracy, format, and the proposed consistency-gated temporal alignment reward. The final formulation is:
\begin{equation}
R_{\text{total}} = 
\lambda_{\text{acc}} R_{\text{acc}} +
\lambda_{\text{fmt}} R_{\text{fmt}} +
\lambda_{\text{tar}} \mathrm{TAR},
\end{equation}
where $R_{\text{acc}}$ equals 1 if the final answer is correct, $R_{\text{fmt}}$ equals 1 if the model output follows the required \texttt{<think>} and \texttt{<answer>} structure, and the $\lambda$ terms are weighting coefficients. This combined objective encourages the model to produce correct, well-structured, and temporally consistent reasoning traces. (\texttt{Appendix}~\ref{supp:implementation_details} for implementation details).

\noindent
\textbf{Summary.}~In summary, the Temporal Alignment Reward measures how well the timestamps in the model’s reasoning align with reference reasoning, while the consistency gate ensures that improvements in timestamp precision contribute to coherent reasoning. Together with accuracy and format rewards, TAR provides a balanced reinforcement signal that enhances temporal precision, logical consistency, and visual focus in video reasoning.

% % % % % % % % % % % % % % % % % % % % % % % % % % % % % % % 

%% file: sec/4_experiments.tex
\section{Experiments}
\label{sec:experiments}

This section presents our dataset, implementation details, and quantitative evaluation of \textbf{\MyModel}.
We describe how the data is collected, filtered, and refined to train timestamp-aware reasoning, followed by implementation and training specifics.
Then, a comprehensive comparison is provided across 11 benchmarks, across generic and reasoning datasets, using TAC, VAS and answer accuracy metrics.

% % % % % % % % % % % % %  TAC/VAS Figure % % % % % % % % % % % % %
\begin{figure*}[!b]
\vspace{-1em}
\centering
    \includegraphics[width=1\linewidth]{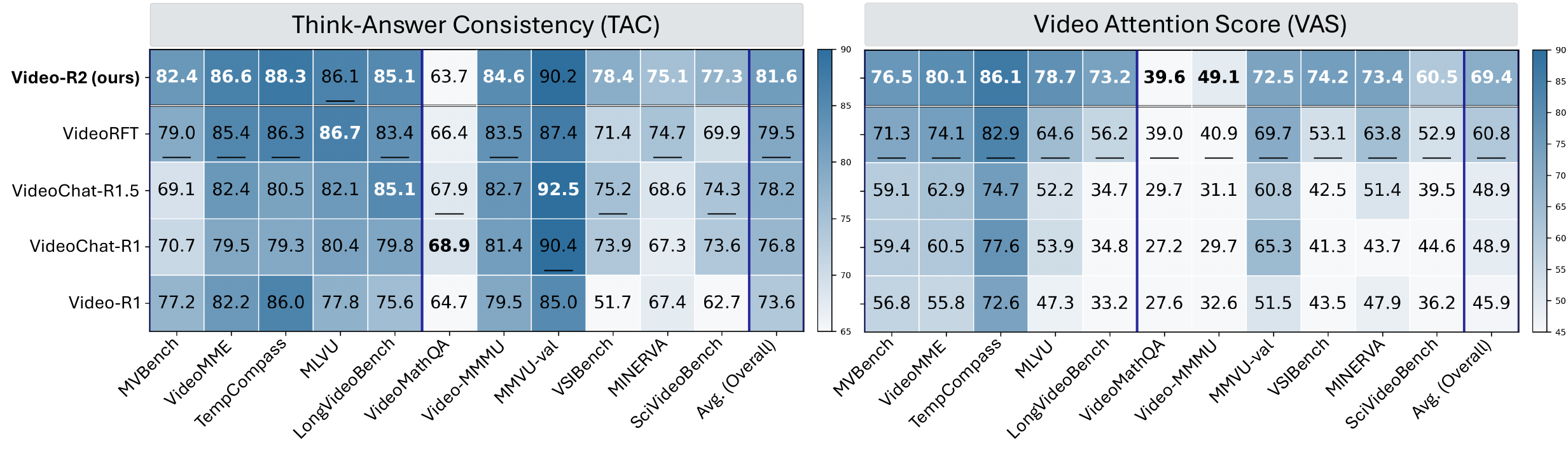}
    \vspace{-2.5em}
    \caption{
    \textbf{Reasoning quality comparison on TAC and VAS metrics.} 
    The left and right heatmaps show \textbf{TAC} and \textbf{VAS} across 11 benchmarks, including five generic~\cite{li2023mvbench,li2023videomme,tempcompass,zhou2025mlvu,wu2024longvideobench} and six reasoning-focused~\cite{rasheed2025videomathqa,hu2025video_mmmu,mmvu2024,vsibench,minerva2024,deng2025scivideobench} datasets. 
    The rightmost columns show the overall average. 
    The {\color{darkblue}\textbf{dark blue}} vertical line separates the generic, reasoning, and average columns. 
    Darker colors indicate higher scores, with the best results in \textbf{bold} and the second-best \underline{underlined} for all benchmarks.
    \MyModel~achieves the best TAC on 8 out of 11 benchmarks and the best VAS on all 11 benchmarks compared with previous reasoning models. 
    Overall, the proposed temporal alignment and consistency gating improve both logical coherence (TAC) and perceptual focus (VAS) in video reasoning multimodal models.
}
    \label{fig:tac_vas_results}
% \vspace{-1em}
\end{figure*}
% % % % % % % % % % % % %  TAC/VAS Figure % % % % % % % % % % % % %

% % % % % % % % % % % % % % % % Dataset % % % % % % % % % % % % % % % %
\subsection{Dataset}
\label{sec:dataset}

\noindent
\textbf{Overview.}~The dataset used to train~\MyModel~is designed to provide timestamp-aware reasoning supervision for both SFT and reinforcement learning. 
The data consists of video–question–answer triplets from multiple public sources, re-ranked and refined to emphasize reasoning-rich and temporally grounded examples. 
In total, the dataset contains 15,271 samples drawn from 11,816 unique videos.

\noindent
\textbf{Data collection.}~We collect video–question–answer pairs from five existing video QA datasets: 
\textbf{LLaVA-Video-178K}~\cite{zhang2024video}, \textbf{NeXtQA}~\cite{xiao2021nextqa}, \textbf{ActivityNet-QA}~\cite{yu2019activitynet}, \textbf{PerceptionTest}~\cite{patraucean2023perception_test}, and \textbf{Clevrer}~\cite{yi2019clevrer}. Together, these sources provide a diverse mixture of short to medium length videos covering everyday, causal, and temporal reasoning scenarios, resulting in approximately 200K QA pairs.

\noindent
\textbf{Difficulty ranking.}~To prioritize challenging examples that demand deeper temporal reasoning, we assign a difficulty score to each QA sample. Following the pipeline in Video-R1~\cite{videor12025}, we first generate intermediate reasoning steps for each QA pair. An LLM then analyzes the reasoning trace associated with each question and estimates how difficult it is to derive the correct answer given the question’s context. Each sample receives a score between 0 and 10, where higher values indicate greater reasoning difficulty. The complete judging rubric and prompt are provided in the \texttt{Appendix}~\ref{supp:dataset_prompts}. We observe that samples with higher difficulty scores often involve multi-step or temporally dependent reasoning, aligning with our goal of enhancing the model’s temporal reasoning ability. We select the 15K most difficult samples for training our model.

\noindent
\textbf{Reasoning regeneration and verification.}~We regenerate reasoning traces for all selected samples using Gemini-2.5-pro~\cite{comanici2025gemini}. For each sample, the model receives both the question and the correct answer along with the video and is tasked with generating intermediate reasoning that explicitly references timestamps and describes relevant spatial and temporal details. Providing the correct answer reduces the likelihood of factual errors, but this setup may still produce occasional fabricated or misaligned reasoning. To ensure quality, we apply a two-stage filtering process: first, we automatically discard samples with inconsistent reasoning (\textit{TAC}~=~0), and second, we conduct human verification on a randomly selected subset of 500 samples. No factual errors or inconsistencies were found in this subset, confirming the overall reliability of the regenerated reasoning. While some minor imperfections may remain since full human verification is impractical at this scale, we observe the data to be of consistently high quality. 
Additional quality analysis and dataset examples are provided in the {\texttt{Appendix}~\ref{supp:dataset_quality},~\ref{supp:dataset_examples}}.

\noindent
\textbf{Dataset statistics.}~The final dataset contains 15,271 QA pairs from 11,816 unique videos, with an average duration of 49 sec. Based on the difficulty ranking, we adopt a 70-30 split, using the top 30\% most difficult samples for GRPO and the remaining 70\% for SFT. This results in 10,467 samples (8,621 unique videos) for SFT and 4,804 samples (4,099 unique videos) for GRPO. Each reasoning trace averages about 450 tokens. (\texttt{Appendix}~\ref{supp:dataset_stats} for more details).
% % % % % % % % % % % % % % % % Dataset % % % % % % % % % % % % % % % %

% % % % % % % % % % % % % % % % Implementation Details % % % % % % % % % % % % % % % %
\subsection{Implementation Details}
\label{sec:implementation_details}
\MyModel~is built upon the non-reasoning Qwen2.5-VL-7B~\cite{bai2025qwen2.5VL} MLLM and is trained following the standard reasoning-tuning pipeline comprising SFT and GRPO, consistent with prior video reasoning works~\cite{videor12025,videochatR12025,yan2025videochat_r1.5,videorft2025}.
%%%%%%%%%%%%%%%%%%%%%%%%%%%%%%%%%
Please refer to \texttt{Appendix}~\ref{supp:implementation_details} for training hyperparameters and additional implementation details.

\noindent
\textbf{Evaluation protocol.}~Previous reasoning models based on Qwen2.5-VL report results under varied configurations (e.g., Video-R1~\cite{videor12025} uses 16/32/64 frames, VideoRFT~\cite{videorft2025} uses 32 frames, and VideoChat-R1.5~\cite{yan2025videochat_r1.5} uses up to 2048 frames), making direct comparison of reported numbers difficult. 
To ensure fairness and reproducibility, we re-evaluate all reasoning models under a unified protocol using the \texttt{lmms-eval}~\cite{zhang2025lmms} framework. 
Specifically, all models are tested with 2\,FPS sampling, a maximum of 128 frames, and a per-frame maximum resolution of 360$\times$420\,px, using the official prompts provided in each paper. 
This standardized setup provides a balance between computation and visual coverage, enabling a fair and comprehensive comparison of reasoning ability and overall performance. 
% % % % % % % % % % % % % % % % Implementation Details % % % % % % % % % % % % % % % %

% % % % % % % % % % % % %  SoTA Comparison Table % % % % % % % % % % % % %
\begin{table*}[!t]
\centering
\setlength{\tabcolsep}{0.75em}
\begin{adjustbox}{max width=\textwidth}
\begin{tabular}{
  l|                               % Model
  *{5}{c}                          % 5 Generic benches
  |>{\columncolor{gray!10}}c|      % Avg Generic (boxed by vertical rules)
  *{6}{c}                          % 6 Reasoning benches
  |>{\columncolor{gray!10}}c|      % Avg Reasoning (boxed)
  >{\columncolor{gray!10}}c        % Avg Overall (boxed; right border too)
}
\toprule
 &
\multicolumn{6}{c|}{\textbf{Generic Benchmarks}} &
\multicolumn{7}{c|}{\textbf{Reasoning Benchmarks}} &
\\
\cmidrule(lr){2-7}\cmidrule(lr){8-14}
\raisebox{-15ex}{\textbf{Model}} &
\rothead{MVBench} &
\rothead{VideoMME} &
\rothead{TempCompass} &
\rothead{MLVU} &
\rothead{LongVideoBench} &
\rothead{\textbf{Avg. (Generic)}} &
\rothead{VideoMathQA} &
\rothead{Video-MMMU} &
\rothead{MMVU-Val} &
\rothead{VSiBench} &
\rothead{MINERVA} &
\rothead{SciVideoBench} &
\rothead{\textbf{Avg. (Reasoning)}} &
\rothead{\textbf{Avg. (Overall)}} \\
\midrule
% ------------------- Data rows -------------------
\multicolumn{15}{l}{\textbf{Non-Reasoning Models}} \\

LLaVA-OV~\cite{liu2024llavaonevision} & 
56.7 & 58.2 & 67.2 & 63.7 & 56.3 &
60.6 &
20.7 & 33.9 & 54.7 & 32.4 & 31.6 & 18.8 &
32.0 &
45.0 \\

InternVL2.5~\cite{chen2024expanding} &
72.0 & 64.2 & 70.0 & 68.9 & 60.0 &
67.0 &
24.3 & 48.6 & 60.8 & 38.9 & 36.6 & 22.8 &
38.7 &
51.6 \\

Qwen2.5-VL~\cite{qwen2025qwen25technicalreport} &
68.1 & 66.1 & 74.2 & 67.9 & 61.3 &
67.5 &
26.7 & 50.3 & 67.2 & 37.7 & 35.3 & 16.4 &
38.9 &
51.9 \\

PerceptionLM~\cite{cho2025PerceptionLM} &
77.1 & 58.3 & 72.7 & 66.2 & 60.9 &
67.0 &
22.1 & 39.9 & 56.5 & 34.3 & 35.6 & 18.9 &
34.5 &
49.3 \\

InternVL3~\cite{zhu2025internvl3} &
75.4 & 66.3 & 72.2 & 71.2 & 58.8 &
68.8 &
29.1 & 50.3 & 66.1 & 42.1 & 34.1 & 29.4 &
41.9 &
54.1 \\

\midrule

\multicolumn{15}{l}{\textbf{Reasoning Models}} \\

Video-R1~\cite{videor12025} &
65.1 & 64.3 & 73.5 & \underline{67.7} & \underline{57.6} &
\underline{65.6} &
23.6 & 47.0 & 64.0 & 37.8 & 33.9 & 26.8 &
38.8 &
51.0 \\

VideoChat-R1~\cite{videochatR12025} &
63.6 & 64.1 & \underline{74.5} & 62.5 & 54.3 &
63.8 &
24.3 & \textbf{52.0} & 64.8 & 33.0 & 33.8 & 26.5 &
39.1 &
50.3 \\

VideoChat-R1.5~\cite{yan2025videochat_r1.5} &
\underline{65.7} & \textbf{64.8} & 73.9 & 65.3 & 53.6 &
64.7 &
\underline{26.4} & 50.0 & \underline{67.0} & 36.1 & 33.7 & 25.8 &
\underline{39.8} &
51.1 \\

VideoRFT~\cite{videorft2025} &
64.8 & 64.1 & 73.8 & 66.6 & 57.0 &
65.3 &
25.2 & 48.1 & 66.7 & 38.5 & 34.0 & 25.7 &
39.7 &
\underline{51.3} \\

\rowcolor{gray!10}
\textbf{\MyModel~(Ours)} &
\textbf{67.5} & 63.8 & \textbf{74.9} & \textbf{68.3} & \textbf{59.2} &
\textbf{66.7} &
\textbf{28.8} & \underline{50.8} & \textbf{67.4} & \textbf{39.4 }& \underline{34.9} & \underline{28.4} &
\textbf{41.6} &
\textbf{53.0} \\

% --- Add more rows below; keep the same column structure ---
% ModelName &  &  &  &  &  &  &  &  &  &  &  &  &  &  \\
\bottomrule
\end{tabular}
\end{adjustbox}
\vspace{-0.5em}
\caption{
    \textbf{Accuracy comparison on generic and reasoning benchmarks.} 
    Results are reported for both non-reasoning and reasoning models (7B/8B scale) across five generic~\cite{li2023mvbench,li2023videomme,tempcompass,zhou2025mlvu,wu2024longvideobench}
    and six reasoning-focused~\cite{rasheed2025videomathqa,hu2025video_mmmu,mmvu2024,vsibench,minerva2024,deng2025scivideobench} benchmarks. 
    Columns list per-benchmark accuracy along with averages for generic, reasoning, and overall performance, with the best results shown in \textbf{bold} and the second-best results \underline{underlined} in each column.
    Our \textbf{\MyModel}~achieves the highest averages across generic (66.7), reasoning (41.6), and overall (53.0), showing improved reasoning ability and balanced accuracy across multiple benchmarks. 
}
\label{tab:sota_table}
\vspace{-1em}
\end{table*}
% % % % % % % % % % % % %  SoTA Comparison Table % % % % % % % % % % % % %

% % % % % % % % % % % % % % % % Results % % % % % % % % % % % % % % % %
\subsection{Results}
\label{sec:results}

We evaluate~\MyModel~across 11 benchmarks, including five generic and six reasoning datasets, using three complementary metrics:~Think--Answer Consistency~\textbf{(TAC)}, Video Attention Score~\textbf{ (VAS)}, and final answer \textbf{Accuracy}. 

\noindent
\textbf{Reasoning quality (TAC).}~Fig.~\ref{fig:tac_vas_results} (left) shows TAC results across all benchmarks.~\MyModel~ranks first on eight of the 11 benchmarks and achieves the highest overall TAC average, demonstrating logical alignment between the reasoning of the model and its final answers.
Among recent reasoning models, VideoRFT~\cite{videorft2025} achieves the second-best TAC on average, which aligns with its use of semantic consistency reward that explicitly enforces alignment between the video description and SigLIP~\cite{zhai2023sigmoid} visual embeddings. 
The gains of~\MyModel~are attributed to the proposed temporal alignment reward (TAR), which encourages timestamp-accurate reasoning and filters temporally precise but logically inconsistent traces through consistency gating (ablation in Sec.~\ref{sec:ablations}). 
Overall, these results indicate that~\MyModel~produces reasoning with stronger logical coherence between the \texttt{<think>} and \texttt{<answer>} content.

\noindent
\textbf{Reasoning quality (VAS).}~Fig.~\ref{fig:tac_vas_results} (right) shows VAS results across all benchmarks. 
\MyModel~shows clear improvement over previously established reasoning models~\cite{videor12025,videochatR12025,yan2025videochat_r1.5,videorft2025}, and ranks first across all 11 benchmarks.
Compared with VideoRFT~\cite{videorft2025}, which employs a semantic consistency reward,~\MyModel~achieves an average VAS improvement of~8.6\%, indicating that the proposed Temporal Alignment Reward (TAR) effectively enhances visual correspondence by reinforcing timestamp-referenced reasoning, leading to consistently higher VAS scores across benchmarks.

\noindent
\textbf{Benchmark accuracy.}~Tab.~\ref{tab:sota_table} presents results on 11 benchmarks and shows that~\MyModel~achieves the best overall average in all three categories including generic, reasoning and combined.
For generic benchmarks,~\MyModel~reaches an average accuracy of~66.7, higher than the previous best of~65.6~from Video-R1~\cite{videor12025}. 
For reasoning-focused benchmarks, the gap is even clear, where~\MyModel~achieves~41.6 compared to~39.8 from VideoChat-R1.5~\cite{yan2025videochat_r1.5} and~39.7 from VideoRFT~\cite{videorft2025}.
Overall,~\MyModel~achieves an average of~53.0, higher than the second best of~51.3 achieved by 
VideoRFT. 
These results demonstrate that the proposed temporal alignment and consistency gating contribute to both stronger reasoning capability and robust predictive performance under a unified evaluation setup.

\noindent
\textbf{Comparison with non-reasoning models.}~For completeness, Tab.~\ref{tab:sota_table} also reports results from recent non-reasoning MLLMs, including LLaVA-OneVision~\cite{liu2024llavaonevision}, InternVL2.5~\cite{chen2024expanding}, Qwen2.5-VL~\cite{bai2025qwen2.5VL}, PerceptionLM~\cite{cho2025PerceptionLM}, and InternVL3~\cite{zhu2025internvl3}.
These models achieve strong performance on generic benchmarks, reflecting their large-scale pretraining and broad visual coverage. 
For instance, InternVL2.5, Qwen2.5-VL, and PerceptionLM obtain average scores of 67.0, 67.5, and 67.0 on generic benchmarks, respectively, whereas~\MyModel~achieves a comparable 66.7. 
However, on reasoning-focused benchmarks the gap becomes clear: InternVL2.5, Qwen2.5-VL and PerceptionLM average 38.7, 38.9 and 34.5, while~\MyModel~achieves 41.6, indicating a clear improvement in reasoning ability through temporal alignment and consistency training. 
We also include recent InternVL3~\cite{zhu2025internvl3} for completeness, which achieves slightly higher accuracy (41.9 on reasoning benchmarks vs 41.6 of ours) due to extended pretraining on larger datasets.
Notably, \textbf{\MyModel}~is trained on the non-reasoning base Qwen2.5-VL and achieves a higher overall average of 53.0 compared to 51.9, demonstrating that the proposed reasoning optimization effectively enhances reasoning performance while maintaining general visual understanding.

\noindent
\textbf{Summary.}~Overall,~\MyModel~achieves consistent improvements across all three evaluation metrics, reasoning coherence (TAC), visual correspondence (VAS), and accuracy, on a wide range of benchmarks. 
These findings confirm that explicitly rewarding temporal precision and logical consistency leads to more reliable and interpretable video reasoning. 
A detailed comparison of the \textit{training data} used across prior methods versus ours
and an extended technical discussion highlighting \textit{key differences} between~\MyModel~and prior approaches are provided in the \texttt{Appendix}~\ref{supp:comparison_with_prior_methods} and~\ref{supp:novelty}.

% % % % % % % % % % % % % % % % Resuls % % % % % % % % % % % % % % % %

% % % % % % % % % % % % %  SFT/RL/TAR Ablations (Extended Metrics Style) % % % % % % % % % % % % %
\begin{table*}[!t]
\centering
\setlength{\tabcolsep}{0.6em}
\begin{adjustbox}{max width=\textwidth}
\begin{tabular}{
  l|cccc|*{3}{c}|*{3}{c}|*{3}{c}
}
\toprule
& \multicolumn{4}{c|}{\textbf{Components}} &
\multicolumn{3}{c|}{\textbf{Accuracy} $\uparrow$} &
\multicolumn{3}{c|}{\textbf{TAC} $\uparrow$} &
\multicolumn{3}{c}{\textbf{VAS} $\uparrow$} \\
\cmidrule(lr){2-5}\cmidrule(lr){6-8}\cmidrule(lr){9-11}\cmidrule(lr){12-14}
\raisebox{-7ex}{\textbf{Method}} &
\rotheadablation{\textbf{SFT}} &
\rotheadablation{\textbf{RL}} &
\rotheadablation{\textbf{TAR}} &
\rotheadablation{\textbf{Cons. ($g$)}} &
\rotheadablation{\textbf{Generic}} &
\rotheadablation{\textbf{Reasoning}} &
\rotheadablation{\textbf{Overall}} &
\rotheadablation{\textbf{Generic}} &
\rotheadablation{\textbf{Reasoning}} &
\rotheadablation{\textbf{Overall}} &
\rotheadablation{\textbf{Generic}} &
\rotheadablation{\textbf{Reasoning}} &
\rotheadablation{\textbf{Overall}} \\
\midrule

Qwen2.5-VL (non-reasoning) &
- & - & - & - &
67.5 & 38.9 & 51.9 &
- & - & - &
- & - & -  \\

\midrule

Qwen2.5-VL (CoT) &
\xmark & \xmark & \xmark & \xmark &
59.6 & 37.0 & 47.3 &
63.2 & 49.1 & 55.5 &
53.2 & 36.9 & 44.3 \\

\textbf{+ SFT} &
\cellcolor{gray!10}\cmark & \xmark & \xmark & \xmark &
60.9 & 38.8 & 48.9 &
\textbf{96.5} & \textbf{97.1} & \textbf{96.8} &
76.8 & \textbf{63.7} & 69.7 \\

\textbf{+ SFT + GRPO} &
\cellcolor{gray!10}\cmark & \cellcolor{gray!10}\cmark & \xmark & \xmark &
65.9 & 41.0 & 52.3 &
82.9 & 71.6 & 76.8 &
75.8 & 61.5 & 61.5 \\

\textbf{+ SFT + GRPO + TAR} &
\cellcolor{gray!10}\cmark & \cellcolor{gray!10}\cmark & \cellcolor{gray!10}\cmark & \xmark &
\underline{66.5} & \underline{41.2} & \underline{52.7} &
83.0 & 72.9 & 77.5 &
\underline{78.4} & \underline{62.9} & \textbf{69.9 }\\

\rowcolor{gray!10}
\textbf{+ SFT + GRPO + TAR w/ Cons. ($g$)} &
\cellcolor{gray!10}\cmark & \cellcolor{gray!10}\cmark & \cellcolor{gray!10}\cmark & \cellcolor{gray!10}\cmark &
\textbf{66.7} & \textbf{41.6} & \textbf{53.0} &
\underline{85.7} & \underline{78.2} & \underline{81.6} &
\textbf{78.9} & 61.5 & 69.4\\

\bottomrule
\end{tabular}
\end{adjustbox}
\vspace{-0.5em}
\caption{
\textbf{Ablation on key components of \MyModel.}
Checkmarks denote enabled modules: supervised fine-tuning (SFT), group relative policy optimization (GRPO), Temporal Alignment Reward (TAR), and consistency gating ($g$). 
Metrics are reported as average scores across five generic~\cite{li2023mvbench,li2023videomme,tempcompass,zhou2025mlvu,wu2024longvideobench}, six reasoning-focused~\cite{rasheed2025videomathqa,hu2025video_mmmu,mmvu2024,vsibench,minerva2024,deng2025scivideobench}, and all benchmarks for \textbf{Accuracy}, \textbf{TAC}, and \textbf{VAS}. 
The results highlight the progressive improvements achieved by introducing temporal alignment and consistency gating in~\MyModel.
}
\label{tab:ablation_components}
\vspace{-1.3em}
\end{table*}
% % % % % % % % % % % % %  SFT/RL/TAR Ablations (Extended Metrics Style) % % % % % % % % % % % % %

% % % % % % % % % % % % %  Ablations (Extended Metrics Style) % % % % % % % % % % % % %
% % % % % % % % % % % % %  Ablations (Extended Metrics Style) % % % % % % % % % % % % %
\subsection{Ablations}
\label{sec:ablations}

To analyze the contribution of each training component, we conduct a step-by-step ablation over SFT, GRPO, and the proposed temporal alignment reward (TAR) w/ and w/o consistency gate ($g$). 
The results in Tab.~\ref{tab:ablation_components} report averages for Accuracy, TAC, and VAS across all 11 benchmarks. 
The top row corresponds to the non-reasoning Qwen2.5-VL~\cite{bai2025qwen2.5VL} baseline that directly predicts the final answer without intermediate reasoning; therefore, TAC and VAS cannot be computed. 
The second row, Qwen2.5-VL (CoT), shows the same model evaluated with chain-of-thought prompting, where it is asked to \texttt{think} before \texttt{answer}. We note that CoT decreases overall accuracy and yields the lowest TAC, suggesting that the non-reasoning model struggles to maintain coherence when forced to generate reasoning traces.

\noindent
\textbf{Effect of SFT.}~SFT over the non-reasoning baseline significantly improves both TAC and VAS, with TAC increasing from 55.5 to 96.8 and VAS from 44.3 to 69.7. 
This demonstrates that SFT effectively teaches the model to produce coherent, timestamp-aware reasoning in the \texttt{<think>}--\texttt{<answer>} format. 
Accuracy also improves slightly over the CoT baseline, confirming that reasoning supervision benefits both interpretability and predictive performance.

\noindent
\textbf{Effect of GRPO.}~Adding GRPO further improves accuracy (48.9~$\rightarrow$~52.3) but leads to a decrease in both TAC (96.8~$\rightarrow$~76.8) and VAS (69.7~$\rightarrow$~61.5). 
This behavior arises because the GRPO reward is only applied to the final answer; as a result, the model tends to rely on simpler \textit{answer-based shortcuts} similar to those used by the non-reasoning baseline. 
Although this improves answer correctness, it reduces reasoning alignment and visual correspondence, highlighting the need for an additional reward component that explicitly targets reasoning quality.

\noindent
\textbf{Effect of TAR.}~Incorporating TAR improves both reasoning quality metrics: TAC improves from 76.8 to 77.5 overall, and VAS shows a notable increase from 61.5 to 69.9. 
These gains directly result from rewarding timestamp-verified reasoning claims, which encourage the model to rely more on visual and temporal evidence when reasoning. 

\noindent
\textbf{Effect of consistency gating ($g$).}~Finally, enabling the consistency gate produces the overall best results, improving accuracy from 52.7 to 53.0, reasoning TAC jumps from 72.9 to 78.2, and overall TAC from 77.5 to 81.6. 
The gating ensures that temporally precise reasoning is rewarded only when the \texttt{<think>} conclusion aligns with the final \texttt{<answer>}, thus stabilizing training and reinforcing coherent reasoning. 
Additional ablations are in~\texttt{Appendix}~\ref{supp:additional_ablations}, including (i) the effect of introducing a recall-based term in TAR, (ii) the stability of the proposed VAS metric using different LLM judges, (iii) sensitivity analyses for the semantic similarity ($\tau$) and temporal thresholds ($\Delta$).

% Discussion Point
\noindent
\textbf{The SFT consistency paradox.}~An interesting trend in Tab.~\ref{tab:ablation_components} is the high TAC observed after SFT, where the model achieves near-perfect consistency but relatively low accuracy. 
When trained with cross-entropy next-token prediction loss, the model learns to generate well-structured and logically coherent reasoning traces that, however, do not always lead to correct final answers. 
In contrast, GRPO increases answer accuracy by explicitly rewarding generations with correct answer but lowers reasoning consistency, indicating that \textit{policy optimization tends to exploit answer-level shortcuts rather than improve actual reasoning behavior}. 
The introduction of the Temporal Alignment Reward (TAR) and the consistency gate ($g$) mitigates this trade-off, yielding a relatively balanced model that maintains coherence while improving accuracy. 
We further study this effect in \texttt{Appendix}~\ref{supp:additional_ablations_stronger_consistency_reward}, showing that stronger consistency-oriented reward can close the TAC gap between SFT and GRPO but reduce accuracy, underscoring an open trade-off between interpretability and performance. \textbf{Video-R2} is a significant step toward addressing this challenge,
though further research is required to fully resolve this balance.
% 

% % % % % % % % % % % % %  Ablations (Extended Metrics Style) % % % % % % % % % % % % %
% % % % % % % % % % % % %  Ablations (Extended Metrics Style) % % % % % % % % % % % % %

%% file: sec/6_conclusion.tex
\section{Conclusion}
\label{sec:conclusion}
Current video reasoning models often produce reasoning traces that are not consistent with their answers or grounded in visual content. We identify this gap and introduce two diagnostic metrics; Think--Answer Consistency (\textbf{TAC}) and Video Attention Score (\textbf{VAS}), to quantify these aspects. To address this gap, we introduce a reinforcement learning reward named Temporal Alignment Reward (\textbf{TAR}) with a consistency gate, which encourages temporally precise and logically consistent reasoning. The resulting \textbf{\MyModel} model improves reasoning quality and accuracy across multiple benchmarks, showing that explicit temporal alignment and consistency rewards lead to more reliable video reasoning. Our codes, data and model will be open-sourced.

\section{Acknowledgement}
Muhammad Maaz is supported by the 2025 Google PhD Fellowship in Machine Perception. The computations were enabled by resources provided by the LUMI supercomputer hosted by CSC (Finland) and the LUMI consortium.

%% file: supplemental/suppl.tex
\renewcommand{\thefigure}{A\arabic{figure}}
\setcounter{figure}{0}
\renewcommand{\thetable}{A\arabic{table}}
\setcounter{table}{0}
\maketitlesupplementary

% % % % % % % % % TOC % % % % % % % % % 
\part{} 
\parttoc
% % % % % % % % % TOC % % % % % % % % % 

%%%%%%%%%%%%%%%%%% Related Works %%%%%%%%%%%%%%%%%%
\input{supplemental/related_work}
%%%%%%%%%%%%%%%%%% Implementation Details %%%%%%%%%%%%%%%%%%
\input{supplemental/implementation_details}
%%%%%%%%%%%%%%%%%% TAC %%%%%%%%%%%%%%%%%%
\input{supplemental/TAC}
%%%%%%%%%%%%%%%%%% VAS %%%%%%%%%%%%%%%%%%
\input{supplemental/VAS}
%%%%%%%%%%%%%%%%%% TAR %%%%%%%%%%%%%%%%%%
\input{supplemental/TAR}
%%%%%%%%%%%%%%%%%% DATASET %%%%%%%%%%%%%%%%%%
\input{supplemental/dataset}
%%%%%%%%%%%%%%%%%% Comparison %%%%%%%%%%%%%%%%%%
\input{supplemental/comparison_with_prior_methods}
%%%%%%%%%%%%%%%%%% Novelty %%%%%%%%%%%%%%%%%%
\input{supplemental/novelty}
%%%%%%%%%%%%%%%%%% Additional Ablations %%%%%%%%%%%%%%%%%%
\input{supplemental/additonal_ablations}
%%%%%%%%%%%%%%%%%% Train/Eval Prompts %%%%%%%%%%%%%%%%%%
\input{supplemental/train_eval_prompts}

%% file: supplemental/related_work.tex
\section{Related Works}
\label{supp:extended_related_work}
\subsection{Large Language Models}

The trajectory of modern AI has been shaped by the rapid evolution of Large Language Models (LLMs), beginning with the Transformer architecture~\cite{vaswani2017attention}. This innovation enabled autoregressive models such as GPT-1~\cite{radford2018improving}, GPT-2~\cite{radford2019language}, and the paradigm-shifting GPT-3~\cite{brown2020language}, which showcased powerful in-context learning. Complementary masked-language models like BERT~\cite{devlin2018bert} advanced language understanding tasks, while T5~\cite{raffel2019exploring} unified multiple NLP tasks under a text-to-text formulation. The scaling era then followed, with PaLM~\cite{chowdhery2022palm} and Chinchilla~\cite{hoffmann2022training} investigating the limits of compute-optimal scaling, and models like Gopher~\cite{rae2021scaling} pushing the envelope of dataset curation.  

\noindent
\textbf{Open weights efforts.} The open-weights movement accelerated progress by making strong LLMs broadly available. Meta's LLaMA series~\cite{touvron2023llama, touvron2023llama2, meta2024llama3} demonstrated that high performance is possible with carefully curated open data. 
% This line is expected to extend to LLaMA-4. 
Other competitive families soon followed: Mistral and its Mixture-of-Experts variant Mixtral~\cite{jiang2023mistral, jiang2024mixtral}, Falcon~\cite{almazrouei2023falcon}, Orca~\cite{mukherjee2023orca}, the Qwen series (Qwen / Qwen1.5~\cite{bai2023qwen}, Qwen2~\cite{qwen2024qwen2}, Qwen2.5~\cite{qwen2025qwen25technicalreport} and Qwen3~\cite{yang2025qwen3}), DeepSeek series (DeepSeek LLM~\cite{deepseek2024llm}, Deepseek-MoE~\cite{dai2024deepseekmoe}, Deepseek-V2~\cite{deepseek2024deepseekv2}, and Deepseek-V3~\cite{liu2024deepseekv3}), Microsoft’s Phi series (Phi-1~\cite{gunasekar2023textbooksphi1}, Phi-1.5~\cite{li2023textbooksphi1.5}, Phi-3~\cite{abdin2024phi3}, and Phi-4~\cite{abdin2024phi4}), and Google’s Gemma family~\cite{gemma2024}. Together, these models have created a vibrant ecosystem of both proprietary and open alternatives to GPT, Gemini and Claude.  

\noindent
\textbf{Reasoning in LLMs.} Reasoning has emerged as a central focus of LLM development. Prompting-based strategies like Chain-of-Thought (CoT) prompting~\cite{wei2022chain}, Self-Consistency~\cite{wang2022self}, and Tree-of-Thoughts (ToT)~\cite{yao2023tree} enable models to produce intermediate reasoning steps. More interactive frameworks, such as ReAct~\cite{yao2022react} and Reflexion~\cite{shinn2023reflexion}, extend reasoning with tool use and self-reflection. Recently, the ``O1/R1'' family of reasoning-first models has gained prominence: DeepSeek-R1~\cite{deepseek2025r1} demonstrated that reinforcement learning with outcome and format rewards can substantially improve mathematical reasoning; Qwen-3~\cite{yang2025qwen3} introduced explicit reasoning-tuned variants; and OpenAI’s o1/o3/o4~\cite{jaech2024openai, openai2025_o3_o4mini} represents proprietary advances in this space. Some relevant open-source efforts include Vision-R1~\cite{visionr12025}, VLM-R1~\cite{vlmr12025}, and Video-R1~\cite{videor12025}, which adapt reasoning-first reinforcement learning to multimodal and temporal settings.

\subsection{Multimodal Large Language Models}

\textbf{Early vision-language models:}  
The connection between vision and language started with dual-stream models such as ViLBERT~\cite{lu2019vilbert} and LXMERT~\cite{tan2019lxmert}, where images and text were processed separately before being combined. Single-stream models like VisualBERT~\cite{li2019visualbert} and UNITER~\cite{chen2019uniter} instead fused the two modalities in a shared space. Later, contrastive learning methods became dominant. CLIP~\cite{radford2021clip} and ALIGN~\cite{jia2021align} showed that training on large collections of image-text pairs can give strong zero-shot recognition abilities. MetaCLIP~\cite{xu2023metaclip} followed the same contrastive objective as CLIP but focused on transparent and reproducible data curation, making it more open and reliable. SigLIP~\cite{zhai2023sigmoid} and SigLIP-2~\cite{siglip2} further improved contrastive training by refining the learning objectives and scaling the training data. More recently, the Perception Encoder~\cite{bolya2025PerceptionEncoder} introduced a scalable family of vision encoders that achieve state-of-the-art performance across image, video, vision-language, and dense prediction tasks.  

\noindent
\textbf{Image-focused MLLMs.}
The landscape of image-focused Multimoddal Large Language Models (MLLMs) is defined by both powerful proprietary systems, such as OpenAI's GPT-4V~\cite{openai2023gptv}, and a rapidly growing ecosystem of open models. Foundational open-source work established two main ways to connect pre-trained vision encoders to LLMs. The first uses adapter modules, such as the Q-Former in BLIP-2~\cite{li2023blip2} and MiniGPT-4~\cite{zhu2023minigpt4}, which serve as a translator between visual features and the LLM. The second uses a much simpler projection approach, where the visual features are mapped directly into the LLM’s embedding space with just a linear layer or a small neural layer, as seen in LLaVA~\cite{liu2023llava}. These designs significantly lowered the barrier for MLLM creation and inspired a wave of successors such as LLaVA-v1.5~\cite{llavav1.5}, LLaVA-NeXT Interleave~\cite{li2024llavanextinterleaved}, and InstructBLIP~\cite{dai2023instructblip}.

Building on these strong foundations, a number of strong open model families have emerged, pushing the state of the art through scaling and architectural refinement. The Qwen-VL~\cite{wang2024qwen2VL, bai2025qwen2.5VL}, InternVL~\cite{chen2024internvl, chen2024far, gao2024mini, wang2024mpo, chen2024expanding, zhu2025internvl3}, and LLaMA 3.2 Vision~\cite{meta2024llama3} and Phi-Vision~\cite{abdin2024phi3, abouelenin2025phi} series represent large-scale efforts that achieve top results on broad vision-language benchmarks. Other notable efforts include Molmo~\cite{deitke2025molmo}, and Pixtral~\cite{pixtral2024} which also deliver highly competitive performance.

Alongside scaling, research has focused on enhancing specific capabilities. To improve efficiency and accessibility, models like MiniCPM-V~\cite{yao2024minicpm} have been developed with smaller parameter counts. For handling high-resolution imagery and complex document understanding, models such as Idefics2~\cite{laurenccon2024matters} and InternLM-XComposer2-4KHD~\cite{dong2024internlm} have introduced specialized techniques. CogVLM~\cite{wang2023cogvlm} explored deeper fusion strategies between the vision and language backbones. For fine-grained understanding that connects language to specific image regions, several models have introduced grounding capabilities. The Kosmos family~\cite{kosmos1,kosmos2} enabled referring to objects via bounding boxes, a concept extended by Ferret 1/2~\cite{you2023ferret, you2023ferret2} to ground any-shape regions and by GLaMM~\cite{hanoona2023GLaMM} to provide pixel-level segmentation. Together, these systems highlight the rapid and diverse growth of image-focused MLLMs.

\noindent
\textbf{Video-focused MLLMs.}  
Extending to video, early works such as Video-ChatGPT~\cite{maaz2023videochatgpt}, Video-LLaMA~\cite{videollama}, and Video-LLaVA~\cite{lin2023videollava} adapted image-based models to temporal sequences.~Specialized models then appeared, including TimeChat~\cite{li2023timechat}, MovieChat~\cite{song2023moviechat}, LLaMA-VID~\cite{li2023llamavid}, VideoChat2~\cite{li2023mvbench}, ChatUniVi~\cite{jin2024chatunivi}, and VTimeLLM~\cite{huang2023vtimellm}.~More~recent~contributions~focus~on~efficiency~and~broader~coverage:~BT-Adapter~\cite{bt_adapter}, ST-LLM~\cite{stllm}, IG-VLM~\cite{igvlm2024}, GoldFish~\cite{goldfish2024}, VideoGPT+~\cite{maaz2024videogptplus}, and Mobile-VideoGPT~\cite{mobilevideogpt2024}.~Methods such as LongVLM~\cite{longvlm2024}, FTFV-LVLM~\cite{chen2024ftfvlvlm}, and LLaVA-MR~\cite{zhang2024llavamr} address token reduction, long-context reasoning, and efficient retrieval, which are critical for scaling video-based multimodal systems.  

\noindent
\textbf{Joint image-video MLLMs.}  
A recent trend is to design models that handle both images and videos within a single framework. LLaVA-OneVision~\cite{liu2024llavaonevision} shows that a unified architecture can support single-image, multi-image, and videos together. The Qwen-VL series~\cite{wang2024qwen2VL,bai2025qwen2.5VL} extends this idea with stronger dynamic resolution and long-context abilities, while the InternVL family~\cite{chen2024expanding,zhu2025internvl3} achieves state-of-the-art results in both document and video understanding. Perception Language Model (PLM)~\cite{bolya2025PerceptionEncoder} builds on the Perception Encoder~\cite{bolya2025PerceptionEncoder} to align powerful vision encoders with language models, enabling competitive performance across image and video benchmarks. These works suggest a shift toward general-purpose multimodal assistants that can seamlessly move between static and temporal visual inputs.  

\noindent
\textbf{Reasoning in MLLMs.} Beyond recognition and understanding, recent works aim to equip MLLMs with explicit reasoning. For images, LLaVA-O1~\cite{xu2024llavao1}, LLaMA-V-o1~\cite{thawakar2025llamavo1}, Vision-R1~\cite{visionr12025}, VLM-R1~\cite{vlmr12025}, Visionary-R1~\cite{visionaryr12025}, and Perception-R1~\cite{yu2025perceptionr1} introduce reasoning-first training strategies. For videos, methods include Video-R1~\cite{videor12025}, VideoChat-R1~\cite{videochatR12025}, VideoChat-R1.5~\cite{yan2025videochat_r1.5}, VideoRFT~\cite{videorft2025}, TW-GRPO~\cite{dang2025twgrpo}, ReVisual-R1~\cite{chen2025revisualr1}, VideoAgent~\cite{wang2024videoagent}, VersaVid-R1~\cite{chen2025versavidr1}, and Vad-R1~\cite{huang2025vadr1}, each adapting reinforcement or process-level rewards to temporal reasoning. A complementary line explores ``visual chain-of-thought'' reasoning, such as AURORA~\cite{bigverdi2025perception}, which introduces intermediate perception tokens (depth maps, segmentation masks) to bridge perception and symbolic reasoning.  

\noindent
\textbf{Benchmarks for Multimodal Reasoning.}~The advancement of multimodal reasoning has been enabled by increasingly diverse benchmarks. For images, MM-Bench~\cite{liu2023mmbench}, MMMU~\cite{yue2023mmu}, AI2D~\cite{kembhavi2016ai2d}, and MathVista~\cite{lu2023mathvista} target visual knowledge, diagram, and math reasoning. For videos, general-purpose benchmarks such as MVBench~\cite{li2023mvbench}, Video-MME~\cite{li2023videomme}, TempCompass~\cite{tempcompass}, MLVU~\cite{zhou2025mlvu}, and LongVideoBench~\cite{wu2024longvideobench} assess broad multimodal understanding, while reasoning-focused datasets including VideoMathQA~\cite{rasheed2025videomathqa}, Video-MMMU~\cite{hu2025video_mmmu}, MMVU~\cite{mmvu2024}, VSIBench~\cite{vsibench}, MINERVA~\cite{minerva2024}, and SciVideoBench~\cite{deng2025scivideobench} evaluate spatio-temporal reasoning.~Together, these benchmarks highlight both the steady gains and the persistent gap to human-level multimodal reasoning.

%% file: supplemental/implementation_details.tex
\section{Implementation Details}
\label{supp:implementation_details}

\subsection{Model and Training Stages}
\MyModel~is initialized from the non-reasoning backbone Qwen2.5-VL-7B~\cite{bai2025qwen2.5VL}. Following prior reasoning-based approaches such as Video-R1~\cite{videor12025}, VideoChat-R1~\cite{videochatR12025}, and VideoRFT~\cite{videorft2025}, training follows two stages: 
\begin{enumerate}
    \item \textbf{Supervised Fine-Tuning (SFT)} to learn timestamp-aware reasoning in the \texttt{<think>} and \texttt{<answer>} format.
    \item \textbf{Group Relative Policy Optimization (GRPO)} to refine temporal alignment and logical consistency through the proposed Temporal Alignment Reward (TAR, Sec.~\ref{sec:temporal_alignment_reward}).
\end{enumerate}

\noindent
\textbf{Video Processing}
Videos are uniformly sampled at 2\,FPS. 
A maximum of 128 frames are used during SFT, and 32 frames during GRPO. Evaluation uses a maximum of 128 frames for all the reasoning models.
Frames are resized to a maximum of 360$\times$420\,px resolution during both training and evaluation. 
For videos requiring audio cues, subtitles are generated via WhisperX-Large-v3~\cite{radford2022whisper} and rendered at the bottom of frames during training. 
To facilitate temporal reasoning, we overlay frames with timestamps (\texttt{HH:MM:SS} format) at the top-left corner during training. Timestamp visibility is dynamically adjusted (light-on-dark or dark-on-light) for optimal legibility. During evaluation, only timestamps (displayed at the top-left corner) are preserved, while subtitles are omitted to ensure that reasoning relies solely on visual and temporal cues.

\noindent
\textbf{Supervised Fine-Tuning (SFT).}~SFT is performed for one epoch using \texttt{LoRA}~\cite{hu2022lora} adapters (rank $r$=64) applied to both the attention and the feed-forward layers. Training uses a global batch size of 32 and learning rate of $1e^{-5}$, with bf16 precision and a maximum context length of 32K tokens. LoRA-based adaptation preserves the pretrained multimodal knowledge of the base Qwen2.5-VL model while enabling efficient reasoning specialization on the relatively small SFT dataset (we use only $\simeq$10K samples for SFT).

\noindent
\textbf{Reinforcement learning (GRPO).}~Followed by SFT, reinforcement learning using GRPO is conducted for one epoch with group size $K$=8 (eight candidate generations per sample), global batch size of 8, learning rate $1e^{-6}$,~and KL anchor coefficient $\beta$=0.04 for training stability.~All reward components including accuracy reward, format reward, and the proposed consistency-gated temporal alignment reward (TAR), are equally weighted (1:1:1) as described in Sec.~\ref{sec:temporal_alignment_reward}. 
Reward computation uses Qwen3-Next-80B-A3B~\cite{yang2025qwen3} as the LLM for both claim extraction and consistency gating, served via VLLM~\cite{kwon2023efficient} framework for efficient inference. For text similarity, we use \texttt{sentence-transformers/all-MiniLM-L6-v2} model from Sentence Transformers.

\subsection{Evaluation Protocol}
All reasoning and non-reasoning models are evaluated under a unified configuration using lmms-eval~\cite{zhang2025lmms}. Videos are sampled at 2\,FPS (maximum 128 frames, 360$\times$420\,px resolution). Each reasoning model uses its official prompt template as given in the respective paper, and non-reasoning baselines follow the best configurations reported in their respective papers.~Inference uses greedy decoding for all models. Subtitles are excluded during the evaluation to prevent linguistic leakage and ensure a purely visual reasoning evaluation. More details are provided in Sec.~\ref{sec:implementation_details}.

\subsection{Compute and Environment}
All experiments are conducted on 8$\times$AMD Instinct™ MI210 accelerators (64 GB each, 512 GB total) using PyTorch~2.7 and ROCm~6.4. We use bf16 precision throughout both training and evaluation. VLLM~\cite{kwon2023efficient} is used to efficiently serve the Qwen3-Next-80B-A3B~\cite{yang2025qwen3} LLM during reinforcement learning. SFT uses 10,467 samples and completes in approximately 3.8 hours, while GRPO uses 4,804 samples and completes in about 54.4 hours. Evaluation on all 11 benchmarks with a maximum of 128 frames and a maximum of 1024 output tokens takes around 27.5 hours. We use Qwen3-Next-80B-A3B, which maintains a balance between efficiency and accuracy with a total of 80B parameters and only 3B active parameters, as the LLM judge in the Video Attention Score (VAS) calculation, as well as for the extraction of answers from thinking traces in the Think-Answer Consistency (TAC) calculation.
The complete training and evaluation codebase, dataset, and model checkpoints will be open-sourced for reproducibility.

\subsection{Hyperparameters}
A detailed summary of all key hyperparameters for both the training stages and the evaluation is provided in Table~\ref{tab:hyperparams}.

\begin{table*}[t]
\centering
\resizebox{\linewidth}{!}{
\begin{tabular}{llll}

\toprule
\textbf{Category} & \textbf{Setting} & \textbf{Value} & \textbf{Notes} \\

\midrule
\multirow{2}{*}{\textbf{Base model}} 
& Backbone & Qwen2.5-VL-7B & Non-reasoning base model \\
& Training stages & SFT $\rightarrow$ GRPO & Standard reasoning-tuning pipeline \\

\midrule
\multirow{3}{*}{\textbf{Frame sampling}} 
& FPS & 2 & 2\,FPS $\rightarrow$ Uniform frame sampling \\
& Max frames & 128 / 32 / 128 & SFT / GRPO / Eval \\
& Frame resolution & 360$\times$420 px & Applied to all datasets \\

\midrule
\multirow{3}{*}{\textbf{Visual preprocessing}} 
& Subtitles & WhisperX Large-v3 & Printed bottom during training only \\
& Timestamps & HH:MM:SS format & Top-left (train and eval) \\
& Color scheme & Light-on-dark / dark-on-light & Improves visibility and grounding \\

\midrule
\multirow{7}{*}{\textbf{SFT configuration}} 
& Epochs & 1 &  \\
& Batch size & 32 &  \\
& Learning rate & 1e-5 & Cosine LR scheduler \\
& Weight decay & 0.01 &  \\
& Max tokens (context) & 32K & Precision: bf16 \\
& Adapter type & LoRA ($r$=64) & Applied to Q/K/V/FFN modules \\
& Training time& 3.8 hours & 8$\times$AMD Instinct™ MI210 \\

\midrule
\multirow{10}{*}{\textbf{GRPO configuration}} 
& Epochs & 1 &  \\
& Group size ($K$) & 8 & Number of candidate generations \\
& KL coefficient ($\beta$) & 0.04 & Anchor regularization strength \\
& Batch size & 8 &  \\
& Learning rate & 1e-6 &  \\
& Reward weights & $\lambda_{\text{acc}} = \lambda_{\text{fmt}} = \lambda_{\text{tar}} = 1$ & Accuracy, format, TAR components \\\
& Hyperparameters & $\Delta=2$, $\tau=0.75$ & Temporal and semantic thresholds \\
& Judge model & Qwen3-Next-80B-A3B & Served via VLLM API \\
& Sentence Embeddings & \texttt{all-MiniLM-L6-v2} & Used for claim similarity matching \\
& Training time& 54.4 hours & 8$\times$AMD Instinct™ MI210 \\

\midrule
\multirow{8}{*}{\textbf{Evaluation}} 
& Benchmarks & 11 total & 5 generic + 6 reasoning \\
& Framework & lmms-eval & Official prompts for all models \\
& Decoding & Greedy & temperature=0 \\
& Frames used & 2 FPS, max 128 & No subtitles during evaluation \\
& Metrics & TAC, VAS, Accuracy & Consistent setup for all methods \\
& Judge model & Qwen3-Next-80B-A3B &  Used for VAS \\
& Max output tokens& 1024 & Max generation length of 1024\\
& Evaluation time& 27.5 hours & 8$\times$AMD Instinct™ MI210 \\

\midrule
\multirow{2}{*}{\textbf{Environment}} 
& GPUs & 8$\times$AMD Instinct™ MI210 & PyTorch 2.7, ROCm 6.4 \\
% & LLM serving & VLLM & For Qwen3-Next-80B-A3B \\
& Code release & Planned & Full training and evaluation pipeline \\
\bottomrule

\end{tabular}
}
\vspace{-0.5em}
\caption{Comprehensive summary of all hyperparameters used for~\MyModel~training and evaluation. The table details configuration choices across both supervised fine-tuning (SFT) and reinforcement learning (GRPO) stages, including visual preprocessing, frame sampling, model adaptation, and evaluation protocols. These settings ensure reproducible results across all benchmark evaluations.}
\label{tab:hyperparams}
\vspace{-1em}
\end{table*}

%% file: supplemental/TAC.tex
\section{Think--Answer Consistency (TAC)}
\label{supp:tac}

% % % % % % % % % % % % % % Prompt for Answer Extraction % % % % % % % % % % % % % %
\subsection{Prompt for Answer Extraction}
\label{supp:tac_prompt_for_answer_extraction}

To compute the Think--Answer Consistency (TAC) metric, we extract the answer predicted in model's,
\begin{enumerate}
    \item Reasoning trace (\texttt{<think>...</think>} block).
    \item Final output (\texttt{<answer>...</answer>} block). 
\end{enumerate}
We use deterministic parsing prompts that instruct LLM to extract the final choice directly from the model's output.

\noindent
\textbf{(1) Parsing from Reasoning Traces.}~To extract the answer implied by the reasoning trace, we use a stricter prompt that scans for the last explicit conclusion mentioned within the \texttt{<think>...</think>} section. Refer to {\color{blue}\texttt{Prompt}~\textbf{A}}.

\noindent
\textbf{(2) Parsing from Final Answers.}~The {\color{blue}\texttt{Prompt}~\textbf{B}} provides the prompt we use to extract the final answer from the model’s \texttt{<answer>...</answer>} block. 

%%%%%%%%%%%%%% Prompt A: TAC %%%%%%%%%%%%%%
\begin{tcolorbox}[
  breakable,         % allow page breaks
  colback=gray!5!white,
  colframe=gray!60!black,
  title={\textbf{Prompt~A:} TAC - Prompt for Answer Extraction from \texttt{<think>...</think>}},
  width=\linewidth,                  % fit text width
  before skip=4pt,after skip=6pt,    % compact vertical spacing
  boxsep=4pt,left=2pt,right=2pt,top=2pt,bottom=2pt
]
\small
\textbf{System Prompt:}\\
You are a strict extractor.\\
\textbf{Task}: Read ONLY the Reasoning text appended at the end of the user message and output a SINGLE LETTER corresponding to the option that the Reasoning explicitly concludes as the final answer.\\

{\color{blue}Rules (follow in order):}
\begin{enumerate}
    \item Look ONLY at the Reasoning text. Ignore the options text for decision-making.
    \item If the Reasoning contains an explicit final choice (e.g., ``Therefore, D'', ``Answer: A''), output that letter. If multiple explicit finals appear, output the LAST one.
    \item If the Reasoning labels a single option as correct (e.g., ``D is correct''), output that letter. If multiple appear, output the LAST one.
    \item If the Reasoning states the correct option by its text (e.g., ``the correct answer is `red car' ''), match that text exactly to the provided options and output the corresponding letter. Prefer the LAST explicit conclusion.
    \item Do NOT judge correctness or infer meaning. Output only the explicit final conclusion.
    \item Output a single uppercase letter (A--Z). No punctuation, no explanations.\\
\end{enumerate}

\textbf{User Prompt:}\\
\texttt{\color{blue}If is\_mcq\_task = True:}\\
    \qquad Options:\\
    \qquad A. \ldots\\
    \qquad B. \ldots\\
    \qquad C. \ldots\\[2pt]
    \qquad Reasoning:\\
    \qquad \texttt{\color{blue} text inside <think>...</think> block}\\[2pt]
    \qquad \underline{MCQ output format:}\\
    \qquad - Return ONLY one capital letter A--Z on a single line.\\
    \qquad - Do NOT include any other characters or spaces.\\[4pt]
\texttt{\color{blue}Else (open-ended task):}\\
    \qquad Reasoning:\\
    \qquad \texttt{\color{blue} text inside <think>...</think> block}\\[2pt]
    \qquad \underline{Open-form output format:}\\
    \qquad - If numeric, return ONLY the number.\\
    \qquad - If textual, return ONLY the minimal text answer.\\
    \qquad - Output must be a single line with no extra characters.
\end{tcolorbox}
%%%%%%%%%%%%%% Prompt A: TAC %%%%%%%%%%%%%%
\vspace{0.5em}

%%%%%%%%%%%%%% Prompt B: TAC %%%%%%%%%%%%%%
\begin{tcolorbox}[
  breakable,         % allow page breaks
  colback=gray!5!white,
  colframe=gray!60!black,
  title={\textbf{Prompt~B:} TAC - Prompt for Answer Extraction from \texttt{<answer>...</answer>}},
  width=\linewidth,                  % fit text width
  before skip=4pt,after skip=6pt,    % compact vertical spacing
  boxsep=4pt,left=2pt,right=2pt,top=2pt,bottom=2pt
]
\small
\textbf{System Prompt:}\\
You are a deterministic parsing agent.\\
\textbf{Task}: Read ONLY the provided text and emit a SINGLE-LINE answer in the exact format requested.\\

{\color{blue}Hard rules (apply all):}\\
\begin{enumerate}
    \item Do not infer or reason beyond the text. If the text lacks a valid answer, output exactly: \texttt{UNKNOWN}.
    \item Output must contain no explanations, no extra words, no labels, no code fences, no quotes, no brackets.
    \item Strip leading/trailing whitespace. No trailing punctuation unless required by format (e.g., a \%).
    \item Normalize internal whitespace to single spaces.
    \item Treat case-insensitive tokens like \texttt{option c}, \texttt{(c)}, \texttt{[c]}, \texttt{C)} as the letter C when MCQ is requested.\\
\end{enumerate}

\textbf{User Prompt:}\\
\texttt{\color{blue}If is\_mcq\_task = True:}\\
    \qquad Options:\\
    \qquad A. \ldots\\
    \qquad B. \ldots\\
    \qquad C. \ldots\\[2pt]
    \qquad Text to parse (final answer snippet):\\
    \qquad \texttt{\color{blue} text inside <answer>...</answer> block}\\[2pt]
    \qquad \underline{MCQ output format:}\\
    \qquad - Return ONLY one capital letter A--Z on a single line.\\
    \qquad - Do NOT include any other characters or spaces.\\[4pt]
\texttt{\color{blue} Else (open-ended task):}\\
    \qquad Text to parse (final answer snippet):\\
    \qquad \texttt{\color{blue} text inside <answer>...</answer> block}\\[2pt]
    \qquad \underline{Open-form output format:}\\
    \qquad - If numeric, return ONLY the number.\\
    \qquad - If textual, return ONLY the minimal text answer.\\
    \qquad - Output must be a single line with no extra characters.
\end{tcolorbox}
%%%%%%%%%%%%%% Prompt B: TAC %%%%%%%%%%%%%%
\vspace{0.5em}
\noindent
\textbf{Usage.}~Both {\color{blue}\texttt{Prompt}~\textbf{A} and~\textbf{B}}, are used to extract $\hat{a}_{\text{think}}$ and $\hat{a}_{\text{answer}}$ for each sample, which are compared to compute the TAC metric as defined in Eq.~\ref{main:equation_1} of the main paper:
\begin{equation*}
\text{TAC}(M, D) = \frac{1}{|D_{\text{correct}}|} 
\sum_{i \in D_{\text{correct}}} 
\mathbb{I}[\hat{a}_i^{\text{think}} = \hat{a}_i^{\text{answer}}].
\end{equation*}

% % % % % % % % % % % % % % Prompt for Answer Extraction % % % % % % % % % % % % % %

% % % % % % % % % % % % % % TAC versus TAC-All % % % % % % % % % % % % % %
\subsection{TAC vs TAC-All}
\label{supp:tac_vs_tac_all}

In Eq.~\ref{main:equation_1} of the main paper, the Think--Answer Consistency (TAC) metric is computed only over correctly answered samples, denoted as $D_{\text{correct}}$, to avoid cases where logically consistent but incorrect reasoning could inflate the score. 
For completeness, we also evaluate a variant, \textbf{TAC-All}, which measures the same consistency but across \emph{all} evaluation samples ($D$). 
This variant provides an absolute measure of reasoning stability independent of answer accuracy, capturing how consistently a model maintains internal alignment between its reasoning and final prediction regardless of the correctness.

\noindent
As shown in Tab.~\ref{supp:table_tac_vs_tac_all}, both TAC and TAC-All show a similar trend across all benchmarks. 
Our model, \MyModel, maintains strong logical alignment under both metrics, achieving the highest scores across most of the benchmarks. 
While TAC focuses on interpretable correctness, TAC-All confirms that \MyModel~produces coherent reasoning traces even for incorrect predictions.

%%%%%%%%%%%%%  Table %%%%%%%%%%%%%
\begin{table*}[!t]
\centering
\setlength{\tabcolsep}{0.75em}
\begin{adjustbox}{max width=\textwidth}
\begin{tabular}{
  l|                               % Model
  *{5}{c}                          % 5 Generic benches
  |>{\columncolor{gray!10}}c|      % Avg Generic (boxed by vertical rules)
  *{6}{c}                          % 6 Reasoning benches
  |>{\columncolor{gray!10}}c|      % Avg Reasoning (boxed)
  >{\columncolor{gray!10}}c        % Avg Overall (boxed; right border too)
}
\toprule
% Group headers (span includes averages)
 &
\multicolumn{6}{c|}{\textbf{Generic Benchmarks}} &
\multicolumn{7}{c|}{\textbf{Reasoning Benchmarks}} &
\\
\cmidrule(lr){2-7}\cmidrule(lr){8-14}
% Rotated, bottom-aligned column headers (including averages)
\raisebox{-15ex}{\textbf{Model}} &
\rothead{MVBench} &
\rothead{VideoMME} &
\rothead{TempCompass} &
\rothead{MLVU} &
\rothead{LongVideoBench} &
\rothead{\textbf{Avg. (Generic)}} &
\rothead{VideoMathQA} &
\rothead{Video-MMMU} &
\rothead{MMVU-Val} &
\rothead{VSiBench} &
\rothead{MINERVA} &
\rothead{SciVideoBench} &
\rothead{\textbf{Avg. (Reasoning)}} &
\rothead{\textbf{Avg. (Overall)}} \\
\midrule
% ------------------- Data rows -------------------
\multicolumn{15}{l}{\textbf{Think--Answer Consistency (TAC)}} \\

Video-R1~\cite{videor12025} &
77.2 & 82.2 & 86.0 & 77.8 & 75.6 & 79.8 & 64.7 & 79.5 & 85.0 & 51.7 & 67.4 & 62.7 & 68.5 & 73.6 \\

VideoChat-R1~\cite{videochatR12025} &
70.7 & 79.5 & 79.3 & 80.4 & 79.8 & 77.9 & \textbf{68.9} & 81.4 & \underline{90.4} & 73.9 & 67.3 & 73.6 & 75.9 & 76.8 \\

VideoChat-R1.5~\cite{yan2025videochat_r1.5} &
69.1 & 82.4 & 80.5 & 82.1 & \textbf{85.1} & 79.9 & \underline{67.9} & 82.7 & \textbf{92.5} & \underline{75.2} & 68.6 & \underline{74.3} & \underline{76.9} & 78.2 \\

VideoRFT~\cite{videorft2025} &
\underline{79.0} & \underline{85.4} & \underline{86.3} & \textbf{86.7} & \underline{83.4} & \underline{84.1} & 66.4 & \underline{83.5} & 87.4 & 71.4 & \underline{74.7} & 69.9 & 75.5 & \underline{79.5} \\

\rowcolor{gray!10}
\textbf{\MyModel~(Ours)} &
\textbf{82.4} & \textbf{86.6} & \textbf{88.3} & \underline{86.1} & \textbf{85.1} & \textbf{85.7} & 63.7 & \textbf{84.6} & 90.2 & \textbf{78.4} & \textbf{75.1} & \textbf{77.3} & \textbf{78.2} & \textbf{81.6} \\

\midrule

\multicolumn{15}{l}{\textbf{Think--Answer Consistency (TAC) - All}} \\

Video-R1~\cite{videor12025} &
75.7 & 82.4 & 84.2 & 78.8 & 77.4 & 79.7 & 82.6 & 80.2 & 81.8 & 79.4 & 79.7 & 81.9 & 80.9 & 80.4 \\

VideoChat-R1~\cite{videochatR12025} &
74.8 & 81.5 & 78.4 & 82.9 & 84.2 & 80.4 & 62.2 & 81.2 & \underline{89.1} & \textbf{91.5} & 81.8 & 84.9 & 81.8 & 81.2 \\

VideoChat-R1.5~\cite{yan2025videochat_r1.5} &
72.6 & 83.2 & 78.5 & 84.1 & 88.2 & 81.3 & \textbf{86.0} & 79.7 & \textbf{89.5} & 84.2 & 83.2 & 76.0 & 83.1 & 82.3 \\

VideoRFT~\cite{videorft2025} &
77.4 & 86.2 & 82.6 & 87.2 & 84.8 & \underline{83.6} & 75.5 & \textbf{86.8} & 86.0 & 86.1 & \underline{83.3} & \underline{88.0} & \underline{84.3} & \underline{84.0} \\

\rowcolor{gray!10}
\textbf{\MyModel~(Ours)} &
\textbf{82.7} & \textbf{87.6} & \textbf{86.3} & \underline{86.7} & \underline{85.8} & \textbf{85.8} & \underline{83.1} & \underline{86.4} & 87.8 & \underline{90.6} & \textbf{84.9} & \textbf{89.2} & \textbf{87.0} & \textbf{86.5} \\

% --- Add more rows below; keep the same column structure ---
% ModelName &  &  &  &  &  &  &  &  &  &  &  &  &  &  \\
\bottomrule
\end{tabular}
\end{adjustbox}
\vspace{-0.5em}
\caption{
\textbf{Comparison between TAC and TAC-All.} 
TAC is computed only over correctly answered samples ($D_{\text{correct}}$), while TAC-All considers all predictions ($D_{\text{all}}$) regardless of correctness. 
The upper block reproduces the TAC results corresponding to the left heatmap of Fig.~\ref{fig:tac_vas_results}, whereas the lower block reports TAC-All. 
Both metrics exhibit similar relative trends, confirming that \MyModel~maintains strong reasoning-answer consistency even for incorrect responses.
}
\label{supp:table_tac_vs_tac_all}
\vspace{-1em}
\end{table*}
%%%%%%%%%%%%%  Table %%%%%%%%%%%%%

% % % % % % % % % % % % % % TAC versus TAC-All % % % % % % % % % % % % % %

% % % % % % % % % % % % % % TAC - Qualitative Comparision % % % % % % % % % % % % % %
\subsection{Qualitative Comparison}
\label{supp:tac_qualitative_examples}

\begin{figure*}[!t]
  \centering
  \includegraphics[width=\textwidth]{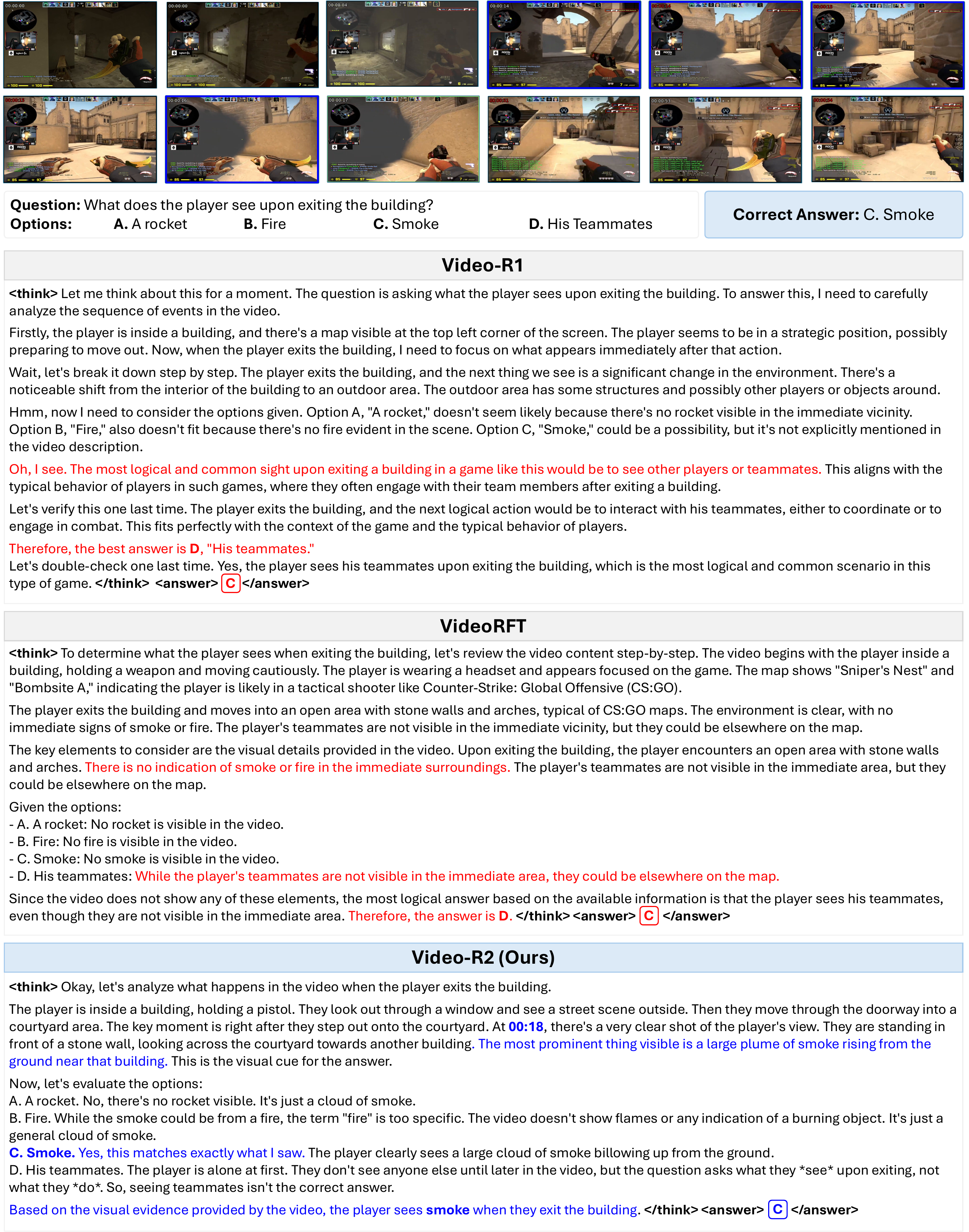}
  \caption{TAC comparison Example 1.}
  \label{fig:tac_ex1}
\end{figure*}

\begin{figure*}[!t]
  \centering
  \includegraphics[width=\textwidth]{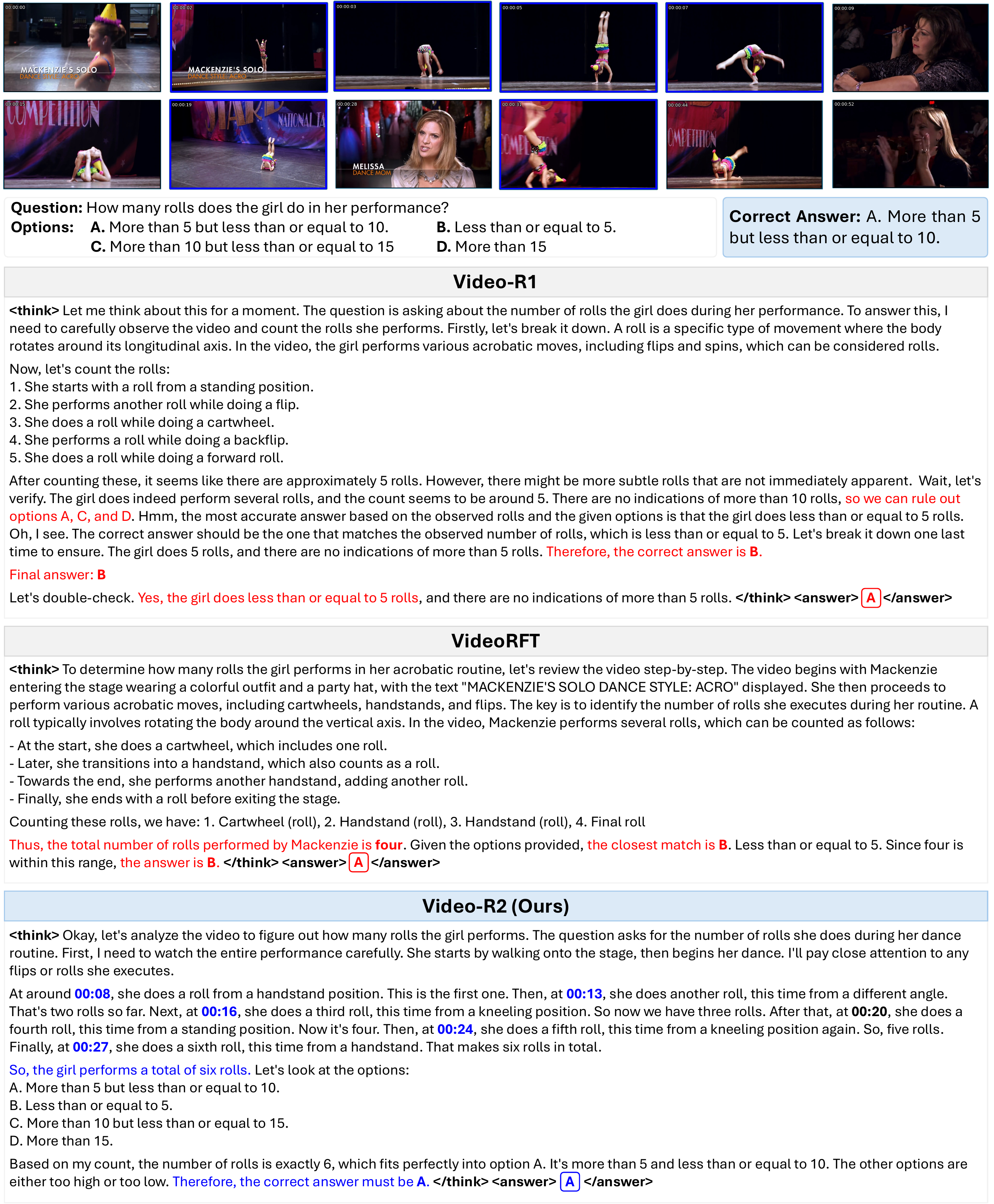}
  \caption{TAC comparison Example 2.}
  \label{fig:tac_ex2}
\end{figure*}

\begin{figure*}[!t]
  \centering
  \includegraphics[width=\textwidth]{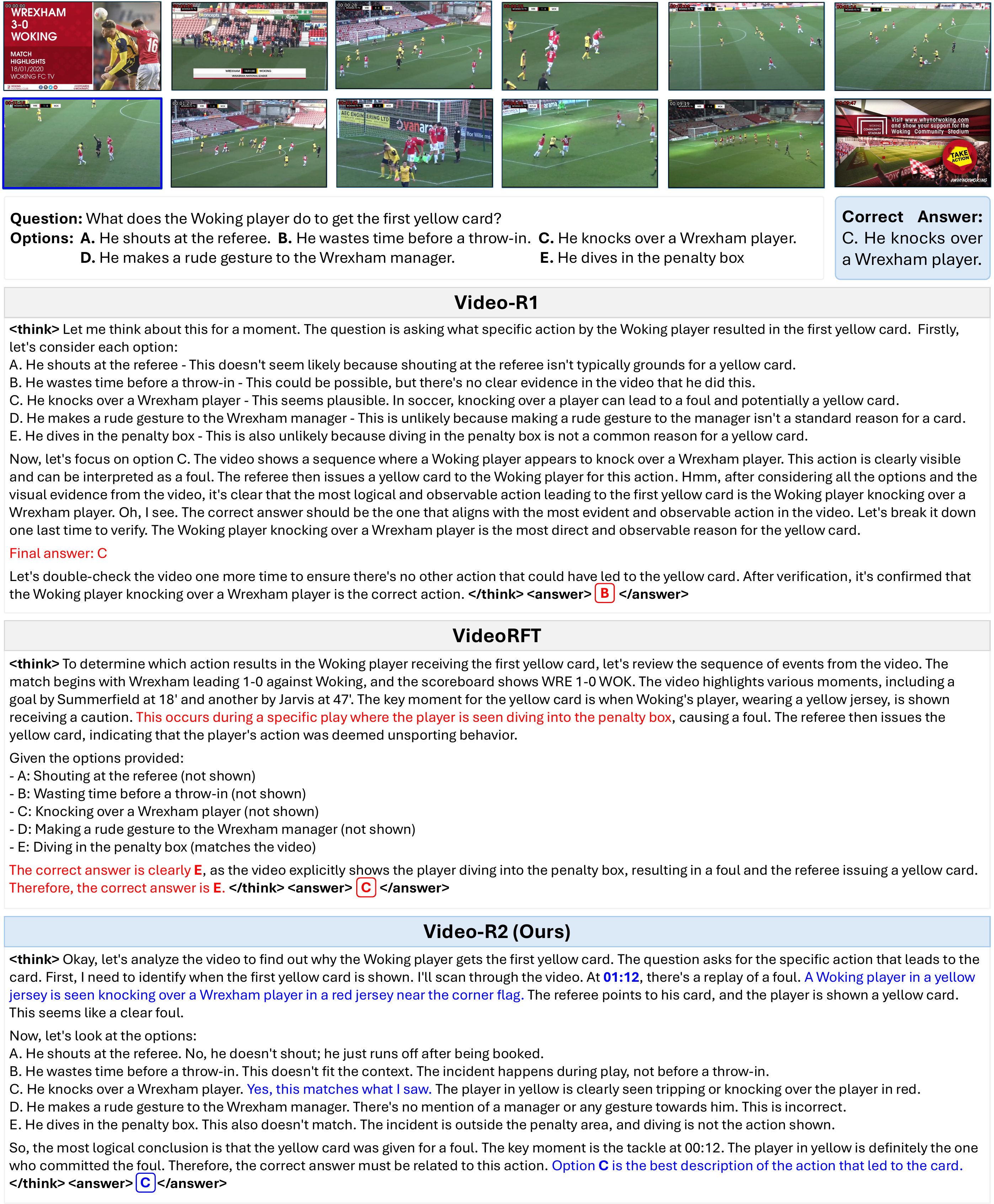}
  \caption{TAC comparison Example 3.}
  \label{fig:tac_ex3}
\end{figure*}

Figs.~\ref{fig:tac_ex1}-\ref{fig:tac_ex3} show qualitative differences in reasoning consistency across models, focusing on the Think--Answer Consistency (TAC) behavior of Video-R1~\cite{videor12025}, VideoRFT~\cite{videorft2025}, and our proposed~\textbf{\MyModel}. 
In each example, we visualize the video, question, options, correct answer, and reasoning traces generated by all models. 
While prior reasoning models such as Video-R1 and VideoRFT often produce logically inconsistent reasoning, where the reasoning trace concludes one option but the final \texttt{<answer>} outputs another, our~\MyModel~maintains internal alignment between reasoning and answer, demonstrating high TAC. 

\begin{itemize}
  \item \textbf{Example 1 (Fig.~\ref{fig:tac_ex1}):} A video-game scenario where Video-R1 and VideoRFT misalign between reasoning and final answer, while~\MyModel~maintains consistency and predicts the correct answer (“Smoke”).
  \item \textbf{Example 2 (Fig.~\ref{fig:tac_ex2}):} Counting acrobatic rolls in a dance performance. Both baselines exhibit reasoning-answer mismatch, whereas~\MyModel~accurately counts and maintains logical coherence.
  \item \textbf{Example 3 (Fig.~\ref{fig:tac_ex3}):} A soccer clip identifying the cause of a yellow card. Baselines misinterpret the reasoning context, producing contradictory answers, while~\MyModel~produces consistent and correct reasoning grounded in the visual evidence.
\end{itemize}

% ---------------- Example 1 ----------------

% ---------------- Example 2 ----------------

% ---------------- Example 3 ----------------

% % % % % % % % % % % % % % TAC - Qualitative Comparision % % % % % % % % % % % % % %

%% file: supplemental/VAS.tex
\section{Video Attention Score (VAS)}
\label{supp:vas}

% % % % % % % % % % % % % % Prompt for VAS Calculation % % % % % % % % % % % % % %
\subsection{Prompts for VAS Calculation}
To compute the Video Attention Score (VAS), we evaluate the degree to which the reasoning text (\texttt{<think>...</think>} block) \emph{claims} to rely on visual evidence from the video. 
The scoring prompt guides a large language model (LLM) to assign a 0-10 score. 
The complete prompt configuration is shown in {\color{blue}\texttt{Prompt}~\textbf{C}}.

%%%%%%%%%%%%%% VAS Prompt C %%%%%%%%%%%%%%
\begin{tcolorbox}[
  breakable,         % allow page breaks
  colback=gray!5!white,
  colframe=gray!60!black,
  title={\textbf{Prompt~C:} VAS - Prompt for Visual Attention Scoring from \texttt{<think>...</think>}},
  label={box:vas_prompt},
  width=\linewidth,                  % fit text width
  before skip=4pt,after skip=6pt,    % compact vertical spacing
  boxsep=4pt,left=2pt,right=2pt,top=2pt,bottom=2pt
]
\small
\textbf{System Prompt:}\\
\quad You are an expert judge of **claimed visual grounding** in video QA chain-of-thought. You will receive an Input (question/options/prompt) and a model Response that includes \texttt{<think>…</think>} and \texttt{<answer>…</answer>} tags. \\
\quad \textbf{Your task}:~Evaluate how much the reasoning inside \texttt{<think>} and \texttt{</think>} only *claims* to rely on concrete visual evidence from the video versus using text patterns, prior knowledge, or guessing. You do not have the video; score what the reasoning asserts, not whether those assertions are true or the final answer is correct.\\

{\color{blue}What counts as visual grounding signals (positive cues):}

\begin{enumerate}

\item Concrete, specific references to visible entities and attributes (colors, positions, counts, on-screen text/OCR).

\item Spatio-temporal descriptions tied to the video (“first/then,” “left/right,” “foreground/background,” scene changes).

\item Actions/interactions clearly described as seen (“picks up,” “points at,” “walks behind”).

\item Visual logic chained to the question (“Because the sign reads ‘Gate B’ and the arrow points left…”).

\item Mentions like “the video shows,” “the video states,” or “the video indicates…” count as *moderate visual grounding* **if** they reference specific factual content (e.g., numbers, object names, or events), even without spatial detail.\\ \\

\end{enumerate}

{\color{blue}What counts as non-visual or weak grounding (negative cues):}

\begin{enumerate}
\item Generic mentions of “the video” with no specific content (“the video talks about cars”).

\item Option pattern matching, stereotypes/common knowledge, or generic phrases (“based on the context,” “usually…”).

\item Restating the question or options with “I see” but no concrete visual detail.

\item Timestamp name-dropping without linked content, irrelevant flourish, or contradictions.\\

\end{enumerate}

{\color{blue}Edge cases:}

\begin{itemize}

\item If multiple \texttt{<think>} blocks exist, judge the first.

\item If \texttt{<think>} is missing or empty → score 0.

\item Ignore \texttt{<answer>} when scoring.\\

\end{itemize}

{\color{blue}Scoring (0-10, integer):}

\begin{enumerate}
\item 10 (Excellent): Dense, highly specific, step-by-step visual reasoning directly addressing the question.

\item 7–9 (Strong): Clearly video-based with several specific claims tied to what’s shown, even if phrased verbally (“the video states…”).

\item 4–6 (Moderate): Claims to recall or cite specific video facts (numbers, entities, or comparisons) but lacks spatial or temporal detail.

\item 1–3 (Weak): Minimal visual content; mostly generic or option-pattern reasoning with only token mentions of the video.

\item 0 (None): No visual grounding; guessing, meta-reasoning only, or \texttt{<think>} absent.\\
\end{enumerate}

\textbf{Output:} Return JSON only with keys ``score" (0–10 integer) and ``rationale" (2–3 concise sentences citing the most important cues; do not mention lack of video).\\

\textbf{User Prompt:}\\
Evaluate ATTENTION-TO-VIDEO in the model’s reasoning. Score only the text inside \texttt{<think>…</think>}, not the final \texttt{<answer>} and not answer correctness.\\

You do not have access to the video. Judge whether the reasoning \textit{claims} to use concrete visual evidence (what is seen, where it appears, how it changes) versus relying on generic patterns, options, or prior knowledge.\\

{\color{blue}Use this rubric (0–10):}

\begin{enumerate}
\item 10: Dense, specific visual descriptions (entities, attributes, counts, on-screen text, spatial relations, temporal order) directly supporting the question.

\item 7–9: Clearly video-based reasoning with multiple specific claims tied to what the video shows, even if phrased as “the video states…” or “the video shows…”.

\item 4–6: Claims to recall or cite specific video facts (numbers, labels, or entities) but lacks spatial or visual detail.

\item 1–3: Minimal visual content; generic “the video talks about…” or option-based logic.

\item 0: No visual grounding; guessing, meta-reasoning only, or missing \texttt{<think>} block.\\
\end{enumerate}

\noindent
Input: \texttt{\color{blue}\{input\_text\}}\\
Response: \texttt{\color{blue}\{response\_text\}}\\

Reply ONLY with JSON like: \texttt{\{"score": int, between 0 and 10, "rationale": "2 to 3 sentences"\}}
\end{tcolorbox}
%%%%%%%%%%%%%% VAS Prompt C %%%%%%%%%%%%%%

% % % % % % % % % % % % % % Prompt for VAS Calculation % % % % % % % % % % % % % %
\subsection{Qualitative Comparison}
\label{supp:vas_qualitative_comparison}

\begin{figure*}[!t]
  \centering
  \includegraphics[width=\textwidth]{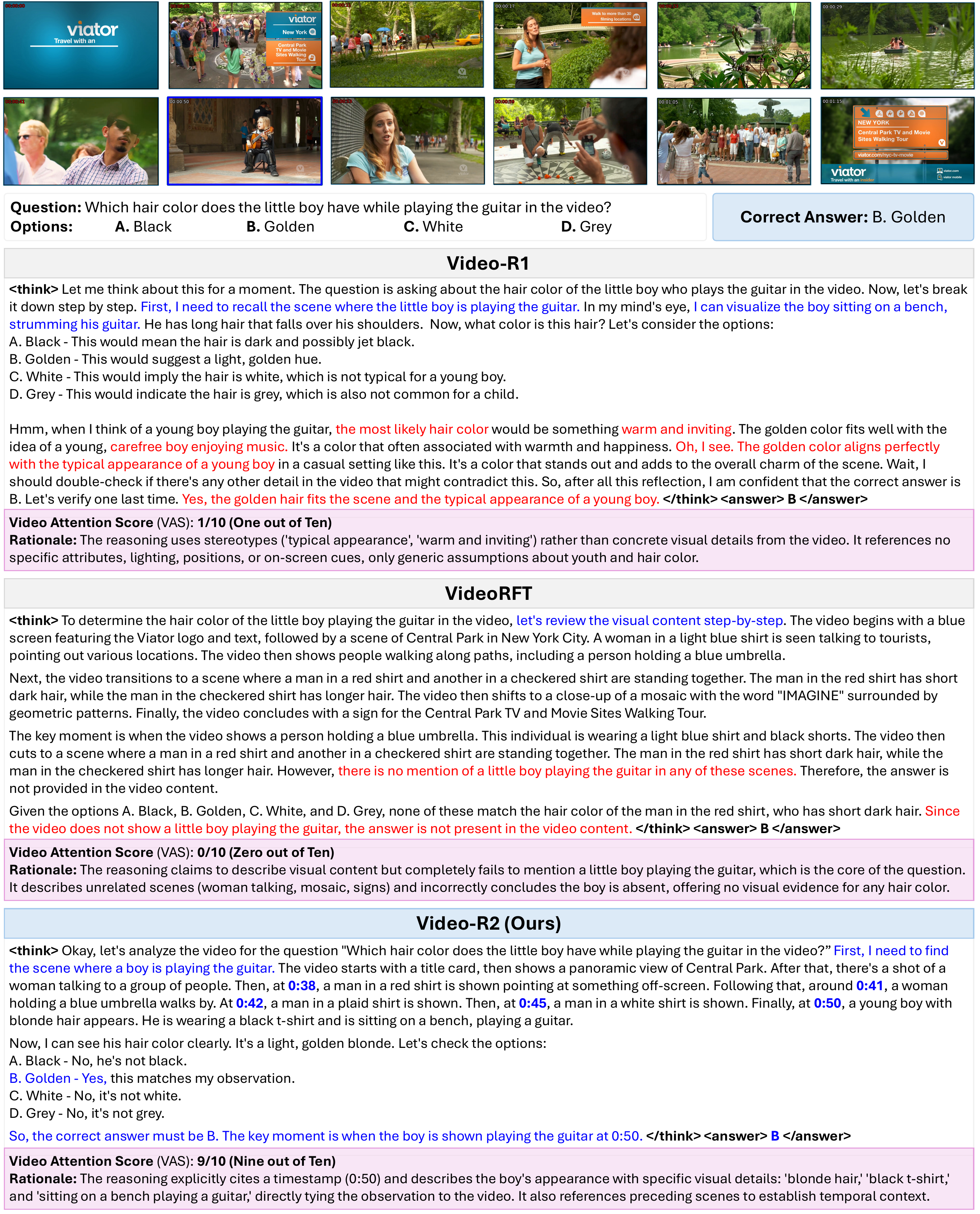}
  \caption{VAS comparison Example 1.}
  \label{fig:vas_ex1}
\end{figure*}

\begin{figure*}[!t]
  \centering
  \includegraphics[width=0.95\textwidth]{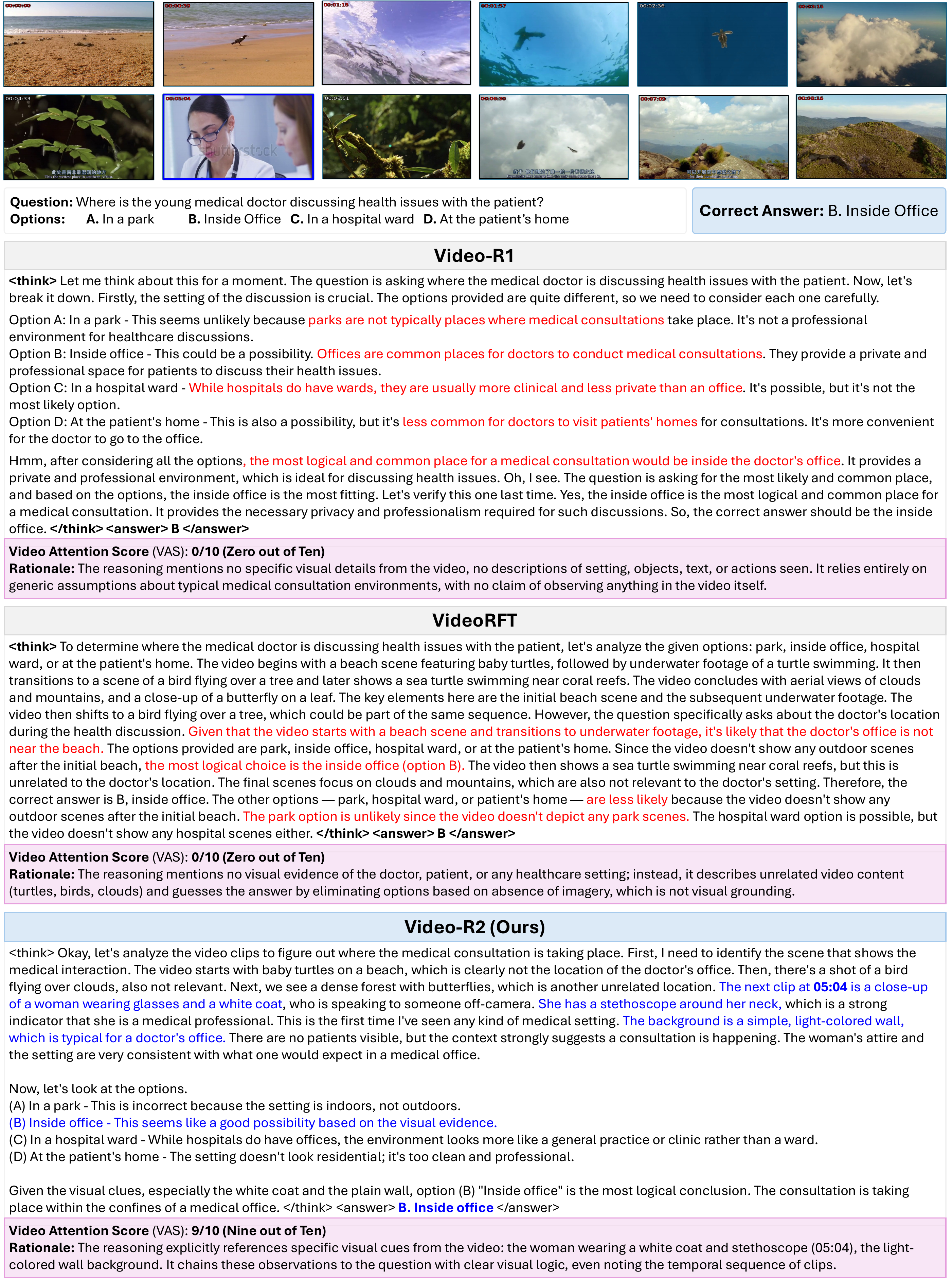}
  \caption{VAS comparison Example 2.}
  \label{fig:vas_ex2}
\end{figure*}

\begin{figure*}[!t]
  \centering
  \includegraphics[width=\textwidth]{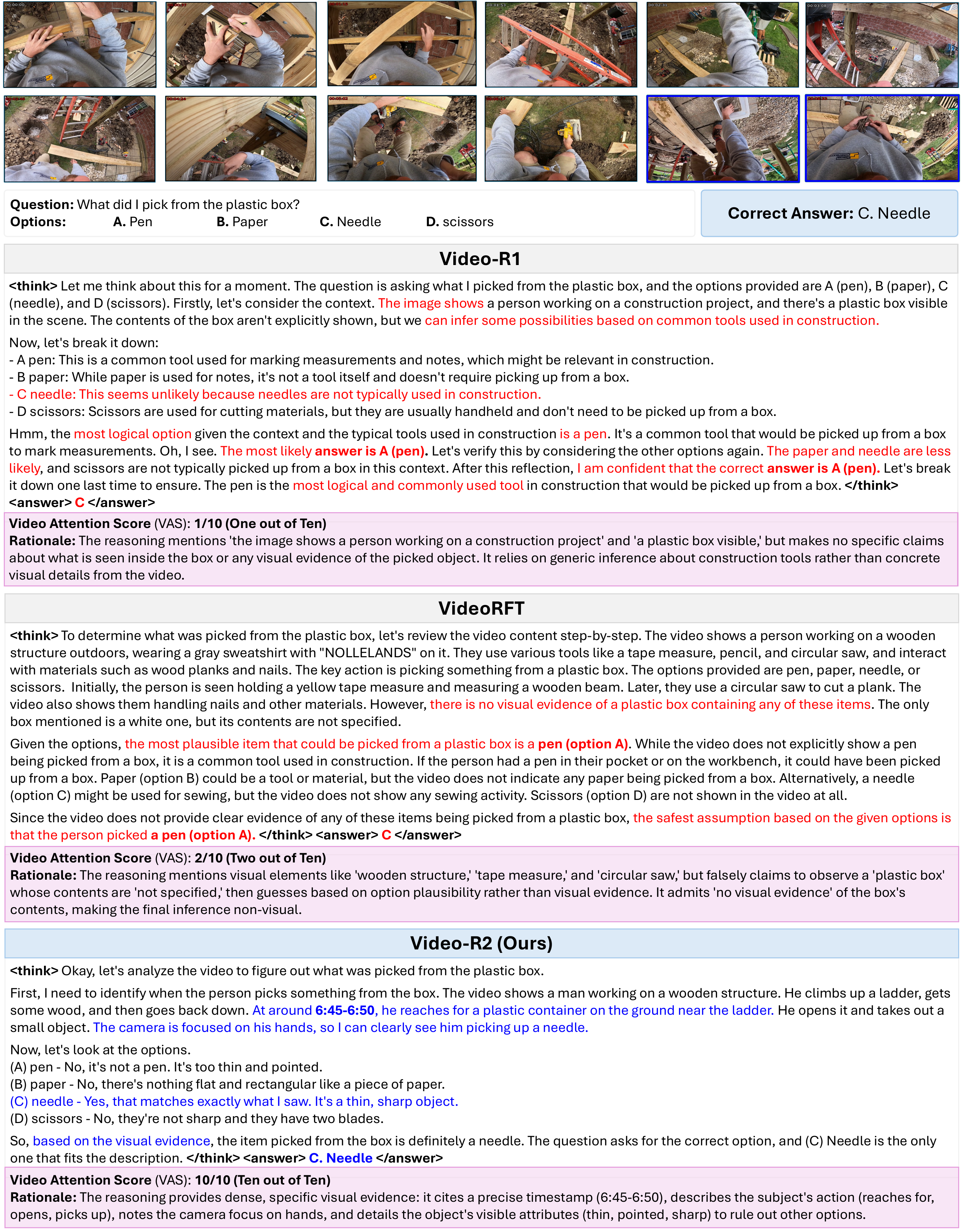}
  \caption{VAS comparison Example 3.}
  \label{fig:vas_ex3}
\end{figure*}

Figs.~\ref{fig:vas_ex1}-\ref{fig:vas_ex3} show how Video Attention Score (VAS) captures the claimed visual grounding quality of model reasoning.  
Each example compares reasoning traces from Video-R1~\cite{videor12025}, VideoRFT~\cite{videorft2025}, and our \textbf{\MyModel}, highlighting how visual specificity and temporal grounding differ across models.  
While baseline models often describe generic contexts or rely on prior knowledge, our~\MyModel~consistently references concrete visual details, such as timestamps, spatial positions, and detailed object descriptions, resulting in significantly higher VAS.  

\begin{itemize}
  \item \textbf{Example 1 (Fig.~\ref{fig:vas_ex1}):} A boy playing guitar in Central Park. Video-R1 and VideoRFT reasoning remain abstract or irrelevant, lacking any visual grounding, whereas~\MyModel~cites timestamps (0:50) and mention visual details such as “blonde hair,” “black t-shirt,” and “sitting on a bench,” achieving a VAS of 9/10.
  \item \textbf{Example 2 (Fig.~\ref{fig:vas_ex2}):} A medical consultation scene. Both baselines fail to describe visual evidence: one relying on generic assumptions, the other misinterpreting unrelated footage, while~\MyModel~grounds its reasoning in specific visual cues (white coat, stethoscope, and wall background) for a VAS of 9/10.
  \item \textbf{Example 3 (Fig.~\ref{fig:vas_ex3}):} A construction worker picking an object from a plastic box. Baselines forms a theory without observation of visual details, while~\MyModel~identifies the exact moment (6:45–6:50) and object features (“thin, sharp, needle-like”) for perfect visual grounding with a VAS of 10/10.
\end{itemize}

% ---------------- Example 1 ----------------

% ---------------- Example 2 ----------------

% ---------------- Example 3 ----------------

%% file: supplemental/TAR.tex
\section{Temporal Alignment Reward (TAR)}
\label{supp:tar}

\subsection{Prompt for Claim Extraction}
\label{supp:tar_prompt_for_claim_extraction}

To compute the Temporal Alignment Reward (TAR), we first extract timestamped \emph{claims} from both the reference and model-generated reasoning.  
Each claim represents a localized event or observation within the \texttt{<think>...</think>} block, expressed as a timestamp and its corresponding textual description.  
The extraction process is handled by an LLM-based parser that ensures temporal precision and semantic clarity.  
The prompt is provided in {\color{blue}\texttt{Prompt}~\textbf{D}}.

%%%%%%%%%%%%%% TAR - Prompt for Claim Extraction %%%%%%%%%%%%%%
\begin{tcolorbox}[
  breakable,         % allow page breaks
  colback=gray!5!white,
  colframe=gray!60!black,
  title={\textbf{Prompt~D:} TAR - Prompt for Claim Extraction},
  width=\linewidth,                  % fit text width
  before skip=4pt,after skip=6pt,    % compact vertical spacing
  boxsep=4pt,left=2pt,right=2pt,top=2pt,bottom=2pt
]
\small
\textbf{System Prompt:}\\
\quad You are a precise temporal information extraction assistant.\\

{\color{blue}Goal:}
Given ONLY the provided text, extract every timestamp mention and produce a compact JSON object mapping each timestamp (key) to ONE short sentence (value) describing what happens at that time according to the text.\\

{\color{blue}Output format (STRICT):}

\begin{enumerate} 

\item Return ONLY a JSON object. No prose, no code fences, no explanations.

\item Keys: timestamps exactly as mentioned, normalized to MM:SS or HH:MM:SS with leading zeros.
  
\quad \quad • Single time example: "00:42"
  
\quad \quad • Range examples: "00:42-00:45", "01:45-02:01"

\item Order keys by start time ascending.

\item If the text uses phrases like "around the 16-second mark" or "at about 1:02", convert to "00:16" or "01:02".

\item If a single continuous action is described across adjacent times (e.g., 00:07 and 00:08), you MAY consolidate into a range "00:07-00:08" with one concise sentence.

\item If the same timestamp appears multiple times, merge into a single key and summarize succinctly.

\item If no timestamps are present, return {}.\\

\end{enumerate}

{\color{blue}Content rules:}

\begin{enumerate}

\item Derive sentences ONLY from the given text. No hallucinations.

\item Each value is a brief, plain-language clause ($\leq$ 10 words), sentence case, ending with a period.

\item Prefer action-focused wording (“Performer executes a cartwheel while holding the jump rope.”).\\

\end{enumerate}

\textbf{User Prompt:}\\
Extract timestamped events from the following text and return ONLY the JSON object as specified.\\

{\color{blue}STRICT KEYS FORMAT:}

\begin{enumerate}

\item Use ONLY zero-padded MM:SS or HH:MM:SS.

\item For ranges, use a single hyphen: "MM:SS-HH:MM:SS".

\item Do NOT output plain seconds without a colon (e.g., "6" or "020" is forbidden).

\item Do NOT output words like "end", "EOF", "+", or any trailing symbols.

\item If the text mentions decimal seconds (e.g., 01.77s or 3.5s), round to the nearest second BEFORE emitting.

\item Normalize phrases like "around 16 seconds" to "00:16".

\item Keys must be unique and ordered by start time ascending.

\item If no timestamps are present, return {{}} exactly.\\

\end{enumerate}

{\color{blue}VALUES:}

\begin{enumerate}
\item Each value is ONE short clause ($\leq$ 10 words), sentence case, action-focused, ends with a period.

\item Derive strictly from the provided text (no hallucinations).\\

\end{enumerate}

Return ONLY the JSON object. No code fences, no explanations.\\

The reasoning text is:\\
{\{\color{blue}{the reasoning text}}\}
\end{tcolorbox}
%%%%%%%%%%%%%% TAR - Prompt for Claim Extraction %%%%%%%%%%%%%%

\subsection{Prompt for Consistency Gating}
\label{supp:tar_prompt_for_consistency_gating}
The consistency gating step in Temporal Alignment Reward (TAR) ensures that temporal alignment contributes to the reward only when the reasoning (\texttt{<think>...</think>}) and the final conclusion (\texttt{<answer>...</answer>}) agree.~The LLM determines whether the reasoning supports the same conclusion as given in the final answer and outputs a binary TRUE or FALSE signal used to gate the TAR computation.  

%%%%%%%%%%%%%% TAR - Prompt for Consistency Gating %%%%%%%%%%%%%%
\begin{tcolorbox}[
  breakable,         % allow page breaks
  colback=gray!5!white,
  colframe=gray!60!black,
  title={\textbf{Prompt~E:} TAR - Prompt for Consistency Gating},
  width=\linewidth,                  % fit text width
  before skip=4pt,after skip=6pt,    % compact vertical spacing
  boxsep=4pt,left=2pt,right=2pt,top=2pt,bottom=2pt
]
\small
\textbf{System Prompt:}\\
\quad You are a meticulous auditor. Determine whether the reasoning (THINK) and the final answer (ANSWER) are logically consistent with each other for the given question. Ignore style, verbosity, or extra details; focus strictly on whether the conclusion in THINK matches and supports the final ANSWER for the same question.\\

\textbf{User Prompt:}\\
\quad \texttt{QUESTION:}\\
\quad \texttt{\color{blue}\{question\}}\\[2pt]
\quad \texttt{THINK (model's internal reasoning):}\\
\quad \texttt{\color{blue}\{reasoning\}}\\[2pt]
\quad \texttt{ANSWER (model's final answer):}\\
\quad \texttt{\color{blue}\{answer\}}\\[4pt]
\quad \texttt{TASK:}\\
\quad 1) Output ONLY one of the \textbf{TRUE} or \textbf{FALSE} on the first line.\\
\quad \phantom{1)} -- \textbf{TRUE} $\Rightarrow$ \texttt{THINK} and \texttt{ANSWER} are consistent and express the same conclusion.\\
\quad \phantom{1)} -- \textbf{FALSE} $\Rightarrow$ \texttt{THINK} contradicts or does not support the \texttt{ANSWER} (e.g., mismatched conclusion).\\[2pt]
\quad 2) On the next line(s), give a brief justification (1--3 sentences) summarizing the key evidence supporting your decision.
\end{tcolorbox}
%%%%%%%%%%%%%% TAR - Prompt for Consistency Gating %%%%%%%%%%%%%%

% ---------------- TAR Calculation Examples ----------------
\subsection{TAR Calculation Examples}
\label{supp:tar_examples}

\begin{figure*}[!t]
  \centering
  \includegraphics[page=1,width=\textwidth]{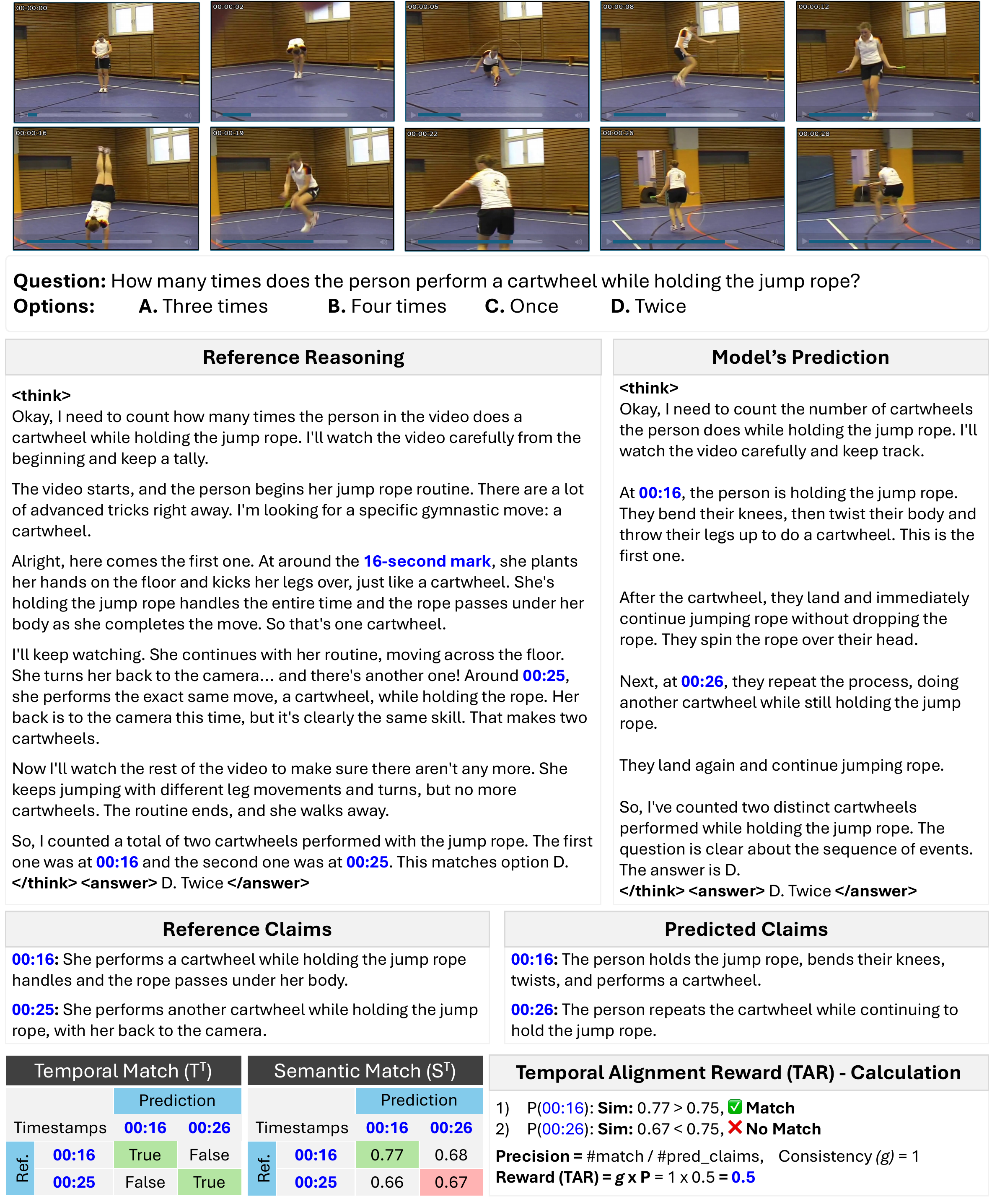}
  \caption{TAR calculation Example 1.}
  \label{fig:tar_ex2}
\end{figure*}

\begin{figure*}[!t]
  \centering
  \includegraphics[page=1,width=\textwidth]{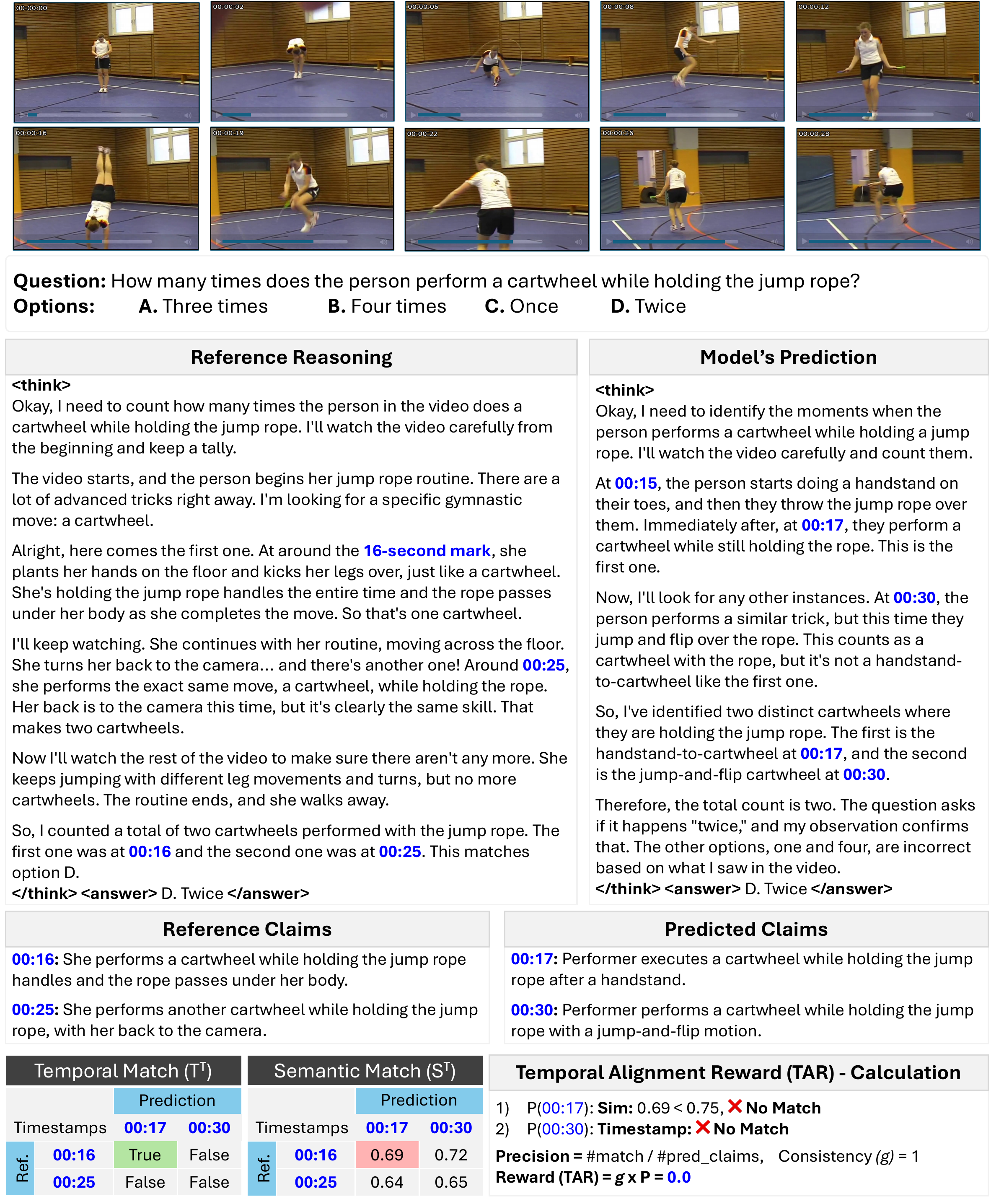}
  \caption{TAR calculation Example 2.}
  \label{fig:tar_ex3}
\end{figure*}

\begin{figure*}[!t]
  \centering
  \includegraphics[page=1,width=\textwidth]{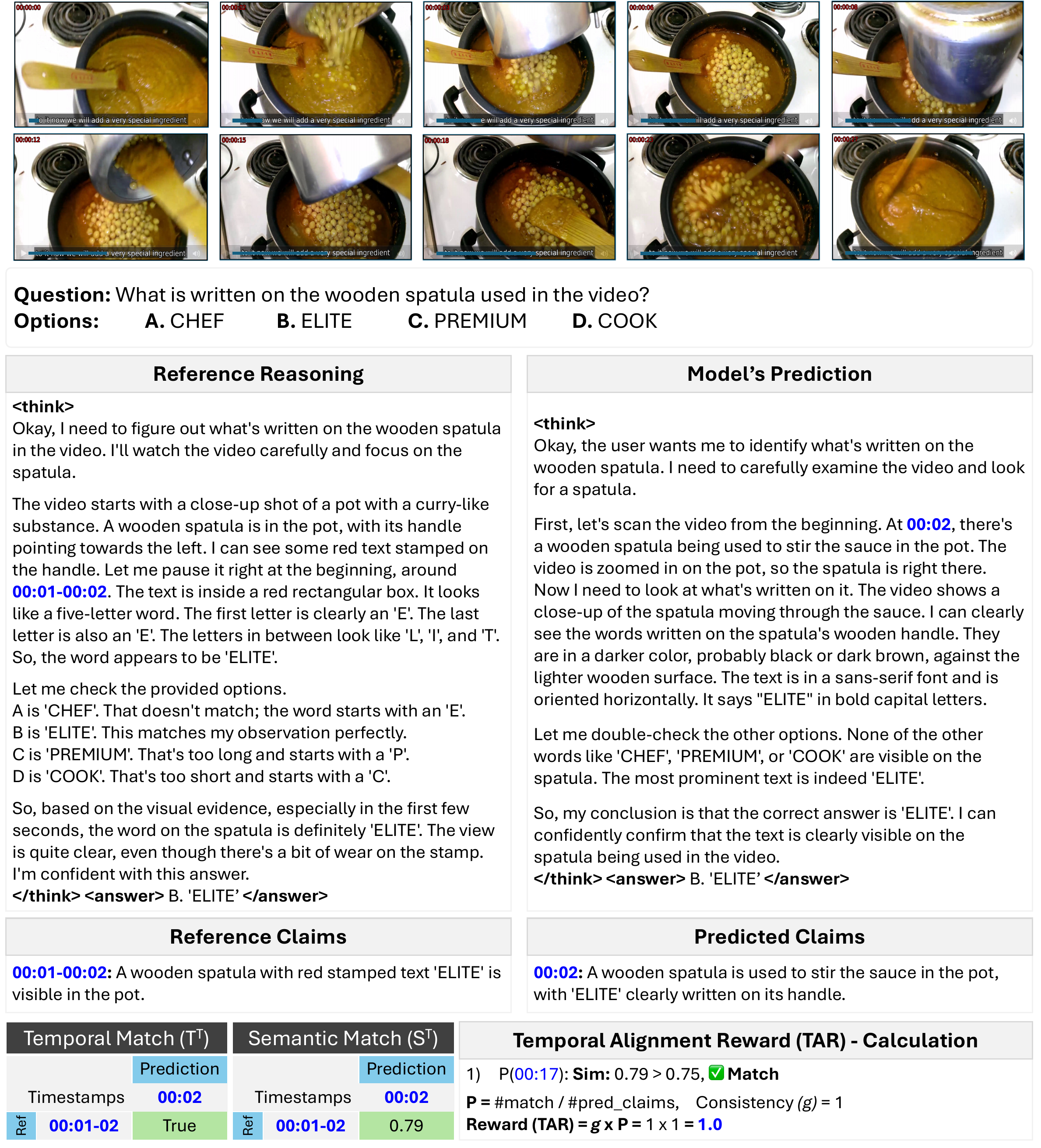}
  \caption{TAR calculation Example 3.}
  \label{fig:tar_ex4}
\end{figure*}

\begin{figure*}[!t]
  \centering
  \includegraphics[page=1,width=\textwidth]{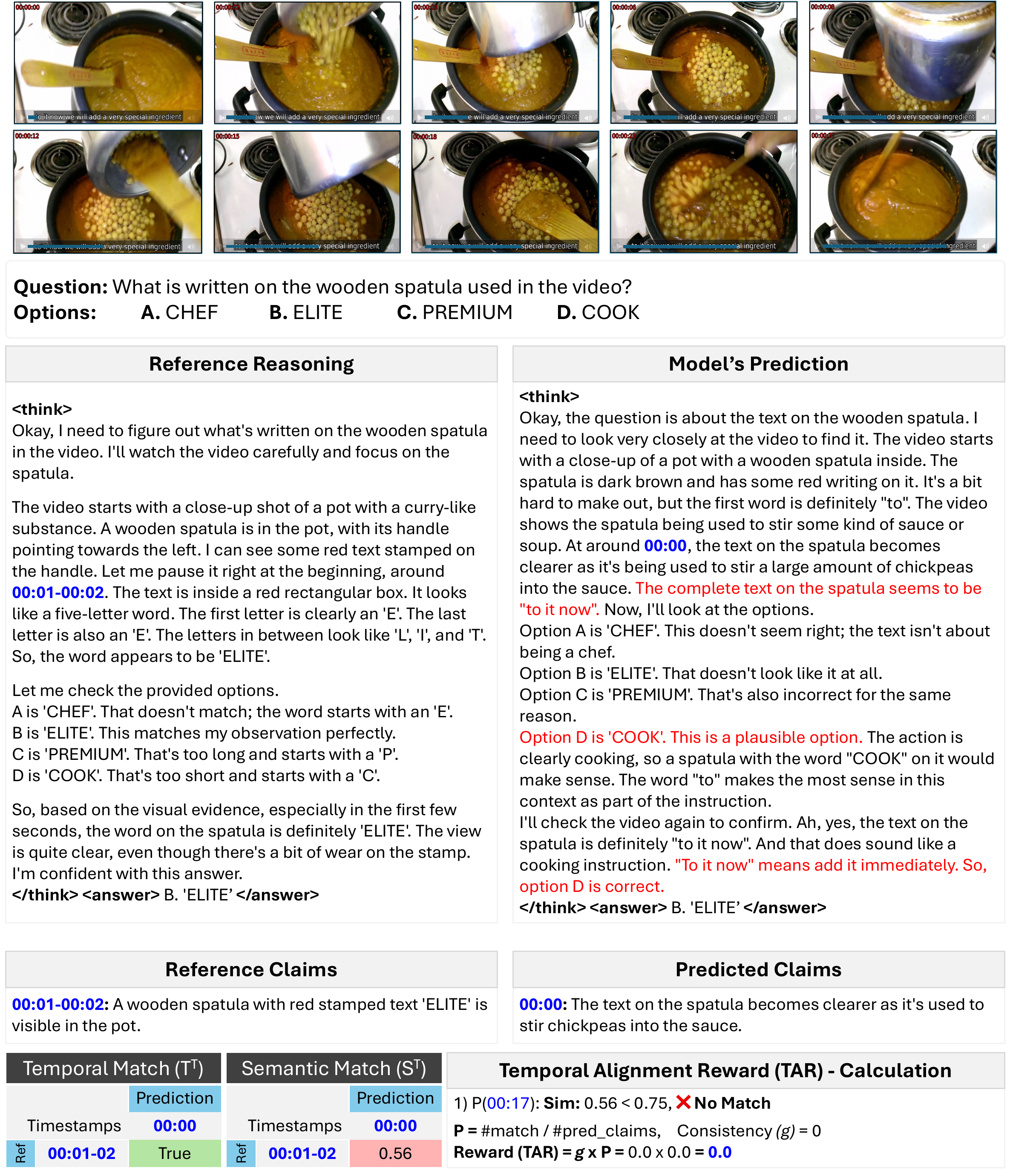}
  \caption{TAR calculation Example 5.}
  \label{fig:tar_ex5}
\end{figure*}

\begin{figure*}[!t]
  \centering
  \includegraphics[page=1,width=\textwidth]{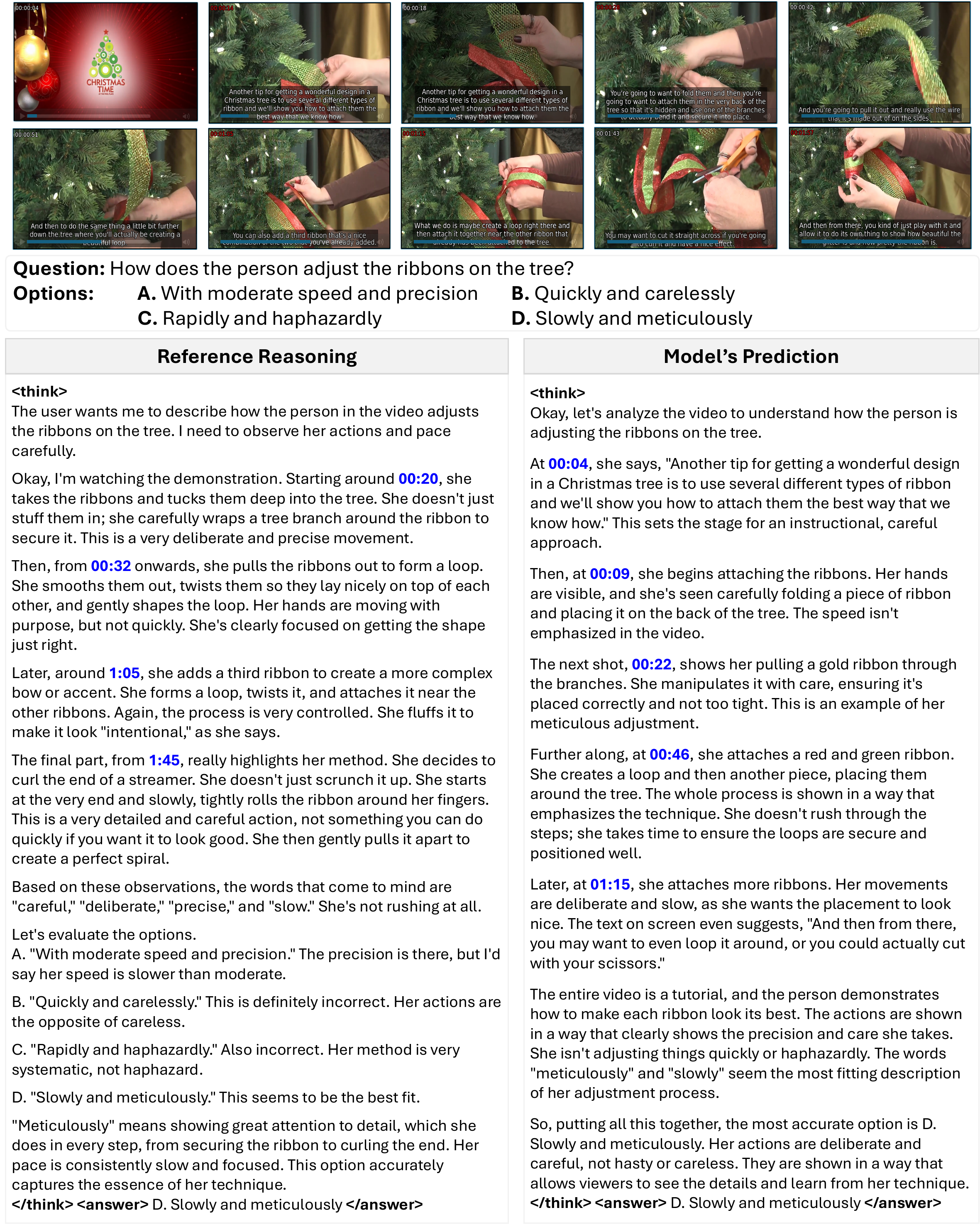}
  \caption{TAR calculation Example 5 (page 1 of 2).}
  \label{fig:tar_ex6_p1}
\end{figure*}

\begin{figure*}[!t]
  \centering
  \includegraphics[page=1,width=\textwidth]{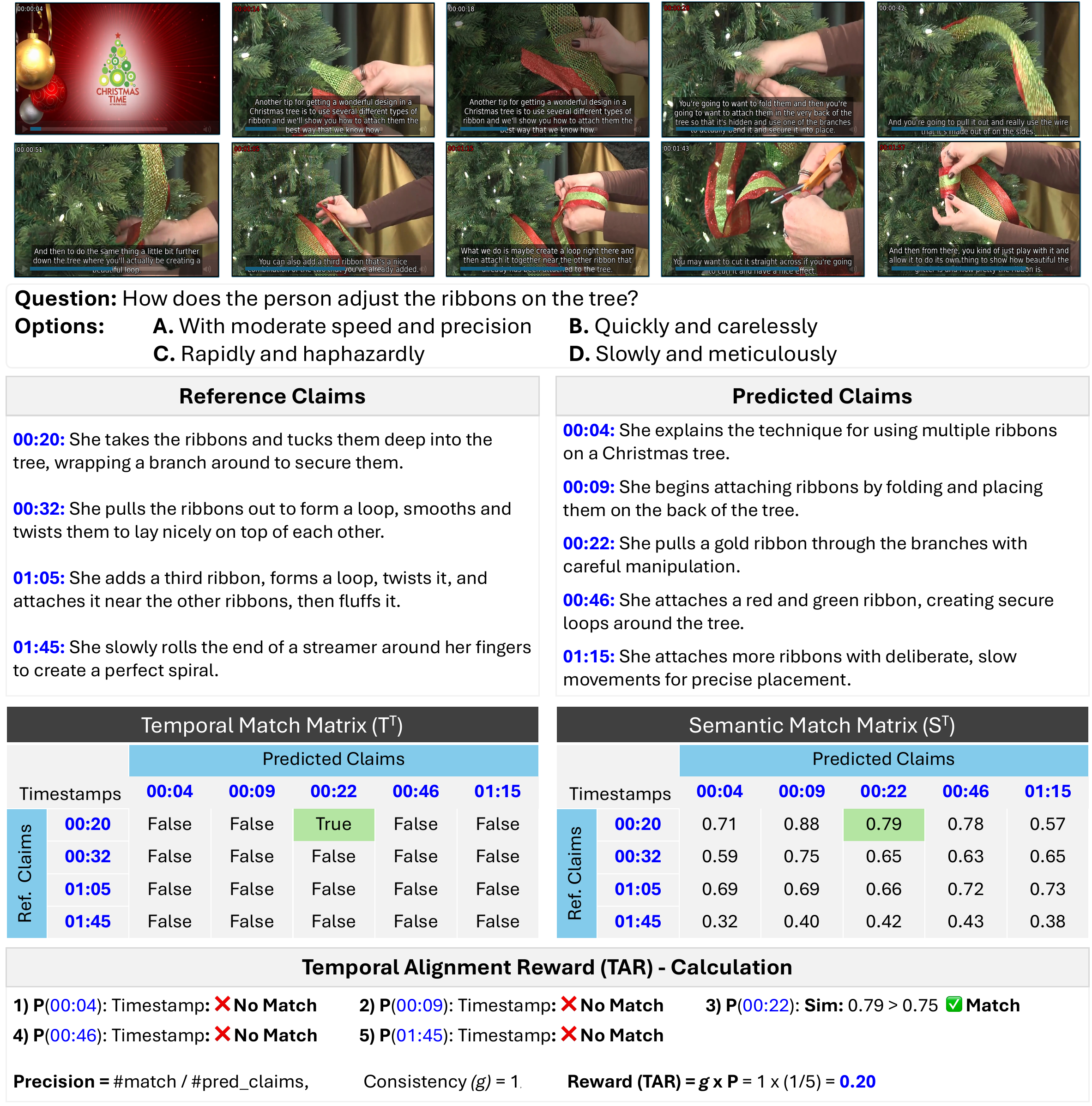}
  \caption{TAR calculation Example 5 (page 2 of 2).}
  \label{fig:tar_ex6_p2}
\end{figure*}

\begin{figure*}[!t]
  \centering
  \includegraphics[page=1,width=\textwidth]{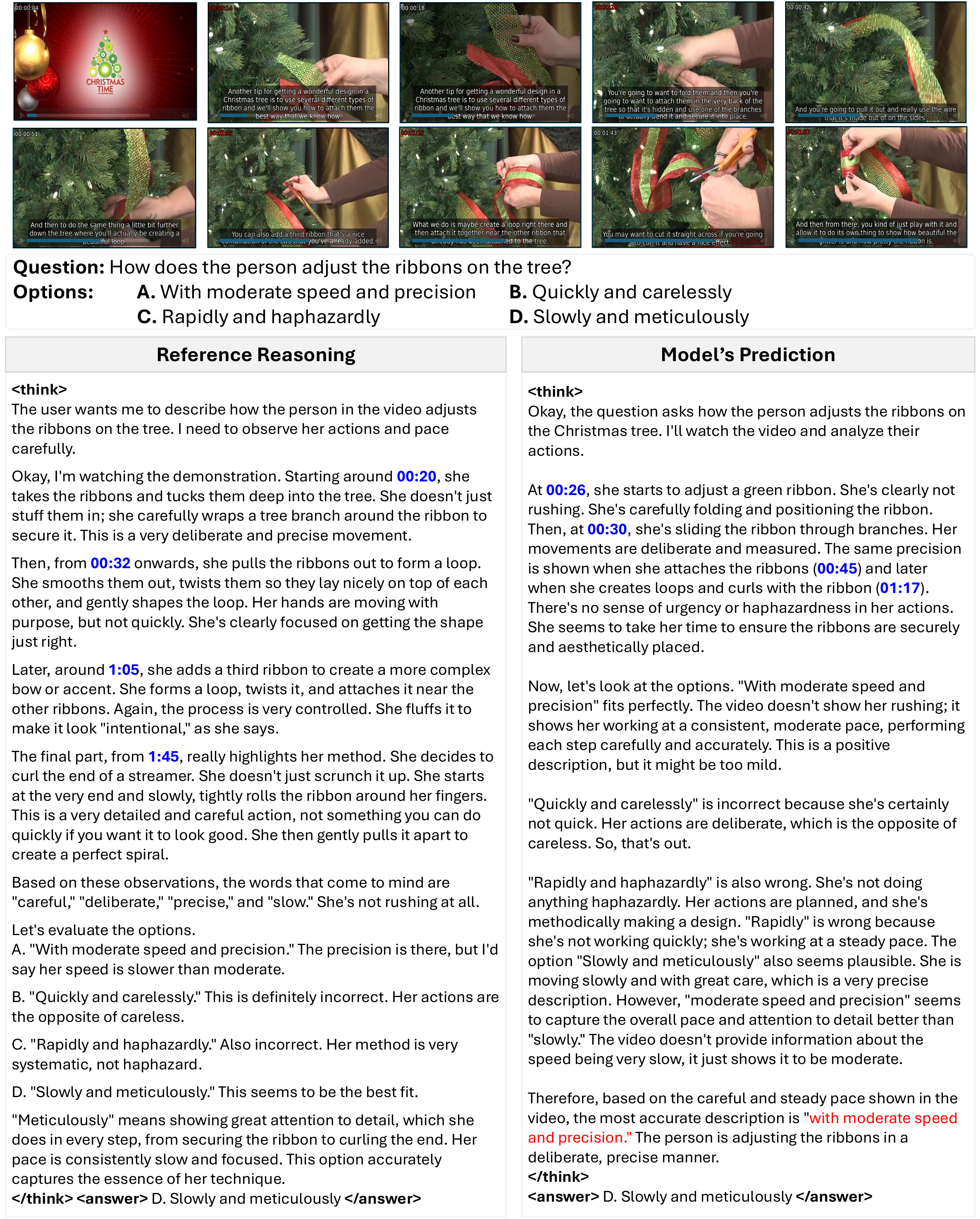}
  \caption{TAR calculation Example 6 (page 1 of 2).}
  \label{fig:tar_ex7_p1}
\end{figure*}

\begin{figure*}[!t]
  \centering
  \includegraphics[page=1,width=\textwidth]{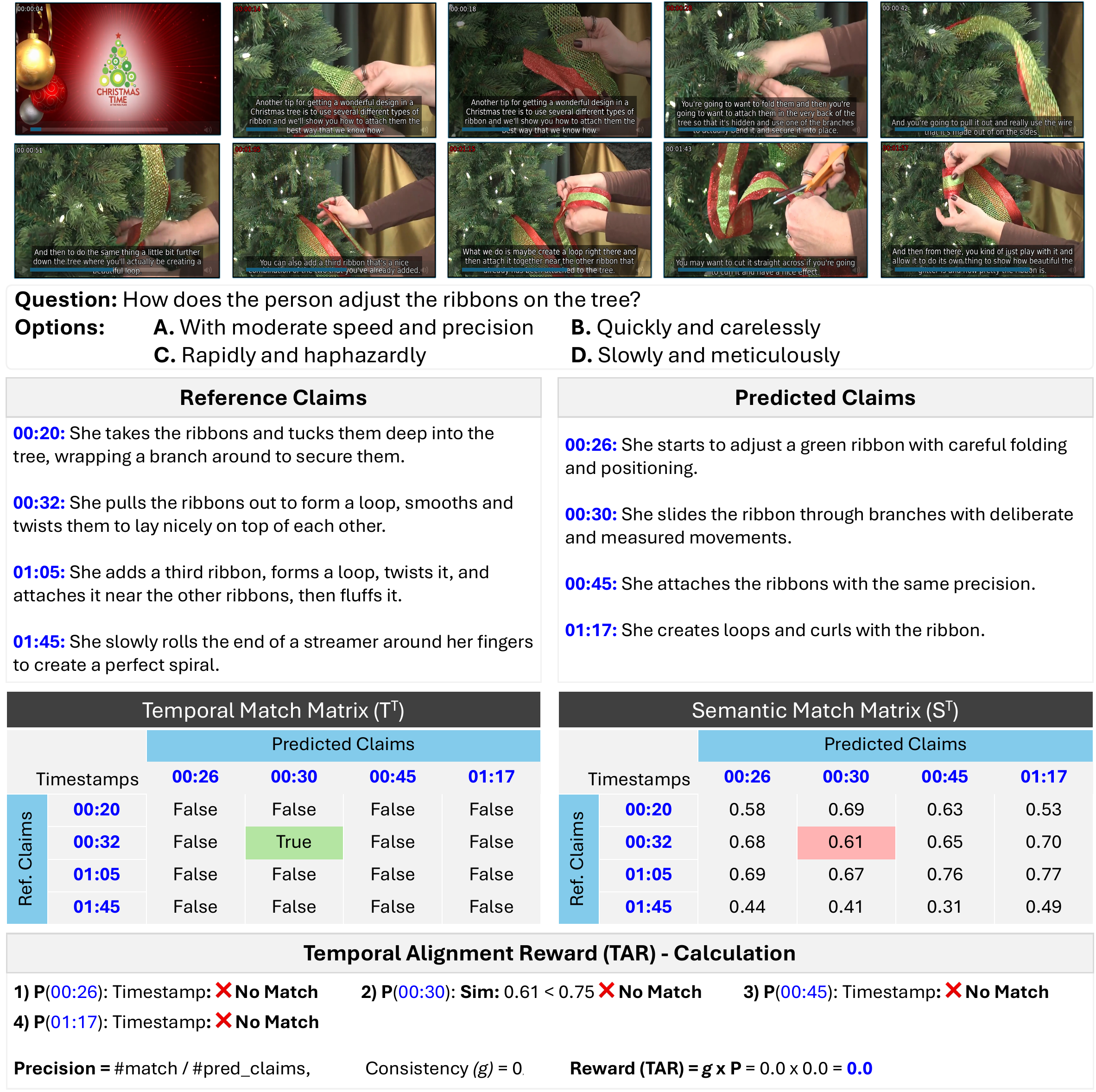}
  \caption{TAR calculation Example 6 (page 2 of 2).}
  \label{fig:tar_ex7_p2}
\end{figure*}

\begin{figure*}[!t]
  \centering
  \includegraphics[page=1,width=\textwidth]{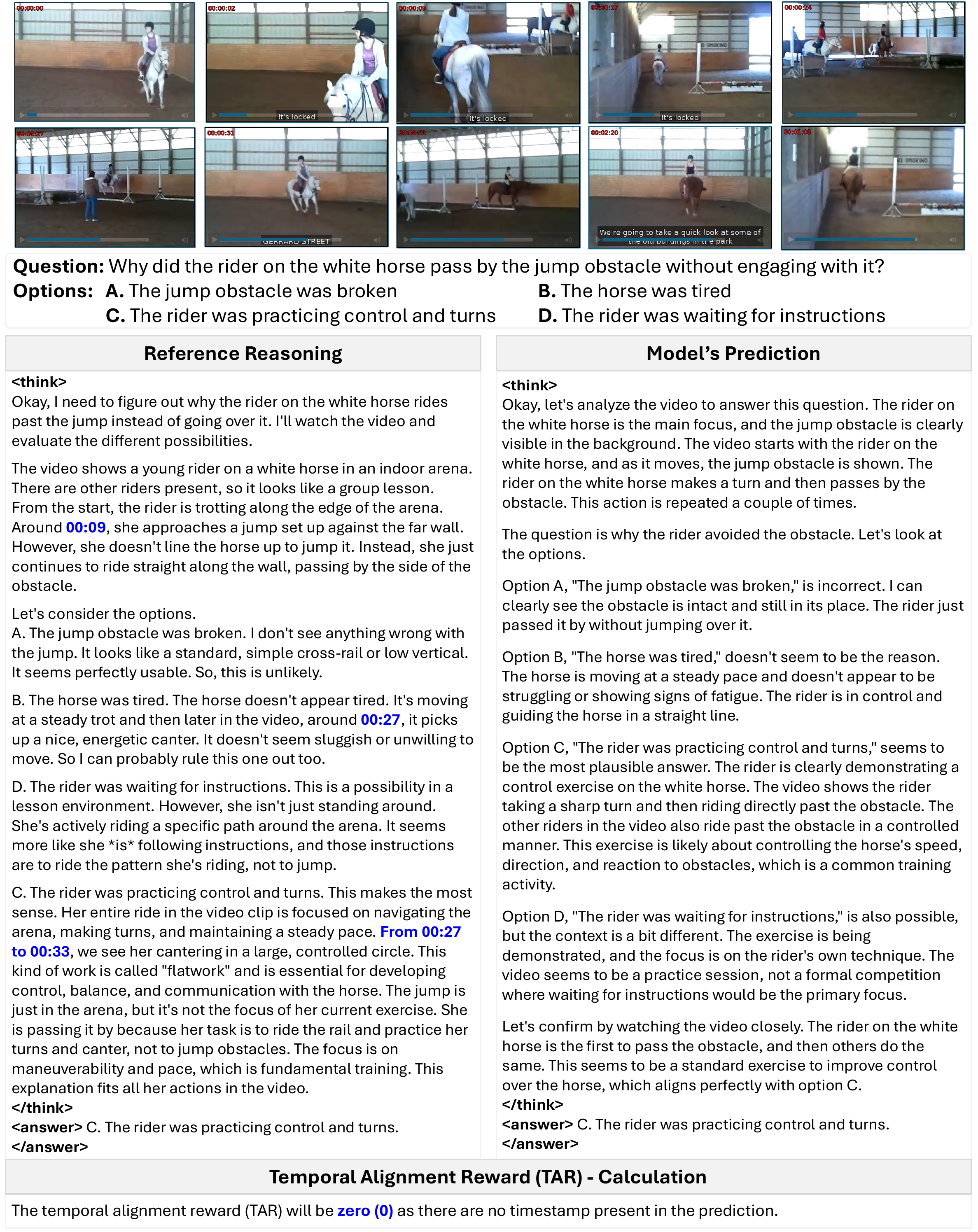}
  \caption{TAR calculation Example 7.}
  \label{fig:tar_ex8}
\end{figure*}

\begin{figure*}[!t]
  \centering
  \includegraphics[page=1,width=\textwidth]{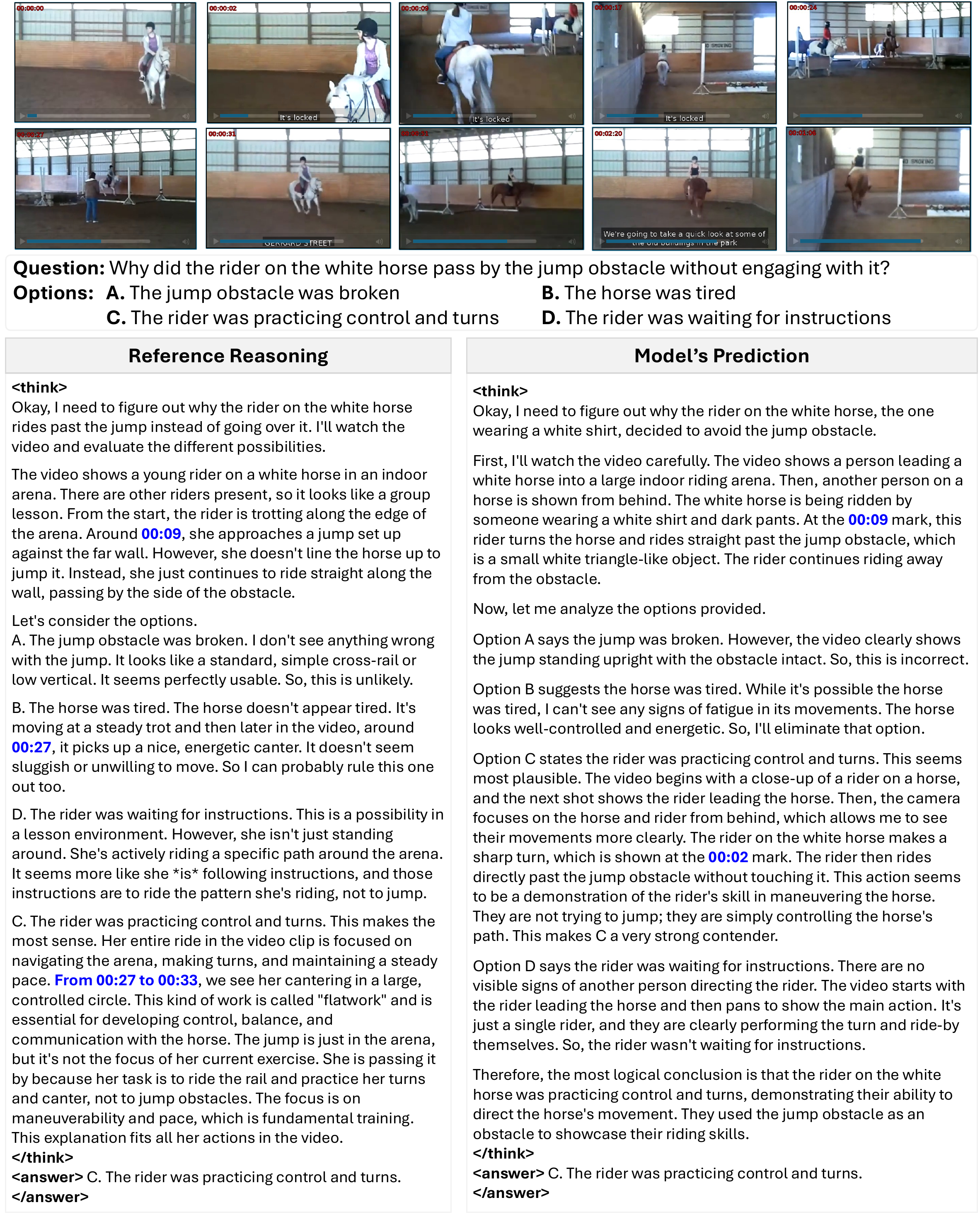}
  \caption{TAR calculation Example 8 (page 1 of 2).}
  \label{fig:tar_ex9_p1}
\end{figure*}

\begin{figure*}[!t]
  \centering
  \includegraphics[page=1,width=\textwidth]{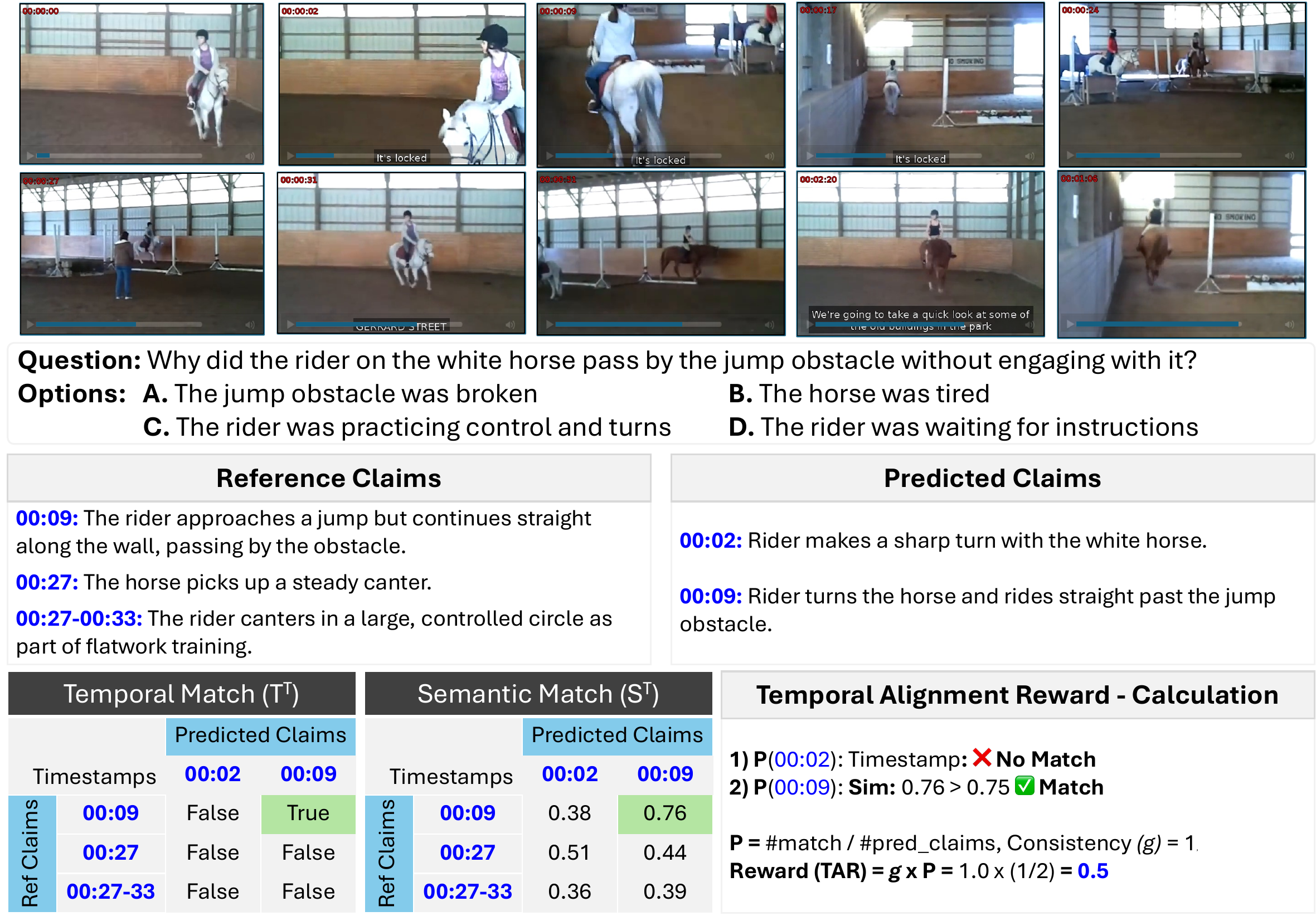}
  \caption{TAR calculation Example 8 (page 2 of 2).}
  \label{fig:tar_ex9_p2}
\end{figure*}

Figs.~\ref{fig:tar_ex2}-\ref{fig:tar_ex9_p1} show end-to-end TAR computation (using candidate generations during GRPO), including claim extraction, temporal and semantic matching, and reward calculation. We show diverse cases. 
% from clear matches to partial or failed alignment. 
Brief summaries:
\begin{itemize}
  \item \textbf{Example 1 (Fig.~\ref{fig:tar_ex2}):} Counting cartwheels. Consistent reasoning with one match and one miss.
  \item \textbf{Example 2 (Fig.~\ref{fig:tar_ex3}):} Counting cartwheels. Temporal mismatch, low similarity score.
  \item \textbf{Example 3 (Fig.~\ref{fig:tar_ex4}):} OCR on spatula text (“ELITE”). Consistent reasoning, perfect alignment.
  \item \textbf{Example 4 (Fig.~\ref{fig:tar_ex5}):} OCR on spatula text (“ELITE”). OCR error and TAC gate failure, TAR is zero.
  \item \textbf{Example 5 (Fig.~\ref{fig:tar_ex6_p1}-\ref{fig:tar_ex6_p2}):} Ribbon-adjustment tutorial. Partial alignment from one matched step.
  \item \textbf{Example 6 (Fig.~\ref{fig:tar_ex7_p1}-\ref{fig:tar_ex7_p2}):} Ribbon-adjustment tutorial. Inconsistent reasoning, no alignment and gated to zero.
  \item \textbf{Example 7 (Fig.~\ref{fig:tar_ex8}):} Horse riding practice. No predicted timestamp.
  \item \textbf{Example 8 (Fig.~\ref{fig:tar_ex9_p1}-\ref{fig:tar_ex9_p2}):} Horse riding practice. Partial alignment with one match and one miss.
\end{itemize}

% -------- Example 2 (single page) --------

% -------- Example 3 (single page) --------

% -------- Example 4 (single page) --------

% -------- Example 5 (single page) --------

% -------- Example 6 (two pages) --------

% -------- Example 7 (two pages) --------

% -------- Example 8 (single page) --------

% -------- Example 9 (two pages) --------

%% file: supplemental/dataset.tex
\section{Dataset}
\label{supp:dataset}

\subsection{Dataset Prompts}
\label{supp:dataset_prompts}

\paragraph{Prompt for Difficulty Ranking:}~To identify samples requiring deeper temporal reasoning, we assign a difficulty score to each video QA pair based on the complexity of its assistant response. 
The complete prompt used for this difficulty evaluation is provided in {\color{blue}\texttt{Prompt}~\textbf{F}}.

\begin{tcolorbox}[
  breakable,
  colback=gray!5!white,
  colframe=gray!60!black,
  title={\textbf{Prompt~F:} Dataset - Prompt for Difficulty Ranking},
  width=\linewidth,
  before skip=4pt,after skip=6pt,
  boxsep=4pt,left=2pt,right=2pt,top=2pt,bottom=2pt
]
\small
You are an expert model evaluator. Evaluate how difficult it would be for a vision-language model 
to generate the assistant responses in the following conversation.\\[4pt]

{\color{blue}The conversation is:}\\
\texttt{User: <user message>}\\
\texttt{Assistant: <assistant message>}\\[4pt]

Now, analyze the assistant responses:\\
Write the difficulty score in the format:\\
\texttt{**Difficulty score: X**}, where \texttt{X} is a number from 1 (easy) to 10 (very hard).
\end{tcolorbox}

\paragraph{Prompt for Reasoning Generation}~To regenerate timestamp-aware reasoning traces for each question-answer pair, we use the Gemini-2.5-Pro~\cite{comanici2025gemini} model to produce reasoning traces that explicitly reference key temporal cues. 
The model follows the instructions provided in {\color{blue}\texttt{Prompt}~\textbf{G}}, which ensures consistent reasoning style and timestamp annotation across the dataset.

\begin{tcolorbox}[
  breakable,
  colback=gray!5!white,
  colframe=gray!60!black,
  title={\textbf{Prompt~G:} Dataset - Prompt for Reasoning Generation},
  width=\linewidth,
  before skip=4pt,after skip=6pt,
  boxsep=4pt,left=2pt,right=2pt,top=2pt,bottom=2pt
]
\small
\textbf{Instructions}\\
Adopt the persona of someone carefully thinking through the following video-based question. Your goal is to show your reasoning process as a natural, internal monologue.\\

\textbf{Formatting Rules:}\\
\begin{enumerate}
    \item \textbf{Thinking Block:} Write your entire thought process within \texttt{<think>} and \texttt{</think>} tags. It should sound like you're talking to yourself to figure out the answer. Use conversational phrases like ``Okay, let's see...", ``Hmm, that's interesting," ``Wait a minute...", ``That makes sense because..."
    \item \textbf{NO LISTS:} Your thought process must be written in prose (a flowing paragraph or two). **Do not use numbered lists, bullet points, or a formal step-by-step breakdown.** This should feel like a stream of consciousness.
    \item \textbf{Answer Block:} After the \texttt{<think>} block, provide the final, concise answer inside \texttt{<answer>} and \texttt{</answer>} tags.
    \item \textbf{Key Frame Block: }After the answer, include a \texttt{<key\_frame>} block where you point out the specific timestamp(s) of the critical video frames without which the question cannot be answered. The timestamps should be as precise as possible (e.g., ``00:12", ``01:05–01:08"). If no key frame is required, leave the block empty. \\
\end{enumerate}

\textbf{Here is your question:}\\
{\color{blue}{\texttt{<question>}}}\\

\textbf{The final answer is provided below.} Use it as a reference to guide your thinking process, ensuring it logically leads to the correct conclusion. Also, make sure to highlight the key frame(s) that were essential for answering.\\
{\color{blue}{\texttt{<answer>}}}
\end{tcolorbox}

% % % % % % % % % % % %  Dataset Quality % % % % % % % % % % % % 
\subsection{Dataset Quality Analysis}
\label{supp:dataset_quality}

The dataset used for training~\MyModel~combines samples from five public video QA datasets: {LLaVA-Video-178K}~\cite{zhang2024video}, {NeXtQA}~\cite{xiao2021nextqa}, {ActivityNet-QA}~\cite{yu2019activitynet}, {PerceptionTest}~\cite{patraucean2023perception_test}, and {Clevrer}~\cite{yi2019clevrer}. 
It contains a total of 15{,}271 samples collected from 11{,}816 unique videos, annotated to provide timestamp-aware grounded supervision for both the SFT and GRPO stages.

\noindent
\textbf{Data verification.}~To ensure high quality of the data, the verification was performed in two stages. 
First, all samples with $TAC=0$ were automatically discarded. 
Second, a randomly selected subset of 500 samples (approximately 3\% of the total) was manually verified, checking both the correctness of the reasoning and its agreement with the provided answer. 
The goal of this verification was to assess reasoning quality of the data, focusing specifically on the correctness and answer consistency.

\noindent
\textbf{Timestamp accuracy.}
During early inspection, we find that the predicted timestamps sometimes do not match the video.
To fix this issue, we print timestamps directly on the video frames during reasoning regeneration, which greatly improves temporal precision.
After printing timestamps, small offsets still appear in some cases, usually within ±2 seconds of the real event in the video. These small differences are within the acceptable time range and do not affect the effectiveness of temporal supervision (see Tab.~\ref{tab:ablation_components} of the main paper for the ablation on SFT and GRPO).

\noindent
\textbf{Overall quality.}~While full manual verification of all 15,271 samples is not possible at our scale, the verified subset shows high reasoning accuracy and consistent temporal alignment.
Including ground truth final answers during reasoning generation reduces possible annotation noise, and TAC-based filtering further ensures strong internal consistency across the dataset.
Although small ambiguities may remain, they have little effect on downstream performance, as shown by steady improvements in TAC, VAS, and accuracy during model training (see Fig.~\ref{fig:tac_vas_results}, Tab.~\ref{tab:sota_table} and Tab.~\ref{tab:ablation_components} of the main paper for the quantitative results).
% Overall, the dataset provides a reliable and balanced base for timestamp-aware video reasoning.

% % % % % % % % % % % %  Dataset Quality % % % % % % % % % % % % 

% % % % % % % % % % % %  Dataset Examples % % % % % % % % % % % % 
\subsection{Dataset Examples}
\label{supp:dataset_examples}

\begin{figure*}[!t]
  \centering
  \includegraphics[width=\textwidth]{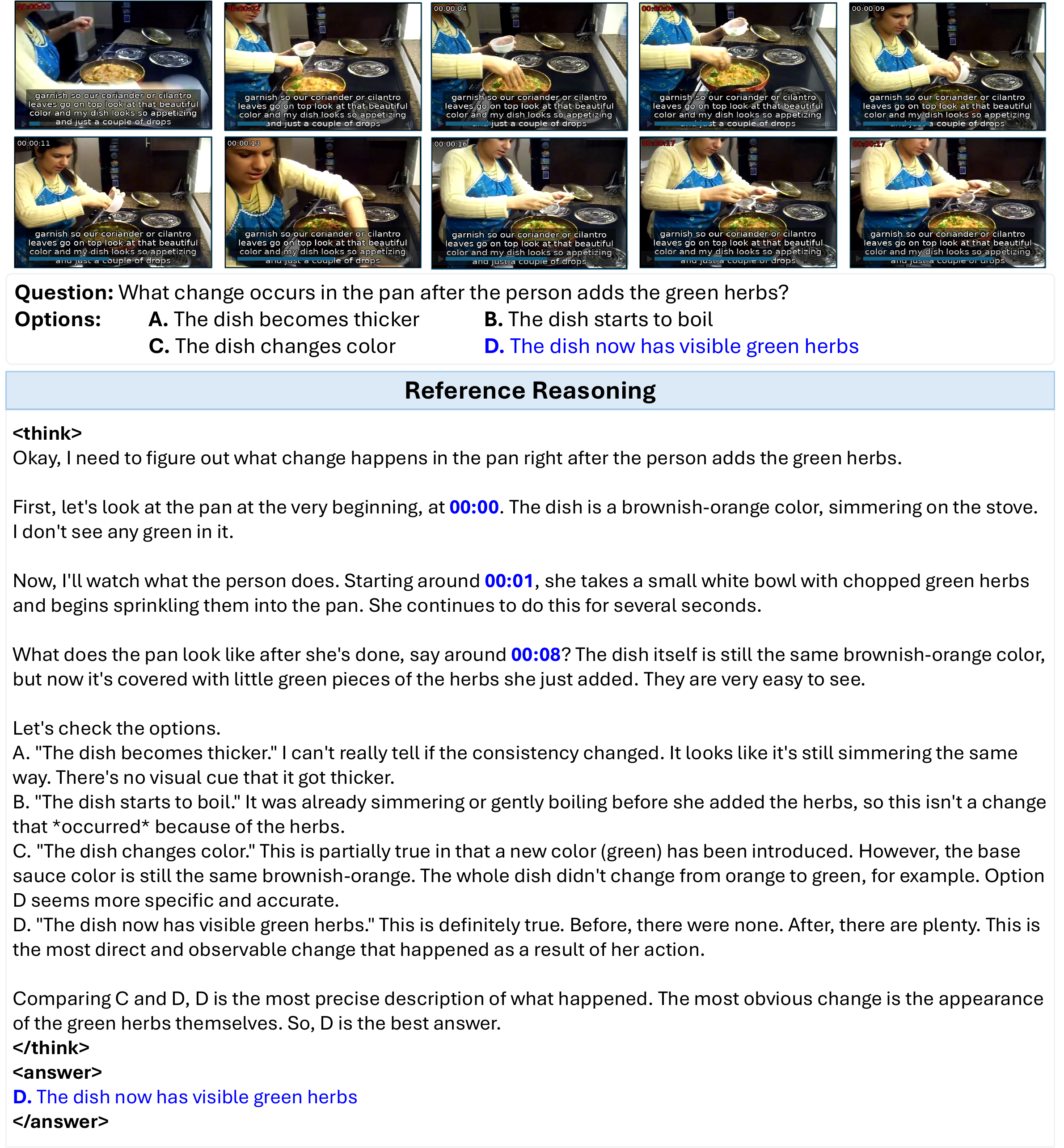}
  \caption{Dataset Example 1}
  \label{fig:dataset_ex1}
\end{figure*}

\begin{figure*}[!t]
  \centering
  \includegraphics[width=\textwidth]{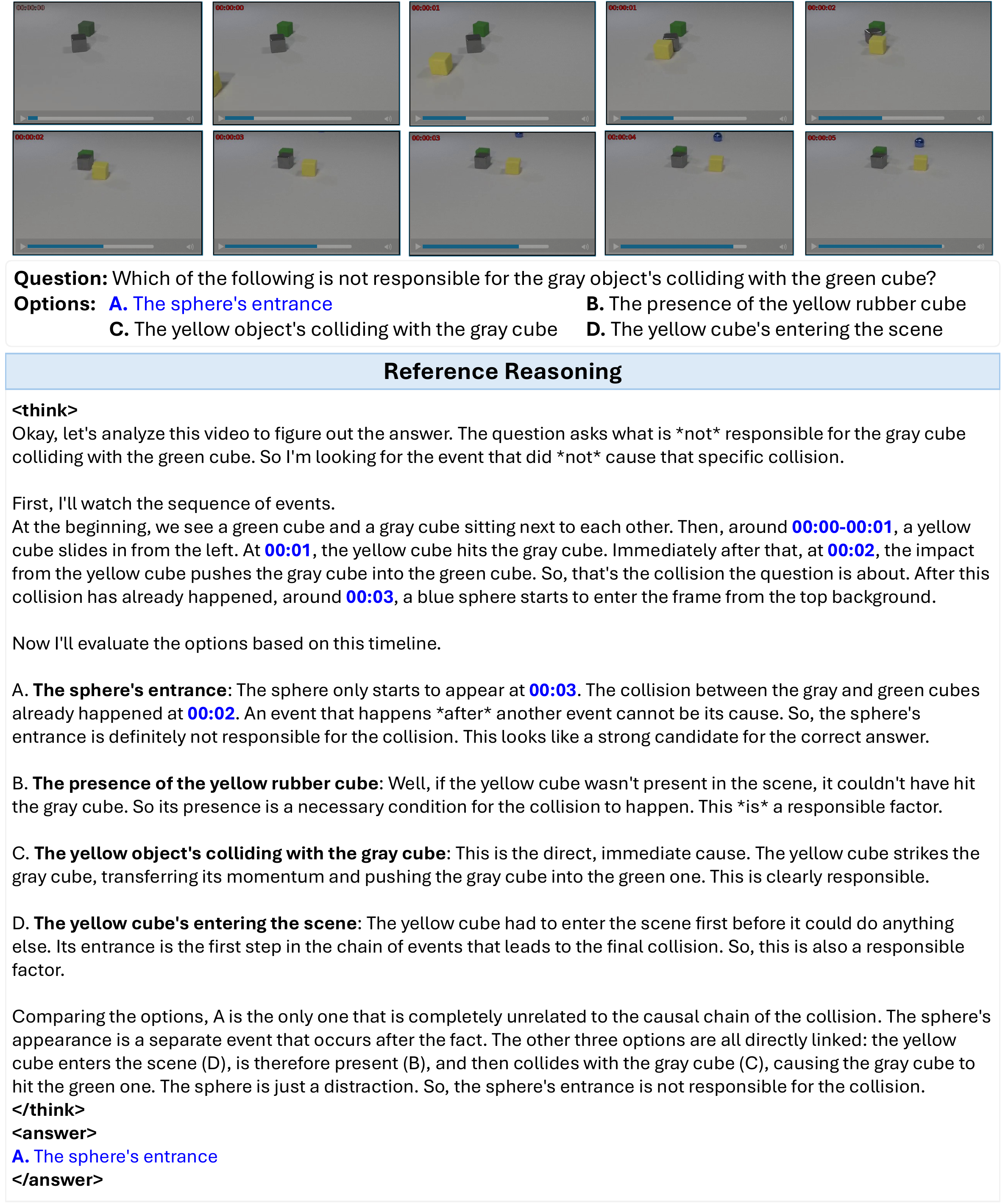}
  \caption{Dataset Example 2}
  \label{fig:dataset_ex2}
\end{figure*}

\begin{figure*}[!t]
  \centering
  \includegraphics[width=\textwidth]{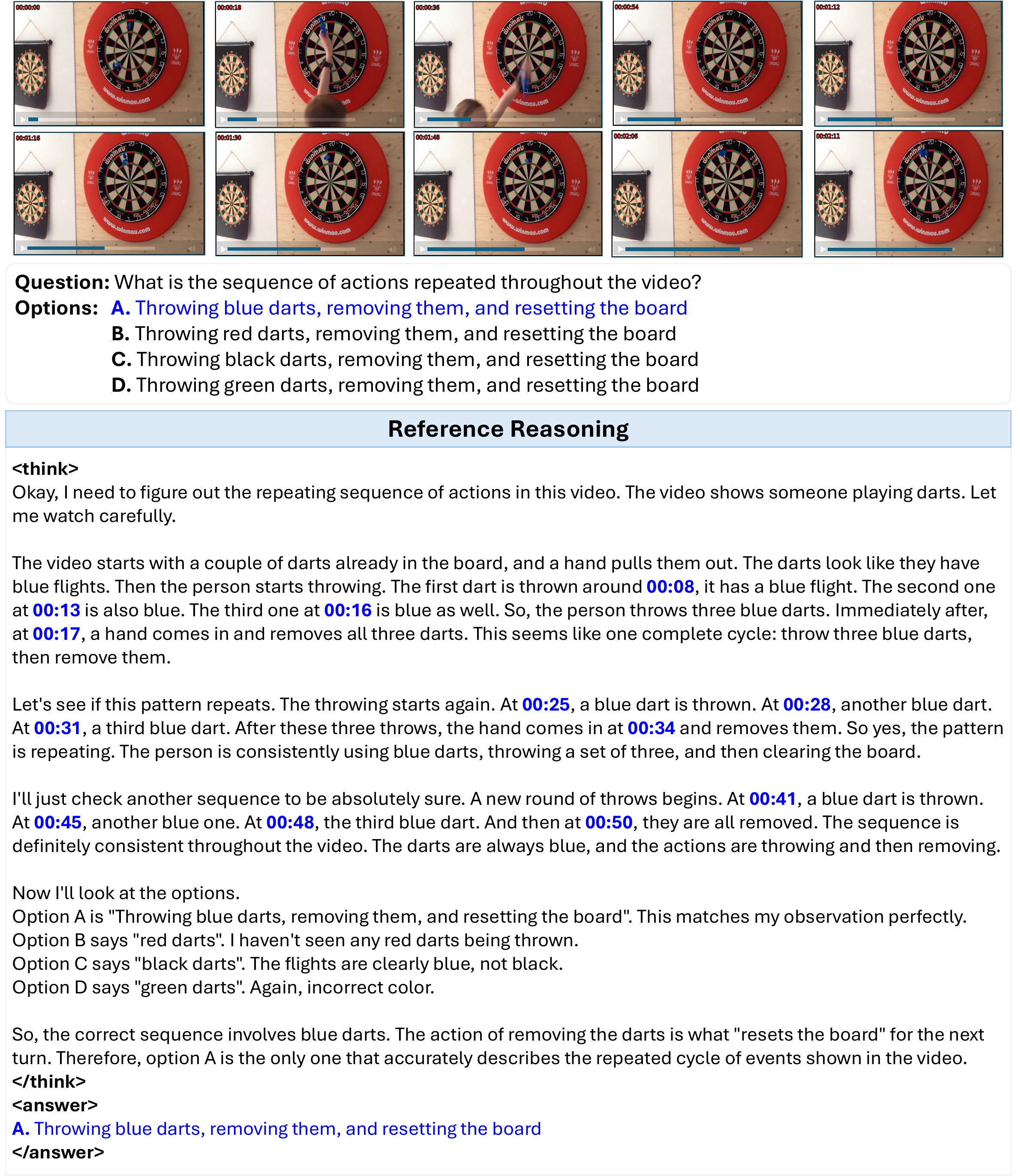}
  \caption{Dataset Example 3}
  \label{fig:dataset_ex3}
\end{figure*}

Figs.~\ref{fig:dataset_ex1}-\ref{fig:dataset_ex3} show a few samples from the dataset, covering both simple and complex temporal reasoning cases. 
Each visualization shows the video context, question, options and generated reference reasoning.

\begin{itemize}
  \item \textbf{Example 1 (Fig.~\ref{fig:dataset_ex1}):} Cooking video with fine-grained temporal reasoning on appearance change after adding ingredients.
  \item \textbf{Example 2 (Fig.~\ref{fig:dataset_ex2}):} Causal reasoning in a synthetic simulation, identifying which event did not contribute to a collision.
  \item \textbf{Example 3 (Fig.~\ref{fig:dataset_ex3}):} Reasoning across repeated motion cycles (throwing and removing darts) with strong temporal alignment.
\end{itemize}

% % % % % % % % % % % %  Dataset Examples % % % % % % % % % % % % 

% % % % % % % % % % % %  Dataset Stats % % % % % % % % % % % % 
%%%%%%%%%%%%%  Dataset Stats Table %%%%%%%%%%%%%
\begin{table*}[!t]
\centering
\setlength{\tabcolsep}{0.75em}
\begin{adjustbox}{max width=\textwidth}
\begin{tabular}{
  l|                 % Split
  *{5}{c}|           % Core stats
  *{2}{c}|           % Resolution
  *{5}{c}            % Sources
}
\toprule
&
\multicolumn{5}{c|}{\textbf{Core Stats}} &
\multicolumn{2}{c|}{\textbf{Resolution}} &
\multicolumn{5}{c}{\textbf{Datasets Source}} \\
\cmidrule(lr){2-6}\cmidrule(lr){7-8}\cmidrule(lr){9-13}
\raisebox{-12ex}{\textbf{Split}} &
\rothead{Samples} &
\rothead{Unique Videos} &
\rothead{Avg. Duration (s)} &
\rothead{Avg. Reasoning Words} &
\rothead{Avg. Reasoning Timestamps} &
\rothead{$<720$ p} &
\rothead{$\geq 720$ p} &
\rothead{LLaVA-Video-178K} &
\rothead{NeXtQA} &
\rothead{ActivityNet-QA} &
\rothead{PerceptionTest} &
\rothead{Clevrer} \\
\midrule
SFT &
10{,}467 & 8{,}621 & 47.3 & 328.6 & 2.54 &
5{,}993 & 2{,}628 &
4{,}635 & 1{,}185 & 2{,}553 & 520 & 1{,}574 \\
GRPO &
4{,}804 & 4{,}099 & 57.8 & 305.8 & 2.22 &
2{,}493 & 1{,}606 &
2{,}198 & 614 & 1{,}328 & 472 & 192 \\
\midrule
Combined &
15{,}271 & 11{,}816 & 48.9 & 321.4 & 2.44 &
7{,}977 & 3{,}839 &
6{,}833 & 1{,}799 & 3{,}881 & 992 & 1{,}766 \\
\bottomrule
\end{tabular}
\end{adjustbox}
\vspace{-0.5em}
\caption{
\textbf{Dataset statistics for SFT, GRPO, and combined.}~We report the number of samples, unique videos, average video duration, average reasoning length in words, and average number of timestamps per reasoning trace. Resolution counts are split below and at least 720p.
The last five columns list per source counts for LLaVA-Video-178K, NeXtQA, ActivityNet-QA, PerceptionTest, and Clevrer.
}
\label{tab:dataset_stats}
\vspace{-1em}
\end{table*}
%%%%%%%%%%%%%  Dataset Stats Table %%%%%%%%%%%%%

\subsection{Dataset Stats}
\label{supp:dataset_stats}

Tab.~\ref{tab:dataset_stats} provides a breakdown of the stats for the Supervised Fine-Tuning (SFT) and Group Relative Policy Optimization (GRPO) subsets of our dataset.~The overally dataset contains 15,271 video-question-answer samples collected from five public sources: LLaVA-Video-178K~\cite{zhang2024video}, NeXtQA~\cite{xiao2021nextqa}, ActivityNet-QA~\cite{yu2019activitynet}, PerceptionTest~\cite{patraucean2023perception_test}, and Clevrer~\cite{yi2019clevrer}.

% Each subset maintains diverse temporal lengths, video resolutions, and reasoning complexities. 
% We also report the average number of timestamps per reasoning trace to highlight the degree of temporal grounding across samples.

\noindent
Across all splits, the dataset maintains balanced coverage of video lengths and resolutions, with roughly one-third of videos above 720p. 
The average reasoning trace spans around 320~words, indicating detailed yet concise temporal reasoning. The presence of multiple datasets ensures diversity in domain and task type, facilitating generalization across multiple reasoning tasks.

% % % % % % % % % % % %  Dataset Stats % % % % % % % % % % % % 

%% file: supplemental/comparison_with_prior_methods.tex
% % % % % % % % % % % % %  SoTA Comparison Table (DATA) % % % % % % % % % % % % %
\begin{table*}[!t]
\centering
\setlength{\tabcolsep}{0.75em}
\begin{adjustbox}{max width=\textwidth}
\begin{tabular}{
  l|                          % Model
  *{3}{c} c|                  % Left panel: SFT, RL, Combined, Annotation Type
  c c >{\columncolor{gray!10}}c|   % TAC: Avg Gen, Avg Reas, Avg Overall (shaded)
  c c >{\columncolor{gray!10}}c|   % VAS: Avg Gen, Avg Reas, Avg Overall (shaded)
  c c >{\columncolor{gray!10}}c    % Acc: Avg Gen, Avg Reas, Avg Overall (shaded)
}
\toprule
&
\multicolumn{4}{c|}{\textbf{Dataset Info}} &
\multicolumn{3}{c|}{\textbf{TAC}} &
\multicolumn{3}{c|}{\textbf{VAS}} &
\multicolumn{3}{c}{\textbf{Accuracy}} \\
\cmidrule(lr){2-5}\cmidrule(lr){6-8}\cmidrule(lr){9-11}\cmidrule(lr){12-14}
\raisebox{-12ex}{\textbf{Model}} &
\rothead{SFT Dataset Size} &
\rothead{RL Dataset Size} &
\rothead{Combined Dataset Size} &
\rothead{Reasoning Annotation Type} &
\rothead{{Avg. (Generic)}} &
\rothead{{Avg. (Reasoning)}} &
\rothead{\textbf{Avg. (Overall)}} &
\rothead{{Avg. (Generic)}} &
\rothead{{Avg. (Reasoning)}} &
\rothead{\textbf{Avg. (Overall)}} &
\rothead{{Avg. (Generic)}} &
\rothead{{Avg. (Reasoning)}} &
\rothead{\textbf{Avg. (Overall)}} \\
\midrule

% ------------------- Data rows -------------------
Video-R1~\cite{videor12025} &
165K & 260K & 260K & \textbf{A} &
79.8 & 68.5 & 73.6 &
53.2 & 39.9 & 45.9 &
\underline{65.6} & 38.8 & 51.0 \\

VideoChat-R1~\cite{videochatR12025} &
- & 18K & - & - &
77.9 & 75.9 & 76.8 &
57.3 & 42.0 & 48.9 &
63.8 & 39.1 & 50.3 \\

VideoChat-R1.5~\cite{yan2025videochat_r1.5} &
- & 80K & 80K & \textbf{A} &
79.9 & \underline{76.9} & 78.2 &
56.7 & 42.5 & 48.9 &
64.7 & \underline{39.8} & 51.1 \\

VideoRFT~\cite{videorft2025} &
102K & 310K & 412K & \textbf{A} &
\underline{84.1} & 75.5 & \underline{79.5} &
\underline{69.8} & \underline{53.2} & \underline{60.8} &
65.3 & 39.7 & \underline{51.3} \\

\midrule

\textbf{\MyModel~(Ours)} &
10.4K & 4.8K & 15.3K & \textbf{A+M} &
\textbf{85.7} & \textbf{78.2} & \textbf{81.6} &
\textbf{78.9} & \textbf{61.5} & \textbf{69.4} &
\textbf{66.7} & \textbf{41.6} & \textbf{53.0} \\

\bottomrule
\end{tabular}
\end{adjustbox}
\vspace{-0.5em}
\caption{
\textbf{Training dataset comparison with prior video reasoning models.}
All baselines rely primarily on large-scale, automatically generated (\textbf{A}) reasoning datasets, whereas our dataset (\textbf{A+M}) combines automatic annotation with partial manual verification of approximately 3\% of the total samples (500 out of 15{,}271). 
Despite being trained on an order of magnitude fewer samples,~\MyModel~achieves higher TAC, VAS, and accuracy across both generic and reasoning benchmarks, indicating that dataset quality and consistency contribute more strongly to reasoning performance than raw dataset scale. 
Dashes (\textbf{-}) indicate unavailable information or unused stage.
}
\label{supp_tab:sota_comparison_data}
\vspace{-1em}
\end{table*}
% % % % % % % % % % % % %  SoTA Comparison Table (DATA) % % % % % % % % % % % % %

\section{Comparison with Prior Methods}
\label{supp:comparison_with_prior_methods}

\subsection{Training Dataset Based Comparison}
\label{supp:training_dataset_comparison}

Tab.~\ref{supp_tab:sota_comparison_data} compares our dataset and results with prior video reasoning models.
Unlike previous works that depend on much larger training sets (about 100K-400K samples), our model reaches higher TAC, VAS, and accuracy while using only 15.2K samples in total.
This shows that data quality and temporal supervision are more important than dataset size.
Our dataset (\textbf{A+M}) combines automatically generated reasoning annotations with manual checks for about 3\% of all samples, making it more reliable.
Using timestamp-aware reasoning and TAC-based filtering also improves consistency, allowing more effective learning from a smaller dataset.
In contrast, earlier models mainly use automatically generated (\textbf{A}) reasoning traces, which can lead to inconsistency or weak visual grounding.

%% file: supplemental/novelty.tex
\begin{table*}[!b]
\centering
\small
\begin{tabular}{l|ccc}
\toprule
\textbf{Method} & \textbf{Reward Type} & \textbf{Granularity} & \textbf{Consistency Handling} \\
\midrule
Video-R1~\cite{videor12025} & T-GRPO (ordered vs. shuffled) & Sequence & Answer-gated \\
VideoChat-R1~\cite{videochatR12025} & Task-verifiable (IoU / Acc) & Task-level & None \\
VideoChat-R1.5~\cite{yan2025videochat_r1.5} & VTTS Iterative Perception & Inference-level & None \\
VideoRFT~\cite{videorft2025} & Semantic Consistency (SigLIP) & Description-span & Answer-gated \\
\midrule
\textbf{Video-R2 (Ours)} & \textbf{TAR (Precision-based)} & \textbf{Claim-level} & \textbf{TAC-gated} \\
\bottomrule
\end{tabular}
\vspace{-0.5em}
\caption{
\textbf{Comparison of reward formulation and consistency design across recent video reasoning models.}
Unlike previous approaches that focus on task-specific or semantic alignment rewards, our method introduces a novel claim-level precision-based temporal alignment reward combined with TAC-gated consistency enforcement, providing fine-grained supervision of reasoning alignment.}
\label{supp:novelty_comparison}
\vspace{-1em}
\end{table*}

\section{Detailed Discussion on Novelty}
\label{supp:novelty}

\noindent
\textbf{Comparison with Video-R1.}~Video-R1~\cite{videor12025} employs a two-stage training pipeline similar to ours, consisting of supervised fine-tuning (SFT) followed by group relative policy optimization (GRPO). Its key novelty lies in the temporal-GRPO (T-GRPO), which compares model performance on ordered versus shuffled frames. While this provides a proxy for ensuring that the model explores the temporal sequence present in the video frames, no direct reinforcement is applied on the reasoning traces. In contrast, our approach introduces the Temporal Alignment Reward (TAR), which operates at the claim level by aligning timestamped reasoning to reference claims with consistency gating, resulting in fine-grained temporal credit assignment. 

\noindent
\textbf{Comparison with VideoChat-R1.}~VideoChat-R1~\cite{videochatR12025} focuses on improving video perception through reinforcement fine-tuning with GRPO. Its reward functions are task-specific and are all verifiable, such as intersection-over-union (IoU) for temporal range grounding, accuracy for classification, and recall for captioning. Although this yields strong perceptual grounding, it does not evaluate the reasoning process of the model. Our method differs by directly rewarding temporally aligned reasoning through predicted and reference claims, offering an explicit link between reasoning steps and video content.

\noindent
\textbf{Comparison~with~VideoChat-R1.5.}~VideoChat-R1.5~\cite{yan2025videochat_r1.5}~introduces Visual Test-Time Scaling (VTTS) and Iterative Perception, where the model iteratively refines its output at test time. The improvement primarily comes from the test-time scaling, not from reward formulation. Unlike VideoChat-R1.5, our method strengthens temporal reasoning during training via a temporal alignment reward and a consistency gating, leading to more coherent and grounded reasoning traces.

\noindent
\textbf{Comparison with VideoRFT.}~VideoRFT~\cite{videorft2025} introduces a semantic consistency reward that aligns generated video descriptions with visual embeddings using SigLIP~\cite{zhai2023sigmoid}.
It works at a coarse level of text-video similarity and focuses on visual matching rather than detailed reasoning or timing accuracy.
In contrast, our method directly supervises the reasoning process by aligning timestamped claims in the reasoning text with reference claims through the proposed Temporal Alignment Reward (TAR).
Further, instead of conditioning on answer correctness as in VideoRFT, our gating mechanism is based on Think--Answer Consistency (TAC), which enforces logical agreement between reasoning and answer before applying any temporal reward.
This design focuses learning on fine-grained temporal reasoning instead of broad video-text similarity, representing a conceptually distinct reinforcement formulation.

\noindent
\textbf{Summary.}~Overall, \textbf{\MyModel}~unifies timestamp-aware supervision and reinforcement fine-tuning through a precision-based Temporal Alignment Reward (TAR) combined with a consistency gate $(g)$. 
Unlike prior works that focus on perceptual or embedding-level correspondence, our method explicitly rewards temporally accurate and logically consistent reasoning at the claim level. 
This joint supervision reinforces both the temporal correspondence and internal coherence of the reasoning process. 
To the best of our knowledge, \textbf{\MyModel}~is the first to employ consistency-gated temporal alignment as a reinforcement objective for video reasoning, offering a path toward more interpretable and causally grounded multimodal reasoning.
A summary of comparison with recent video reasoning methods is provided in Tab.~\ref{supp:novelty_comparison} for completeness.

%% file: supplemental/additonal_ablations.tex
%%%%%%%%%%%%%  Supp: Ablations (Stronger Consistency & Recall Term) %%%%%%%%%%%%%
\begin{table*}[!t]
\centering
\setlength{\tabcolsep}{0.75em}
\begin{adjustbox}{max width=\textwidth}
\begin{tabular}{
  l|                               % Model
  *{5}{c}                          % 5 Generic benches
  |>{\columncolor{gray!10}}c|      % Avg Generic (boxed by vertical rules)
  *{6}{c}                          % 6 Reasoning benches
  |>{\columncolor{gray!10}}c|      % Avg Reasoning (boxed)
  >{\columncolor{gray!10}}c        % Avg Overall (boxed; right border too)
}
\toprule
% Group headers (span includes averages)
 &
\multicolumn{6}{c|}{\textbf{Generic Benchmarks}} &
\multicolumn{7}{c|}{\textbf{Reasoning Benchmarks}} &
\\
\cmidrule(lr){2-7}\cmidrule(lr){8-14}
% Rotated, bottom-aligned column headers (including averages)
\raisebox{-15ex}{\textbf{Model}} &
\rothead{MVBench} &
\rothead{VideoMME} &
\rothead{TempCompass} &
\rothead{MLVU} &
\rothead{LongVideoBench} &
\rothead{\textbf{Avg. (Generic)}} &
\rothead{VideoMathQA} &
\rothead{Video-MMMU} &
\rothead{MMVU-Val} &
\rothead{VSiBench} &
\rothead{MINERVA} &
\rothead{SciVideoBench} &
\rothead{\textbf{Avg. (Reasoning)}} &
\rothead{\textbf{Avg. (Overall)}} \\
\midrule
% ------------------- Data rows -------------------
\multicolumn{15}{l}{\textbf{Accuracy}} \\

% \rowcolor{gray!10}
\MyModel~(Ours) &
\textbf{67.5} & \textbf{63.8} & \textbf{74.9} & \underline{68.3} & \textbf{59.2} & \textbf{66.7} & \textbf{28.8} & \textbf{50.8} & \textbf{67.4} & \textbf{39.4} & \underline{34.9} & \underline{28.4} & \textbf{41.6} & \textbf{53.0} \\

$\rightarrow$ Stronger Consistency Reward &  
62.9 & 60.3 & 71.9 & 63.7 & 54.2 & 62.6 & 20.0 & \underline{49.3} & 64.3 & 35.7 & 33.5 & 26.0 & 38.1 & 49.3 \\

$\rightarrow$ Recall Term in TAR &  
\underline{67.0} & \underline{62.9} & \underline{74.2} & \textbf{68.6} & \underline{57.4} & \underline{66.0} & \underline{22.6} & 47.9 & \underline{66.9} & \underline{37.9} & \textbf{35.5} & \textbf{28.5} & \underline{39.9} & \underline{51.8} \\

\midrule

\multicolumn{15}{l}{\textbf{Think--Answer Consistency (TAC)}} \\

% \rowcolor{gray!10}
\MyModel~(Ours) &
82.4 & 86.6 & 88.3 & 86.1 & 85.1 & 85.7 & 63.7 & 84.6 & \underline{90.2} & 78.4 & 75.1 & \underline{77.3} & 78.2 & 81.6 \\

$\rightarrow$ Stronger Consistency Reward &  
\textbf{98.7} & \textbf{98.3} & \textbf{99.0} & \textbf{98.9} & \textbf{99.0} & \textbf{98.8} & \textbf{96.1} & \textbf{99.0} & \textbf{98.5} & \textbf{98.1} & \textbf{98.5} & \textbf{98.0} & \textbf{98.0} & \textbf{98.4} \\

$\rightarrow$ Recall Term in TAR &  
\underline{89.6} & \underline{88.9} & \underline{88.6} & \underline{89.0} & \underline{89.3} & \underline{89.1} & \underline{77.8} & \underline{92.0} & 86.8 & \underline{85.4} & \underline{82.3} & 73.7 & \underline{83.0} & \underline{85.8} \\

\midrule

\multicolumn{15}{l}{\textbf{Video Attention Score (VAS)}} \\

% \rowcolor{gray!10}
\MyModel~(Ours) &
76.5 & 80.1 & 86.1 & \underline{78.7} & \underline{73.2} & \underline{78.9} & 39.6 & 49.1 & 72.5 & 74.2 & \underline{73.4} & 60.5 & 61.5 & 69.4 \\

$\rightarrow$ Stronger Consistency Reward &  
\textbf{83.1} & \textbf{83.5} & \textbf{88.8} & \textbf{85.3} & \textbf{81.5} & \textbf{84.4} & \underline{43.4} & \textbf{51.9} & \textbf{78.0} & \textbf{79.7} & \textbf{78.0} & \textbf{65.9} & \textbf{66.2} & \textbf{74.5} \\

$\rightarrow$ Recall Term in TAR &  
\underline{78.2} & \underline{80.5} & \underline{86.2} & 76.8 & 72.7 & \underline{78.9} & \textbf{44.5} & \underline{50.8} & \underline{76.3} & \underline{76.9} & 72.9 & \underline{62.7} & \underline{64.0} & \underline{70.8} \\

% --- Add more rows below; keep the same column structure ---
% ModelName &  &  &  &  &  &  &  &  &  &  &  &  &  &  \\
\bottomrule
\end{tabular}
\end{adjustbox}
\vspace{-0.5em}
\caption{
\textbf{Ablations on stronger consistency reward and recall term in TAR.}
The first variant adds a TAC-based reward directly to the total reward instead of consistency gating, producing a nearly \textit{perfect TAC (98.4\%)} and \textit{strong visual grounding (VAS)} but reduced accuracy. 
This highlights a trade-off similar to the \textit{SFT consistency paradox}, where optimizing for internal coherence alone yields a highly consistent but not necessarily stronger model. 
The second variant introduces a recall term in the Temporal Alignment Reward, computing \( \mathrm{TAR}_{F1} \) instead of precision-only TAR. 
While this slightly enhances visual attention (VAS), accuracy drops especially on the reasoning benchmarks, as overgeneration of timestamps is encouraged. 
These experiments confirm that the precision-based, TAC-gated TAR in \textbf{\MyModel} offers the most balanced optimization of accuracy, consistency (TAC), and visual grounding (VAS).
}
\label{supp:ablation_1_2}
% \vspace{-1em}
\end{table*}
%%%%%%%%%%%%%  Supp: Ablations (Stronger Consistency & Recall Term) %%%%%%%%%%%%%

\section{Additional Ablations}
\label{supp:additional_ablations}

\subsection{Effect of Stronger Consistency Reward Factor}
\label{supp:additional_ablations_stronger_consistency_reward}

To study the effect of stronger consistency-oriented optimization, we modify the reward function to explicitly include the \textbf{Think--answer consistency (TAC)} as an additional reward component rather than using it as a gating condition. 
In this configuration, the Temporal Alignment Reward is computed without the consistency gate (e.g. \(TAR_{\text{prec}}\)), and a separate TAC-based term is added to the total reward for GRPO:
\begin{align}
R_{\text{total}} = {} &
\lambda_{\text{acc}} R_{\text{acc}}
+ \lambda_{\text{fmt}} R_{\text{fmt}} \notag \\
& + \lambda_{\text{tar}} TAR_{\text{prec}}
+ \lambda_{\text{tac}} R_{\text{TAC}},
\end{align}
where \(R_{\text{TAC}}\in [0,1]\) directly measures the alignment between the model’s reasoning (\texttt{<think>...</think>}) and final answer (\texttt{<answer>...</answer>}).
Removing the gating allows the model to be explicitly rewarded for self-consistency, irrespective of the other reward components.

As shown in Tab.~\ref{supp:ablation_1_2}, this variant achieves \textit{nearly perfect TAC (98.4\%)} and the \textit{highest VAS} across all benchmarks, but at a clear cost to overall accuracy. 
This behavior reflects a strong parallel to the \textit{SFT consistency paradox} discussed in the main paper, where optimizing solely for internal coherence yields highly structured reasoning but not necessarily a stronger model. 
We attribute this to the model overemphasizing logical self-agreement while underexploring corrective reasoning signals. 
Nevertheless, we note that combining this stronger consistency reward with larger-scale data (on the order of hundreds of thousands or millions of samples) may produce highly consistent and accurate models: a promising direction for future exploration.

%%%%%%%%%%%%%  Supp: Table PCC %%%%%%%%%%%%%
\begin{table*}[!t]
\centering
\setlength{\tabcolsep}{0.75em}
\begin{adjustbox}{max width=\textwidth}
\begin{tabular}{
  l|                               % Model
  *{5}{c}                          % 5 Generic benches
  |>{\columncolor{gray!10}}c|      % Avg Generic (boxed by vertical rules)
  *{6}{c}                          % 6 Reasoning benches
  |>{\columncolor{gray!10}}c|      % Avg Reasoning (boxed)
  >{\columncolor{gray!10}}c        % Avg Overall (boxed; right border too)
}
\toprule
% Group headers (span includes averages)
 &
\multicolumn{6}{c|}{\textbf{Generic Benchmarks}} &
\multicolumn{7}{c|}{\textbf{Reasoning Benchmarks}} &
\\
\cmidrule(lr){2-7}\cmidrule(lr){8-14}
% Rotated, bottom-aligned column headers (including averages)
\raisebox{-15ex}{\textbf{Model}} &
\rothead{MVBench} &
\rothead{VideoMME} &
\rothead{TempCompass} &
\rothead{MLVU} &
\rothead{LongVideoBench} &
\rothead{\textbf{Avg. (Generic)}} &
\rothead{VideoMathQA} &
\rothead{Video-MMMU} &
\rothead{MMVU-Val} &
\rothead{VSiBench} &
\rothead{MINERVA} &
\rothead{SciVideoBench} &
\rothead{\textbf{Avg. (Reasoning)}} &
\rothead{\textbf{Avg. (Overall)}} \\
\midrule
% ------------------- Data rows -------------------
\multicolumn{15}{l}{\textbf{Pearson Correlation Coefficient (PCC)}} \\

Video-R1~\cite{videor12025} &
0.740 & 0.777 & 0.666 & 0.777 & 0.684 & 0.729 & 0.664 & 0.752 & 0.782 & 0.744 & 0.733 & 0.657 & 0.722 & 0.725 \\

VideoChat-R1~\cite{videochatR12025} &
0.689 & 0.768 & 0.690 & 0.779 & 0.745 & 0.734 & 0.762 & 0.707 & 0.753 & 0.756 & 0.819 & 0.728 & 0.754 & 0.745 \\

VideoChat-R1.5~\cite{yan2025videochat_r1.5} &
0.654 & 0.762 & 0.727 & 0.781 & 0.634 & 0.711 & 0.750 & 0.762 & 0.731 & 0.720 & 0.736 & 0.654 & 0.726 & 0.719 \\

VideoRFT~\cite{videorft2025} &
0.709 & 0.715 & 0.646 & 0.796 & 0.657 & 0.705 & 0.649 & 0.707 & 0.682 & 0.731 & 0.759 & 0.676 & 0.701 & 0.702 \\

\MyModel~(Ours) &
0.670 & 0.656 & 0.613 & 0.704 & 0.610 & 0.651 & 0.675 & 0.693 & 0.698 & 0.540 & 0.671 & 0.579 & 0.643 & 0.646 \\

\midrule

\rowcolor{gray!10}
\textbf{Average} &
\textbf{0.692} & \textbf{0.736} & \textbf{0.668} & \textbf{0.767} & \textbf{0.666} & \textbf{0.706} & \textbf{0.700} & \textbf{0.724} & \textbf{0.729} & \textbf{0.698} & \textbf{0.744} & \textbf{0.659} & \textbf{0.709} & \textbf{0.708} \\

% --- Add more rows below; keep the same column structure ---
% ModelName &  &  &  &  &  &  &  &  &  &  &  &  &  &  \\
\bottomrule
\end{tabular}
\end{adjustbox}
\vspace{-0.5em}
\caption{
\textbf{Stability of Video Attention Score (VAS) metric across different LLM judge models.}
Pearson Correlation Coefficient (PCC) between VAS scores computed using Qwen3-Next-80B-A3B (default) and Qwen3-32B across all benchmarks and reasoning models. 
Average PCC values exceed 0.7 in both generic and reasoning categories, confirming a strong and statistically significant linear correlation ($p$-values $=$ $10^{-35}$ -- 0.0). 
This demonstrates that VAS is stable with respect to the choice of LLM judge, and model rankings remain consistent across judge models. 
For full reproducibility and alignment with the reported results, we recommend using Qwen3-Next-80B-A3B as the default LLM judge for VAS computation.
}
\label{supp:table_vas_pcc}
\vspace{-1em}
\end{table*}
%%%%%%%%%%%%%  Supp: Table PCC %%%%%%%%%%%%%

\subsection{Introducing Recall Term in TAR}
\label{supp:additional_ablations_recall_term}

We further explore whether incorporating a recall component into the Temporal Alignment Reward (TAR) improves video reasoning. 
In this variant, we compute both precision- and recall-based temporal alignment scores before applying the TAC gate. 
The recall component (\(\text{TAR}_{\text{rec}}\)) is defined as:
\begin{equation}
\mathrm{TAR}_{\text{rec}} = \frac{1}{m} \sum_{ij} X_{ij},
\end{equation}
where \(m\) is the number of reference claims. 
We then combine the precision (\(\mathrm{TAR}_{\text{prec}}\)) and recall components (\(\mathrm{TAR}_{\text{rec}}\)) using the harmonic mean:
\begin{equation}
\mathrm{TAR}_{F1} = \frac{2 \times \mathrm{TAR}_{\text{prec}} \times \mathrm{TAR}_{\text{rec}}}{\mathrm{TAR}_{\text{prec}} + \mathrm{TAR}_{\text{rec}}}.
\end{equation}
This \( \mathrm{TAR}_{F1} \) replaces the precision-only term in the total reward, while the consistency gating remains active.
\begin{equation}
R_{\text{total}} = 
\lambda_{\text{acc}} R_{\text{acc}} +
\lambda_{\text{fmt}} R_{\text{fmt}} +
\lambda_{\text{tar}} (g \cdot \mathrm{TAR}_{F1}).
\end{equation}
The results in Tab.~\ref{supp:ablation_1_2} show that adding the recall term slightly improves temporal completeness (higher VAS) but lowers accuracy, especially on reasoning benchmarks. 
This drop likely because the recall component encourages overgeneration of timestamps, rewarding redundant or loosely aligned reasoning claims. 
This observation aligns with the inherent incompleteness of reference annotations, where not every relevant event is exhaustively labeled, thus, overfitting to recall may not yield better reasoning quality. 
Empirically, the precision-based TAR used in~\MyModel~remains more stable and better aligned with reasoning quality.

\subsection{Stability of VAS Metric}
\label{supp:stability_of_vas}

To assess the robustness of our proposed Video Attention Score (VAS) metric against the choice of large language model (LLM) used as the judge, we compute VAS using two different models: \textbf{Qwen3-Next-80B-A3B} (our default) and \textbf{Qwen3-32B}. 
The default model features a larger total parameter count but fewer active parameters, offering an optimal balance between speed and evaluation accuracy, while the Qwen3-32B is a dense 32B parameter model used in its non-reasoning mode. 
For each benchmark and reasoning model, we calculate sample-level VAS scores using both LLMs and measure their linear correlation through the \textit{Pearson Correlation Coefficient (PCC)}. 
This process is repeated across all 11 benchmarks (five generic and six reasoning) for five reasoning models, resulting in a comprehensive cross-model stability analysis in Tab.~\ref{supp:table_vas_pcc}. 

As shown in Tab.~\ref{supp:table_vas_pcc}, all PCC values remain high, with overall averages exceeding 0.7, indicating a strong linear correlation between the VAS computed by the two LLMs. 
This demonstrates that the relative ranking of models across benchmarks remains consistent even when the judge model changes, confirming that VAS is a stable and reliable indicator of visual grounding strength. 
Further, the corresponding $p$-values for all correlations were effectively zero (in the range of $10^{-35}$ -- 0.0), showing that these correlations are statistically significant. 
While Qwen3-32B produces consistent rankings, we recommend using \textbf{Qwen3-Next-80B-A3B}, which is openly available, to ensure complete reproducibility of all reported VAS values and maintain exact correspondence with our main results.

\subsection{Sensitivity Analysis}
\label{supp:sensitivity_analysis}

\noindent
\textbf{Sensitivity of $\tau$, $\Delta$, and $\lambda_{\text{tar}}$.}
The hyperparameters used in the Temporal Alignment Reward (TAR), the semantic similarity threshold ($\tau$), the temporal tolerance ($\Delta$), and the TAR weighting factor ($\lambda_{\text{tar}}$), are selected empirically. The search objective focuses solely on maximizing answer accuracy; reasoning quality metrics such as TAC or VAS are not considered during this process. To avoid overfitting and ensure efficient tuning, experiments are conducted on a representative subset of 1,000 randomly sampled instances from the 4.8K GRPO dataset and evaluated using a subset of the generic benchmarks. This setup follows standard machine learning practice, where hyperparameters are optimized on a held-out subset rather than the full training dataset.

For $\tau$ and $\Delta$, a sequential search is performed. First, $\tau$ is fixed at 0.75, and four different $\Delta$ values are tested. Next, using the best $\Delta$, $\tau$ is varied across four candidate settings. From these runs, $\Delta = 2$ s and $\tau = 0.75$ provide the best overall score on generic benchmarks. The selected values also align qualitatively with observations in Figs.~\ref{fig:tar_ex2}--\ref{fig:tar_ex9_p2}.

For $\lambda_{\text{tar}}$, values ${0.5, 1.0, 1.5}$ are tested while keeping all other coefficients fixed. Setting $\lambda_{\text{tar}} = 1.0$, equal to the weights of the accuracy and format rewards, results in the best overall performance. These hyperparameters are determined using limited subsets rather than the complete dataset or all benchmarks. The final values, $\tau = 0.75$, $\Delta = 2$ s, and $\lambda_{\text{tar}} = 1.0$, remain fixed in all subsequent experiments.

%%%%%%%%%%%%%  Supp: Ablations (Comparison with the reported numbers) %%%%%%%%%%%%%
\begin{table*}[!t]
\centering
\setlength{\tabcolsep}{0.75em}
\begin{adjustbox}{max width=\textwidth}
\begin{tabular}{
  l|                               % Model
  c|                               % Max Frames
  *{5}{c}                          % 5 Generic benches
  |>{\columncolor{gray!10}}c|      % Avg Generic (boxed by vertical rules)
  *{6}{c}                          % 6 Reasoning benches
  |>{\columncolor{gray!10}}c|      % Avg Reasoning (boxed)
  >{\columncolor{gray!10}}c        % Avg Overall (boxed; right border too)
}
\toprule
% Group headers (span includes averages)
 &
 &
\multicolumn{6}{c|}{\textbf{Generic Benchmarks}} &
\multicolumn{7}{c|}{\textbf{Reasoning Benchmarks}} &
\\
\cmidrule(lr){3-8}\cmidrule(lr){9-15}
% Rotated, bottom-aligned column headers (including averages)
\raisebox{-15ex}{\textbf{Model}} &
\rothead{Max Frames} &
\rothead{MVBench} &
\rothead{VideoMME} &
\rothead{TempCompass} &
\rothead{MLVU} &
\rothead{LongVideoBench} &
\rothead{\textbf{Avg. (Generic)}} &
\rothead{VideoMathQA} &
\rothead{Video-MMMU} &
\rothead{MMVU-Val} &
\rothead{VSiBench} &
\rothead{MINERVA} &
\rothead{SciVideoBench} &
\rothead{\textbf{Avg. (Reasoning)}} &
\rothead{\textbf{Avg. (Overall)}} \\
\midrule
% ------------------- Data rows -------------------
\multicolumn{16}{l}{\textbf{Accuracy} (Reported)} \\

Video-R1~\cite{videor12025} &  
64 & 64.8 & 61.4 & 73.2 & - & - & - & - & \textbf{52.4} & 63.8 & \underline{37.1} & - & - & - & -  \\

VideoChat-R1~\cite{videochatR12025} &  
NA & 66.2 & - & - & - & - & - & - & - & - & - & - & - & - & - \\

VideoChat-R1.5~\cite{yan2025videochat_r1.5} &  
2048 & \textbf{70.6} & \textbf{65.2} & - & \textbf{70.1} & \textbf{61.4} & - & - & 49.6 & - & - & - & - & - & -  \\

VideoRFT~\cite{videorft2025} &  
32 & 62.1 & 59.8 & \underline{73.7} & - & - & - & - & \underline{51.1} & \textbf{68.5} & 36.8 & - & - & - & -  \\

\rowcolor{gray!10}
\MyModel~(Ours) &
128 & \underline{67.5} & \underline{63.8} & \textbf{74.9} & \underline{68.3} & \underline{59.2} & \textbf{66.7} & \textbf{28.8} & 50.8 & \underline{67.4} & \textbf{39.4} & \textbf{34.9} & \textbf{28.4} & \textbf{41.6} & \textbf{53.0} \\

\midrule

\multicolumn{16}{l}{\textbf{Accuracy} (Reproduced)} \\

Video-R1~\cite{videor12025} & 128 &
65.1 & \underline{64.3} & 73.5 & \underline{67.7} & \underline{57.6} &
\underline{65.6} &
23.6 & 47.0 & 64.0 & 37.8 & 33.9 & \underline{26.8} &
38.8 &
51.0 \\

VideoChat-R1~\cite{videochatR12025} & 128 &
63.6 & 64.1 & \underline{74.5} & 62.5 & 54.3 &
63.8 &
24.3 & \textbf{52.0} & 64.8 & 33.0 & 33.8 & 26.5 &
39.1 &
50.3 \\

VideoChat-R1.5~\cite{yan2025videochat_r1.5} & 128 &
\underline{65.7} & \textbf{64.8} & 73.9 & 65.3 & 53.6 &
64.7 &
\underline{26.4} & 50.0 & \underline{67.0} & 36.1 & 33.7 & 25.8 &
\underline{39.8} &
51.1 \\

VideoRFT~\cite{videorft2025} & 128 &
64.8 & 64.1 & 73.8 & 66.6 & 57.0 &
65.3 &
25.2 & 48.1 & 66.7 & \underline{38.5} & \underline{34.0} & 25.7 &
39.7 &
\underline{51.3} \\

\rowcolor{gray!10}
\textbf{\MyModel~(Ours)} & 128 &
\textbf{67.5} & 63.8 & \textbf{74.9} & \textbf{68.3} & \textbf{59.2} &
\textbf{66.7} &
\textbf{28.8} & \underline{50.8} & \textbf{67.4} & \textbf{39.4 }& \textbf{34.9} & \textbf{28.4} &
\textbf{41.6} &
\textbf{53.0} \\

\bottomrule
\end{tabular}
\end{adjustbox}
\vspace{-0.5em}
\caption{
\textbf{Comparison with previously reported numbers and evaluation under a unified protocol.} 
Upper section shows the accuracy values as reported in the original papers of prior methods, which vary significantly in evaluation settings, frame counts, resolutions, and benchmark coverage. 
Lower section presents results of all models re-evaluated under our standardized setup (2\,FPS, 128 frames, 360$\times$420\,px), ensuring consistency across benchmarks. 
Our unified evaluation eliminates confounding factors and allows fair cross-model comparison. 
All datasets, trained models, and code for training and evaluation will be publicly released to support reproducibility.
}
\label{supp:ablation_comparison_with_reported_numbers}
\vspace{-1em}
\end{table*}
%%%%%%%%%%%%%  Supp: Ablations (Comparison with the reported numbers) %%%%%%%%%%%%%

\subsection{Comparison using Reported Numbers}
\label{supp:comparison_with_reported_numbers}

Tab.~\ref{supp:ablation_comparison_with_reported_numbers} compares the performance of recent video reasoning models using the accuracy values reported in their respective papers, alongside our results obtained under the proposed unified evaluation protocol. 
This comparison highlights the inconsistencies that arise from differing evaluation settings and underscores the need for a standardized evaluation protocol.

Reported numbers across existing methods are not directly comparable due to several confounding factors. 
(1) Different methods evaluate using widely varying maximum frame counts, ranging from 32 frames in VideoRFT~\cite{videorft2025} to 2048 frames in VideoChat-R1.5~\cite{yan2025videochat_r1.5}. 
(2) Frame resolution also differ across implementations and are often unspecified. 
(3) The benchmark coverage itself varies, as some methods report results on only a subset of the benchmarks. 
These discrepancies make a direct quantitative comparison between methods unreliable and can hide the underlying progress in reasoning ability.

To address these inconsistencies, we evaluate all prior reasoning models using their official prompts under our unified evaluation protocol, which standardizes frame sampling (2\,FPS, up to 128 frames, 360$\times$420\,px resolution) and ensures consistent benchmarking across all 11 datasets. 
This controlled setup allows an accurate and fair assessment of reasoning quality, free from evaluation bias. 
Refer to the lower half of Tab.~\ref{supp:ablation_comparison_with_reported_numbers} for the results obtained from our evaluation protocol.

Our unified protocol, along with the complete datasets, models, and training and evaluation code, will be publicly released with detailed documentation to facilitate full reproducibility and future research on video reasoning.

%% file: supplemental/train_eval_prompts.tex
\section{Prompts Used in Training and Evaluation}
\label{supp:training_and_evaluation_prompts}

Both during training and evaluation, we append a fixed post prompt to encourage explicit reasoning within \texttt{<think>...</think>} and a final answer within \texttt{<answer>...</answer>} block. The exact post prompt is: 
% provided below.

\begin{tcolorbox}[
  breakable,
  colback=gray!5!white,
  colframe=gray!60!black,
  title={Post Prompt used in training and evaluation},
  width=\linewidth,
  before skip=4pt,after skip=6pt,
  boxsep=4pt,left=2pt,right=2pt,top=2pt,bottom=2pt
]
\small
Please think about this question as if you were a human pondering deeply. Engage in an internal dialogue using expressions such as 'let me think', 'wait', 'Hmm', 'oh, I see', 'let's break it down', etc, or other natural language thought expressions. It's encouraged to include self-reflection or verification in the reasoning process. Provide your detailed reasoning between the \texttt{<think>} and \texttt{</think>} tags, and then give your final answer between the \texttt{<answer>} and \texttt{</answer>} tags.
\end{tcolorbox}